\documentclass[10pt]{article}
\usepackage[preprint]{tmlr}

\usepackage[utf8]{inputenc}
\usepackage{amsmath,amssymb,amsthm,mathtools}
\usepackage{graphicx}
\usepackage{booktabs}
\usepackage{longtable}
\usepackage{array}
\usepackage{multirow}
\usepackage{algorithm}
\usepackage{algpseudocode}
\usepackage{xcolor}
\usepackage{pifont}
\usepackage{microtype}
\usepackage{hyperref}
\usepackage{url}

\graphicspath{{figures/}}

\providecommand{\tightlist}{\setlength{\itemsep}{0pt}\setlength{\parskip}{0pt}}

\DeclareUnicodeCharacter{00A7}{\S}
\DeclareUnicodeCharacter{2013}{--}
\DeclareUnicodeCharacter{2014}{---}
\DeclareUnicodeCharacter{2026}{\ldots}
\DeclareUnicodeCharacter{2032}{\ensuremath{{}^\prime}}
\DeclareUnicodeCharacter{00B0}{\ensuremath{{}^\circ}}
\DeclareUnicodeCharacter{00B7}{\ensuremath{\cdot}}
\DeclareUnicodeCharacter{00B1}{\ensuremath{\pm}}
\DeclareUnicodeCharacter{00D7}{\ensuremath{\times}}
\DeclareUnicodeCharacter{2192}{\ensuremath{\rightarrow}}
\DeclareUnicodeCharacter{2194}{\ensuremath{\leftrightarrow}}
\DeclareUnicodeCharacter{2208}{\ensuremath{\in}}
\DeclareUnicodeCharacter{2248}{\ensuremath{\approx}}
\DeclareUnicodeCharacter{2260}{\ensuremath{\neq}}
\DeclareUnicodeCharacter{2265}{\ensuremath{\geq}}
\DeclareUnicodeCharacter{27C2}{\ensuremath{\perp}}
\DeclareUnicodeCharacter{2713}{\ding{51}}
\DeclareUnicodeCharacter{2717}{\ding{55}}

\title{EVAF: A Test-Retest Protocol for Selective Parametric\\
Consolidation in Large Language Models}

\author{Haoliang Han\\
\addr Institute of Biomedical Strategy, China Pharmaceutical University\\
\email haolianghan1992@cpu.edu.cn}

\hypersetup{
  pdftitle={EVAF: A Test-Retest Protocol for Selective Parametric Consolidation},
  pdfauthor={Haoliang Han},
  pdfsubject={},
  pdfkeywords={},
  colorlinks=true, linkcolor=black, citecolor=black, urlcolor=black,
}

\begin{document}
\maketitle

\begin{abstract}
Long-horizon agentic LLMs raise a memory question retrieval systems (Mem0, Zep, Letta, HippoRAG, LIGHT) do not answer: \emph{which experiences should leave a persistent parametric imprint on model behavior, rather than be retrieved at query time?} Retrieval fetches facts; consolidation decides which experiences should shape how the agent writes and responds after a long interaction history. We operationalize this question through measurable parametric signatures, in the complementary-learning-systems (CLS) spirit \citep{mcclelland1995complementary}, rather than through claims about cognitive identity.

We introduce \textbf{EVAF} (Echo-Valence Attractor Field) as a framework that \textbf{investigates a mechanistic route toward selective persona consolidation} in LLMs: a fast working-memory state on a product-manifold prefix \((S^{D-1}(\sqrt{D}))^P\) coupled to a slow LoRA store through a dual valence-\(\times\)-surprise write gate and an \textbf{oracle-conditioned} admission protocol (heuristic or DistilBERT-SST2 valence \citep{vershynin2018high, eichenbaum2017role}). EVAF should be read as a \textbf{falsifiable protocol for studying selective parametric consolidation}, not as a complete theory of identity formation.

Four contributions:

\textbf{(C1) A falsifiable mechanistic protocol and four signatures} across five counterfactual ablations (C0--C4), multi-seed robustness, and four architectures (\(\times 57\) parameter range): Test-Retest parametric-write checks (\(22/22\) events negative \(\Delta S\) at seed=42); post-consolidation \textbf{operational dynamical-freeze} signature (\(z > 3\) on all V9 runs); geometric lockdown at FP32 precision; capacity- and maturity-aware re-anchoring rules (§5).

\textbf{(C2) Cross-scale audit with explicit regime-of-validity.} Scaling rules from thin-shell concentration and EWC norm consistency are \emph{partially predictive}: they forecast Qwen3-1.7B-Base hyperparameters from surprise statistics alone, but Mistral-7B-Base requires P-band (§5.2') and maturity-aware (§5.4) corrections. The 4-point audit delimits each rule's class --- predictive, fitted, or conjecture (§6.4, §7.2).

\textbf{(C3) Signatures consistent with selective persona consolidation, not retrieval-competition.} On a controlled persona-fact stream (4 seeds, 2 models), V9 lowers recognition CE vs Frozen in every cell. We use \textbf{persona-imprint} as an \textbf{operational term}: the productive-recall hit rate under content-free prompting (§4.11) --- not a cognitive claim about identity. On GPT-2 this metric shows a \textbf{\(54\times\) amplification} (\(54\% \pm 44\%\) vs \(1\%\) frozen) --- a large but high-variance effect, not significant at \(n=4\) (§7.4). On long streams, V9's signature is \textbf{consistency} (\(3\times\) lower variance than B1b), not higher mean. R2 and a §4.12 eval-time hybrid confirm \textbf{functional orthogonality}: parametric encodes typology, retrieval delivers verbatim specifics (§6.3).

\textbf{(C4) Honest positioning.} EVAF is in the mechanistic-interpretability lineage (ROME, linear-representation hypothesis, superposition) rather than a leaderboard system. Its core deliverables are the \textbf{selective consolidation gate} and \textbf{Test-Retest protocol} --- a falsifiable harness for detecting parametric writes with replay-retention checks, distinguishing optimization artefacts from rote fitting.

The corpus, generator, hyperparameter table, productive-recall dataset, and event-level Test-Retest traces are versioned with the \texttt{v9.0-frozen} state used for all claims.

\end{abstract}

\section{Introduction}
Large language models deployed as long-horizon agents accumulate interaction histories that admit two memory operations. \textbf{Retrieval} answers ``what did the user say earlier?'' at query time. \textbf{Consolidation} asks which experiences should leave a \textbf{persistent parametric imprint} on how the agent writes and responds, rather than remain lookup-only facts. \textbf{We operationalize this question through measurable parametric signatures} (§4, §4.10--§4.11), \textbf{not claims about cognitive identity.} The 2024-2026 landscape of LLM-memory systems has produced excellent retrieval architectures --- Mem0, Zep, Letta/MemGPT \citep{packer2023memgpt}, HippoRAG \citep{gutierrez2024hipporag}, LIGHT \citep{light2025}, EmergenceMem and Mastra OM --- that achieve \(90\)-\(99\%\) on verbatim- recall benchmarks such as LongMemEval \citep{wu2024longmemeval} and LoCoMo \citep{maharana2024evaluating}. These systems work, and we do not propose to replace them. They are also, by construction, retrieval systems: they store and look up; they do not consolidate into the model's parameters.

This paper takes up the orthogonal question: \textbf{what does a selective parametric consolidation mechanism look like, and how does it behave across LLM scales?} Specifically, we investigate whether a parameter-efficient LLM can exhibit \emph{measurable signatures consistent with selective persona consolidation} --- deciding which fragments of stream content are significant enough to admit to a slow parametric store --- in a way that admits falsifiable mechanistic measurements and derivable cross-scale rules. We position our work in the lineage of mechanistic interpretability for memory (ROME \citep{meng2022locating}; the linear-representation hypothesis \citep{park2024geometry}; superposition \citep{elhage2022superposition}) rather than as a competitor to retrieval-augmented systems on benchmarks they were designed for; we are explicit in §6.3 about retrieval and EVAF operating on different timescales (query-time vs.~cumulative persona-formation) and being architecturally complementary, not zero-sum.

\textbf{Why ``persona consolidation'' and not just ``memory''?} The classical complementary- learning-systems (CLS) framework \citep{mcclelland1995complementary, kumaran2016learning} posits that a hippocampal fast store rapidly binds arbitrary episodes while a neocortical slow store consolidates only the \emph{subset} that interacts with prior knowledge and emotional salience. In humans and other animals, this selective consolidation is what produces \emph{personality} and \emph{identity} --- the cumulative shape of the cortical store after a lifetime of experiences --- not what produces fact lookup \citep{eichenbaum2017role, schapiro2017complementary}. The biological filter is a \emph{dual} gate: novel content (high surprise) that is \emph{also} emotionally tagged (high valence) is what shapes the slow store. EVAF instantiates this dual-gate selectivity in an LLM at modern scale; the §4 mechanistic signatures are \textbf{consistent with} selective parametric consolidation and a measurable persona-imprint signature --- not a proof of identity formation or query-time recall superiority. The closest prior LLM-side work that operationalises a CLS-like pattern is \textbf{Larimar} \citep{das2024larimar} (external content-addressable episodic memory, parameter-free at write), \textbf{FastSlow networks} \citep{mujika2017fast}, \textbf{OML} \citep{javed2019meta}, and \textbf{FearNet} \citep{kemker2018fearnet}; none of these contact LLM hidden-state geometry at modern scale, none implement a dual valence-\(\times\)-surprise gate, and none provide the mechanistic-evidence protocols that this paper centres.

Continual-learning baselines that \emph{do} run on LLMs --- naive LoRA updates \citep{razdaibiedina2023continual}, experience replay \citep{rolnick2019experience, chaudhry2019tiny}, and EWC-style anchors \citep{kirkpatrick2017overcoming, zenke2017continual} --- provide the algorithmic ingredients we re-use, but \textbf{none characterise the dynamics that govern the consolidation decision itself, nor provide a falsifiable protocol for detecting when parametric writes occur and persist}. This is the gap we address.

We propose \textbf{EVAF (Echo-Valence Attractor Field)}, a framework for investigating selective parametric consolidation in LLMs. EVAF treats the model's prefix hidden state \(\mathbf{s}(t) \in \mathbb{R}^{P \times D}\) as a fast-timescale working memory constrained to the per-token product manifold \((S^{D-1}(\sqrt{D}))^P \subseteq S^{PD-1}(\sqrt{PD})\) --- a hard-constraint limit of the high-dimensional thin-shell concentration phenomenon \citep{vershynin2018high}, implemented by per-token projection (§3.1) --- and the LoRA-augmented attention weights \(\theta_{\text{LoRA}}\) as a slow-timescale long-term store. The two stores are coupled by a single probabilistic write gate

\[
P_{\text{write}}(t) \;=\; \sigma\!\big(k_v(V_t-\tau_v)\big) \cdot \sigma\!\big(k_s(S_t-\tau_s)\big)
\]

combining sample valence \(V_t\) (from an \textbf{external oracle} --- see §3.3, §7.14) and epistemic surprise \(S_t\) (the cross-entropy of the model on the sample under the current prefix). When \emph{both} signals exceed their thresholds, the sample is admitted to a polarized buffer and consolidated via a brief LoRA inner loop with experience replay and an EWC-style anchor; otherwise it circulates on the working-memory attractor and is implicitly forgotten. This dual gate, the specific spherical-attractor formulation, and the experimental protocol below are, to our knowledge, original; the underlying components --- LoRA \citep{hu2021lora}, experience replay \citep{rolnick2019experience}, EWC \citep{kirkpatrick2017overcoming}, and the FEP-grounded free-energy interpretation \citep{friston2010free} --- are not, and we use them as established machinery.

To characterise whether parametric writes occur and are retained across past content --- and to rule out optimisation artefacts --- we introduce a \textbf{Test-Retest mechanistic protocol}: for every consolidation event we compute the surprise of the model on the buffer sample(s) and on a replay batch of past samples \emph{before} and \emph{after} the LoRA inner loop, while freezing the working-memory state. A negative \(\Delta S_{\text{buf}}\) on the buffer is consistent with a parametric write on the admitted material; a negative \(\Delta S_{\text{rep}}\) on the replay set --- which is included in the \emph{same} inner-loop update --- is a \textbf{replay-retention (anti-forgetting) check on trained past samples}, not a held-out generalization measure (§7.5). The protocol is falsifiable: a counterfactual that nulls the LoRA optimizer produces \(\Delta S = +0.000\) at machine precision across all events on both architectures we evaluate.

Across counterfactual ablations (six variants), multi-seed robustness (five seeds for GPT-2, four for TinyLlama), \textbf{four} base architectures spanning 124M to 7B parameters (GPT-2 with Conv1D attention, TinyLlama-1.1B-Chat with Llama-style RoPE, Qwen3-1.7B-Base with Qwen3 GQA, Mistral-7B-v0.3-Base with Mistral GQA), two valence-oracle implementations (a heuristic and a DistilBERT-SST2 classifier), and a quantitative perturbation study (30 random-direction kicks at three intensities), EVAF exhibits four reproducible mechanistic signatures:

\begin{enumerate}
\def\labelenumi{\arabic{enumi}.}
\tightlist
\item
  \textbf{Test-Retest parametric-write signature} --- across all four models, \(\Delta S_{\text{buf}} < 0\) on \textbf{every} consolidation event of the seed=42 reference run (8/8 on GPT-2, 3/3 on TinyLlama-Chat, 6/6 on Qwen3-Base, 5/5 on Mistral-Base; total \(22/22\) events with no exceptions), and at multi-seed the mean per-event \(\Delta S_{\text{buf}}\) stays negative on every seed of every model (GPT-2 5-seed mean \(-2.89 \pm 0.40\), TinyLlama-Chat 4-seed mean \(-9.75 \pm 6.40\)). The same drop appears on the replayed past samples (\(\Delta S_{\text{rep}}\) tracks \(\Delta S_{\text{buf}}\) with the same sign and comparable magnitude) --- consistent with a parametric write plus replay-driven retention of past content (not held-out generalization; §7.5), not a cognitive claim of ``internalisation succeeded.''
\item
  \textbf{Operational dynamical-freeze signature} --- post-consolidation, the working-memory trajectory collapses from a \(\rho_{10} = -0.87\) limit cycle (V5 baseline, \(17.7°\) per step) to a fixed-point attractor: \(7.73°\) on GPT-2, \(2.39°\) on TinyLlama-Chat, \(2.10°\) on Qwen3-Base, \(4.08°\) on Mistral-Base. A \(z\)-score test on the pre/post step-angle distribution separates V9 from all five counterfactual ablations on every model (§4.3, §7.15; we do not claim a formal dynamical-systems phase transition).
\end{enumerate}

\begin{figure}[t]
\centering
\includegraphics[width=\linewidth]{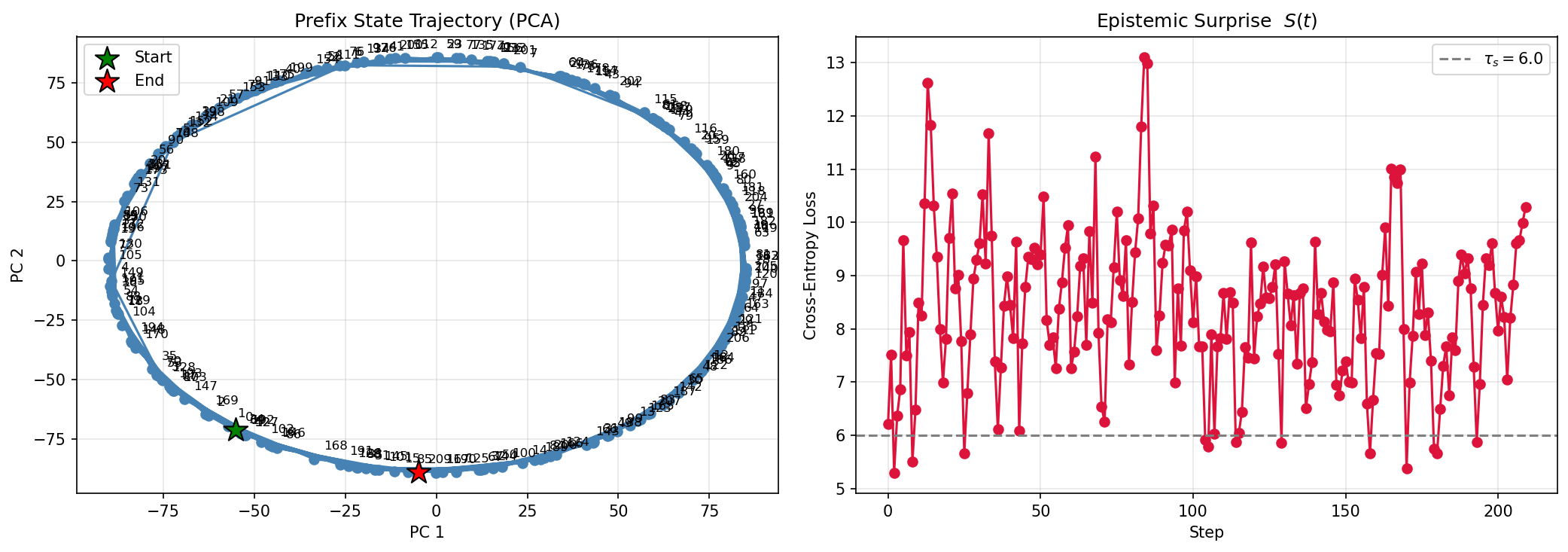}
\caption{V5 limit-cycle working-memory attractor (the \emph{pre-consolidation} trajectory). The hidden state \(\mathbf{s}(t)\) traces a closed orbit on the sphere \(\|\mathbf{s}\|^2 = PD\) with \(\rho_{10} = -0.87\) lag-10 anti-correlation and mean step angle \(17.7°\). Post-consolidation, the same trajectory collapses to a fixed-point attractor (§4.3, Figure~\ref{fig:ablation-seed42}); this collapse is the \emph{operational dynamical-freeze} signature.}
\label{fig:trajectory-v5}
\end{figure}

\begin{enumerate}
\def\labelenumi{\arabic{enumi}.}
\setcounter{enumi}{2}
\tightlist
\item
  \textbf{Geometric lockdown} --- the spherical constraint holds at machine precision, \(\big|\,\|\mathbf{s}\|_2 - \sqrt{PD}\,\big| < 1.3 \times 10^{-5}\) across all four models (max absolute norm deviation from \(\sqrt{PD}\), range \(4.34\times10^{-6}\) to \(1.32\times10^{-5}\)) and survives 8-sigma random kicks with \(100\%\) basin return in \(T_{\text{rec}} \leq 2\) steps, empirically validating the FP32-round-off bound implied by the thin-shell concentration limit (§5.1).
\item
  \textbf{Capacity- and maturity-aware re-anchoring} --- naive transfer of GPT-2-tuned hyperparameters to larger models triggers V8-style catastrophic forgetting; on Mistral-7B-Base, the single-factor scaling rule we initially proposed fails badly (\(\Delta S^{\text{global}}{=}{+}12.57\), i.e., the model worsens rather than improves on the corpus). The empirical resolution requires three rules working in concert (§5): an adaptive percentile \(\tau_s\) that widens from \(\mathrm{Pct}_{50}\) to \(\mathrm{Pct}_{75-90}\) as the warmup-S distribution narrows; a multi-factor Langevin anchor \(\lambda_{\mathrm{reg}}\) with model-class--dependent capacity coupling; and a maturity-aware \(\mathrm{lr}_{\text{LoRA}}\) that shrinks for deeply-pretrained Base models. The 4-model empirical audit precisely delimits each rule's regime of validity and identifies open theoretical questions (§5.5).
\end{enumerate}

\textbf{Contributions.} Our contributions are:

(C1) \emph{Selective consolidation gate + Test-Retest protocol}: the \(P_{\text{write}}\) dual-gate and paired before/after surprise measurements, together with the operational dynamical-freeze \(z\)-score, constitute a \textbf{falsifiable observation toolkit} for selective parametric consolidation that other continual-learning methods can adopt.

(C2) \emph{Cross-architecture and cross-scale validation} over a \(\times 57\) scale range and four distinct attention architectures: GPT-2 (124M, Conv1D), TinyLlama-1.1B-Chat (Llama RoPE), Qwen3-1.7B-Base (Qwen3 GQA), and Mistral-7B-v0.3-Base (Mistral GQA). The four mechanistic signatures (C1) reproduce on every model; the scaling rules of (§5) that we derived from thin-shell concentration and EWC norm consistency are \emph{partially} predictive --- they correctly forecast Qwen3-Base hyperparameters from base-model surprise statistics alone, but fail quantitatively on Mistral-Base, where a maturity-aware revision is required.

(C3) \emph{Counterfactual and oracle-invariant evidence}: a six-variant ablation matrix combined with a five-seed (GPT-2) / four-seed (TinyLlama) robustness study, a DistilBERT-SST2 oracle re-run, and a thirty-trial quantitative perturbation study. All reported numbers reproduce bit-for-bit from the \texttt{v9.0-frozen} state.

(C4) \emph{External-baseline matrix at multi-seed} (§4.9). Five baselines at matched corpus and seed. Multi-seed analysis shows V9 is best characterised as a \textbf{selectivity-and-stability mechanism} --- lowest trigger-count variance, healthy \(\Delta S^{\text{global}}\) distribution, and the only method preserving the operational dynamical-freeze signature on every seed --- rather than deeper per-event Test-Retest than LoRA baselines (per-event \(\Delta S^{\mathcal{B}}\) overlaps within \(1\sigma\) on GPT-2).

(C5) \emph{Mechanism-to-persona transfer} (§4.10--§4.12). Recognition: V9 beats Frozen in every cell; B3b RAG wins on verbatim CE by design. Production: \textbf{operational persona-imprint} metric --- \(54\times\) amplification on GPT-2 short corpus (\(54\% \pm 44\%\) vs \(1\%\) frozen; §4.11); on long streams V9's signature is \textbf{consistency} (\(3\times\) lower variance than B1b), not higher mean. R2 discriminating-token tests and §4.12 eval-time hybrid confirm parametric typology vs retrieval verbatim are \textbf{functionally orthogonal} (§6.3).

(C6) \emph{Second-order findings} of independent interest: an ``Oracle Precision Amplifier'' effect (oracle precision scales mechanism strength without changing direction); ``Capacity-Dependent Mechanism Reweighting'' (at the measured \(\times 1.91\) LoRA capacity ratio the individual necessity of replay and EWC weakens while their joint necessity is preserved); a ``Basin Wall Geometry'' surrounding the limit-cycle attractor; and a ``Noise-Fitting Trap'' delimiting the validity of the Test-Retest parametric-write signal.

\textbf{What we deliberately do \emph{not} claim.} EVAF should be interpreted as a \textbf{falsifiable protocol for studying selective parametric consolidation}, not a complete theory of identity formation. We do not beat retrieval-augmented systems on verbatim-recall benchmarks, and are not designed to. We report a Sentence-BERT MiniLM RAG baseline at parity with our parametric methods in §4.10 and observe that on a verbatim- recall \emph{recognition} metric it produces lower cross-entropy than every parametric method including V9 --- this is structurally expected, since a metric that rewards exact-content retrieval is the natural strength of retrieval. We position the parametric path (EVAF) and the retrieval path (RAG / Mem0 / Zep / LIGHT) as \textbf{operating on different timescales for different purposes}: retrieval is the right tool for query-time fact lookup; EVAF is the right tool for the cumulative shaping of an agent's persona over a long interaction history (§4.11, §6.3). We also do not claim a closed-form predictive scaling rule across all base models; the Mistral-Base audit (§4.8, §5.4) reveals an \(\mathrm{lr}_{\text{LoRA}}\) co-revision supported by a single data point, which we report as a conjecture and list as the single most decision-relevant follow-up experiment (§7.2). We list ten further limitations in §7, including the end-of-stream evaluation-drift caveat that emerged from an extended-stream stress test (§7.11).

The rest of the paper proceeds as follows. Section 2 surveys the closest prior art, with particular care for the boundary between our contributions and the established CLS / replay / EWC / LoRA / memory-augmented-LLM literatures, including the 2024-2026 retrieval-augmented memory landscape (§2.4). Section 3 specifies the framework. Section 4 presents the four mechanistic signatures and their counterfactual / robustness / cross- model / cross-baseline variants, the persona-fact retention task that tests mechanism-to-task transfer (§4.10), and the productive-recall free-generation experiment that demonstrates the long-timescale persona imprint (§4.11). Section 5 develops the thin-shell scaling rule, the adaptive percentile \(\tau_s\), and the multi-factor Langevin EWC scaling that together motivate the cross-model re-anchoring, and reports the 4-point empirical audit that delimits each rule's regime of validity. Section 6 discusses implications, the architectural-complementarity argument between EVAF and retrieval on different timescales (§6.3), and a brief positioning note for the cognitive-science audience. Section 7 lists eleven honest limitations.

\section{Background and Related Work}
\subsection{Complementary Learning Systems (CLS) and dual-store memory}

The complementary learning systems hypothesis \citep{mcclelland1995complementary} posits that the brain maintains two qualitatively distinct memory systems --- a fast-binding hippocampal store and a slow-consolidating neocortical store --- that exchange information through \emph{replay} during off-task periods. Subsequent work refined the timing \citep{kumaran2016learning, schapiro2017complementary} and identified gating signals (novelty / reward) that determine which hippocampal episodes get consolidated. EVAF is \emph{inspired} by CLS but is \textbf{not a cognitive model}: we do not claim to map our Prefix-state attractor to hippocampus or our LoRA-augmented weights to cortex. We use the dual-store \emph{architectural pattern} as a design constraint and reserve neuroscience interpretation for §6 (Discussion).

Three recent ML systems share this architectural pattern. \textbf{FearNet} \citep{kemker2018fearnet} trains a dual-net for incremental classification with generative replay. \textbf{OML} \citep{javed2019meta} meta-learns representations to minimise forgetting. \textbf{FastSlow networks} \citep{mujika2017fast} alternate fast-RNN inner steps with slow-weight consolidation. None of these contact LLM-scale hidden-state geometry, none use the dual-gate \(V \times S\) structure, and none use the spherical-attractor working memory we develop in §3.1. The closest prior work is \textbf{Larimar} \citep{das2024larimar}, which couples an LLM to an external episodic memory addressable by content; Larimar is parameter- free at memory write (it stores keys in a separate non-LLM memory matrix), while EVAF is parameter-internalising (LoRA edits travel into the model itself).

\subsection{Continuous attractor networks and LLM hidden-state geometry}

Continuous attractor networks \citep{wilson1972excitatory,burak2009accurate} realise short-term memory as a ring or sphere of stable states; perturbations relax back to the manifold under linear-stability conditions on the recurrent weights. The \emph{thin-shell concentration} phenomenon for high-dimensional Gaussians \citep{vershynin2018high} --- i.e., for \(\mathbf{x} \sim \mathcal{N}(0, I_D)\) the norm \(\|\mathbf{x}\|\) concentrates at \(\sqrt{D}\) within \(O(1/\sqrt{D})\) --- is what makes a hard sphere constraint approximately faithful at LLM-scale \(D \gtrsim 10^3\). We exploit this in §3.1 by projecting each prefix token onto its own sphere \(S^{D-1}(\sqrt{D})\) --- the per-token product manifold \((S^{D-1}(\sqrt{D}))^P \subseteq S^{PD-1}(\sqrt{PD})\) --- at every step, and verify in §4.7 that the resulting working-memory trajectory exhibits an \textbf{operational dynamical-freeze} signature (limit cycle \(\to\) fixed point) when consolidation events complete --- a signature absent in pure attractor models without our gating layer.

The growing literature on LLM \textbf{hidden-state geometry} \citep{elhage2022superposition, gurnee2024linear, park2024geometry} has documented that intermediate activations live near low-dimensional manifolds embedded in a high-D ambient space. Our spherical projection is consistent with these findings and provides an \emph{analytic} hard-constraint counterpart to the \emph{empirical} low-rank observations.

\subsection{Continual learning: replay, EWC, and LoRA}

The continual-learning literature is structured around three orthogonal mitigations of catastrophic forgetting \citep{french1999catastrophic}:

\begin{itemize}
\tightlist
\item
  \textbf{Replay} \citep{rolnick2019experience, chaudhry2019tiny}: at each update, interleave the current sample with past samples drawn from a buffer.
\item
  \textbf{Regularisation} \citep{kirkpatrick2017overcoming, zenke2017continual}: add a penalty \(\lambda \cdot \mathrm{Fisher}\!\cdot\!\lVert\theta-\theta_0\rVert^2\) that anchors parameters near their pre-task value.
\item
  \textbf{Architecture} \citep{rusu2016progressive, mallya2018piggyback}: allocate fresh capacity per task, leaving prior tasks' parameters frozen.
\end{itemize}

LoRA \citep{hu2021lora} provides a parameter-efficient capacity-allocation substrate; recent LoRA-based continual schemes \citep{razdaibiedina2023continual,
chitale2023task} combine LoRA with replay or regularisation in static task-sequence settings. EVAF integrates all three of replay, an EWC-style L2 anchor (we use a \emph{uniform} anchor in lieu of Fisher; §3.2), and LoRA capacity, but with two important departures: (i) the \emph{trigger} of a consolidation event is the dual gate \(V \times S\) rather than a fixed task boundary; (ii) the trigger fires at a \emph{fraction of a percent} of stream steps (typical: 3--8 out of 210 turns on our corpus), making EVAF a \emph{selectively-consolidating} rather than continually-updating system. As we show in §4.4 (counterfactual C2) and §4.9 (external baseline B1b), removing the gate and consolidating at every step yields shallow internalisation; conversely, removing EWC (C2, B1a) yields collapse on TinyLlama-1.1B+. The \emph{combination} of selective gating + EWC anchor + replay is what we evaluate.

\subsection{Memory-augmented LLMs (retrieval, structured memory, agent systems)}

A parallel and currently-dominant line of work scales memory by externalising it. \textbf{RAG} \citep{lewis2020retrieval, izacard2023atlas} and \textbf{MemGPT / Letta} \citep{packer2023memgpt} retrieve relevant past contexts at inference time and inject them via the prompt. \textbf{Memorizing Transformers} \citep{wu2022memorizing} extend the attention KV cache with a long-term store. \textbf{Prompt-compression} systems \citep{chevalier2023adapting, ge2024incontext} compress past contexts into a small number of soft tokens. The 2024-2026 wave of \emph{production} long-term- memory systems has converged on multi-layer structured-memory architectures: \textbf{Mem0} and \textbf{Zep} \citep{zep2024} maintain typed memory graphs (episodic / semantic / procedural) addressable by content and updateable through agent tool calls; \textbf{HippoRAG} \citep{gutierrez2024hipporag} adapts neuroscience- inspired Personalised PageRank over a knowledge graph; \textbf{LIGHT} \citep{light2025} introduces a cognitively-organised multi-layer episodic / working / scratchpad memory and reports SOTA on the BEAM benchmark; recent adaptive-RAG systems (\textbf{EmergenceMem}, \textbf{Mastra OM}) push LongMemEval \citep{wu2024longmemeval} to \(94\)-\(99\%\) accuracy through dynamic-\(k\) retrieval with MMR diversification.

\textbf{These systems are excellent at what they are designed for --- query-time fact retrieval --- but by construction they leave the parameters of the LLM unchanged.} They are external memory stores, not consolidation mechanisms. EVAF asks a different question --- \emph{which fragments of stream content should shape the model itself, on the persona-formation timescale?} --- and is therefore positioned as architecturally complementary to retrieval rather than competing with it on the recall benchmarks these systems were built for (§6.3).

The strongest evidence for the complementarity is the productive-recall amplification of §4.11: an EVAF-trained model freely generates persona- specific keywords at a \(4\)-seed mean \(\mathbf{54\times}\) the frozen-base rate on GPT-2 (\(54\% \pm 44\%\) vs \(1\%\); a large but high-variance effect, §7.4) on a prompt that contains \emph{no fact content for a retriever to match}. Retrieval cannot produce this signature even in principle, because the prompt is below the retrieval-relevance threshold; the persona imprint lives in the parameters or it does not exist at all. On the §4.10 recognition CE metric our Sentence-BERT MiniLM RAG baseline (B3b) does produce lower CE than every parametric method including V9, as it should for any metric that admits exact-content retrieval; this is the right outcome for the right metric, and we explicitly do not claim parametric superiority on recall. We position EVAF and retrieval as \textbf{operating on different timescales for different purposes}: retrieval is the query-timescale fact-lookup tool; EVAF is the persona-formation-timescale consolidation tool. We implement a minimal eval-time hybrid in §4.12 (V9 stream + RAG at eval): it confirms the two paths are orthogonal but shows naive concatenation is not sufficient fusion --- write-time routing remains the natural follow-up (§6.3).

\subsection{Mechanistic interpretability for memory}

A small but growing body of work probes how LLMs encode and recall factual knowledge \citep{meng2022locating, geva2022transformer, hernandez2023linearity} and how that encoding updates under fine-tuning \citep{de2021editing, hartvigsen2023aging}. These works diagnose \emph{which parameters} hold a given fact and \emph{how an edit propagates}. EVAF operates at a different level: we do not localise individual facts but instead instrument the \emph{dynamics} of the consolidation decision through the Test-Retest protocol (\(\Delta S_{\text{buf}}\) and \(\Delta S_{\text{rep}}\), §3.4) and the working-memory dynamical-freeze signature (\(z\)-score on the step-angle distribution, §4.3). The two tool-sets are complementary: mechanistic-interpretability work tells us \emph{where} a fact ends up after a write; EVAF tells us \emph{whether and when} the system decides to write.

\subsection{Summary: what is and is not new in EVAF}

\begin{longtable}[]{@{}lll@{}}
\toprule
\begin{minipage}[b]{0.30\columnwidth}\raggedright
Element\strut
\end{minipage} & \begin{minipage}[b]{0.30\columnwidth}\raggedright
Established\strut
\end{minipage} & \begin{minipage}[b]{0.30\columnwidth}\raggedright
Our contribution\strut
\end{minipage}\tabularnewline
\midrule
\endhead
\begin{minipage}[t]{0.30\columnwidth}\raggedright
Dual-store CLS framing\strut
\end{minipage} & \begin{minipage}[t]{0.30\columnwidth}\raggedright
\citep{mcclelland1995complementary}\strut
\end{minipage} & \begin{minipage}[t]{0.30\columnwidth}\raggedright
(architectural inspiration only)\strut
\end{minipage}\tabularnewline
\begin{minipage}[t]{0.30\columnwidth}\raggedright
Spherical hard-constraint on a prefix state\strut
\end{minipage} & \begin{minipage}[t]{0.30\columnwidth}\raggedright
thin-shell limit \citep{vershynin2018high}\strut
\end{minipage} & \begin{minipage}[t]{0.30\columnwidth}\raggedright
the \emph{projection on every step} and its falsifiable lock signature (§4.7)\strut
\end{minipage}\tabularnewline
\begin{minipage}[t]{0.30\columnwidth}\raggedright
LoRA\strut
\end{minipage} & \begin{minipage}[t]{0.30\columnwidth}\raggedright
\citep{hu2021lora}\strut
\end{minipage} & \begin{minipage}[t]{0.30\columnwidth}\raggedright
parameter-efficient capacity substrate\strut
\end{minipage}\tabularnewline
\begin{minipage}[t]{0.30\columnwidth}\raggedright
Replay\strut
\end{minipage} & \begin{minipage}[t]{0.30\columnwidth}\raggedright
\citep{rolnick2019experience}\strut
\end{minipage} & \begin{minipage}[t]{0.30\columnwidth}\raggedright
replay sampling from past stream excluding current buffer\strut
\end{minipage}\tabularnewline
\begin{minipage}[t]{0.30\columnwidth}\raggedright
EWC-style anchor\strut
\end{minipage} & \begin{minipage}[t]{0.30\columnwidth}\raggedright
\citep{kirkpatrick2017overcoming}\strut
\end{minipage} & \begin{minipage}[t]{0.30\columnwidth}\raggedright
uniform-Fisher L2 anchor with scaling rule §5\strut
\end{minipage}\tabularnewline
\begin{minipage}[t]{0.30\columnwidth}\raggedright
Dual gate \(V \times S\)\strut
\end{minipage} & \begin{minipage}[t]{0.30\columnwidth}\raggedright
---\strut
\end{minipage} & \begin{minipage}[t]{0.30\columnwidth}\raggedright
\textbf{new} (\(P_{\text{write}}\) in §3.3)\strut
\end{minipage}\tabularnewline
\begin{minipage}[t]{0.30\columnwidth}\raggedright
Test-Retest protocol\strut
\end{minipage} & \begin{minipage}[t]{0.30\columnwidth}\raggedright
---\strut
\end{minipage} & \begin{minipage}[t]{0.30\columnwidth}\raggedright
\textbf{new} (mechanistic-evidence harness, §3.4)\strut
\end{minipage}\tabularnewline
\begin{minipage}[t]{0.30\columnwidth}\raggedright
Operational dynamical-freeze \(z\)-score\strut
\end{minipage} & \begin{minipage}[t]{0.30\columnwidth}\raggedright
---\strut
\end{minipage} & \begin{minipage}[t]{0.30\columnwidth}\raggedright
\textbf{new} (working-memory freeze signature, §4.3)\strut
\end{minipage}\tabularnewline
\begin{minipage}[t]{0.30\columnwidth}\raggedright
Multi-factor \(\lambda_{\text{reg}}\) scaling rule\strut
\end{minipage} & \begin{minipage}[t]{0.30\columnwidth}\raggedright
---\strut
\end{minipage} & \begin{minipage}[t]{0.30\columnwidth}\raggedright
\textbf{new} (§5, §11.5 of math framework)\strut
\end{minipage}\tabularnewline
\begin{minipage}[t]{0.30\columnwidth}\raggedright
Persona-fact retention as a downstream evaluation of mechanism→task transfer\strut
\end{minipage} & \begin{minipage}[t]{0.30\columnwidth}\raggedright
---\strut
\end{minipage} & \begin{minipage}[t]{0.30\columnwidth}\raggedright
\textbf{new} in the sense of being our specific design (§4.10); the dataset family (PersonaChat) is well-known \citep{zhang2018personalizing}\strut
\end{minipage}\tabularnewline
\begin{minipage}[t]{0.30\columnwidth}\raggedright
Productive-recall persona-imprint metric (operational: free generation from content-free prompt; §4.11)\strut
\end{minipage} & \begin{minipage}[t]{0.30\columnwidth}\raggedright
---\strut
\end{minipage} & \begin{minipage}[t]{0.30\columnwidth}\raggedright
\textbf{new}: generative-level probe consistent with selective consolidation; not a cognitive identity claim\strut
\end{minipage}\tabularnewline
\begin{minipage}[t]{0.30\columnwidth}\raggedright
Reframing of dual-gate selectivity as a \emph{persona-relevance filter} rather than a generic surprise filter\strut
\end{minipage} & \begin{minipage}[t]{0.30\columnwidth}\raggedright
---\strut
\end{minipage} & \begin{minipage}[t]{0.30\columnwidth}\raggedright
\textbf{new} framing, motivated by §4.11 amplification and §6.3 complementarity argument\strut
\end{minipage}\tabularnewline
\bottomrule
\end{longtable}

The remaining sections develop the framework (§3), report the experimental findings (§4), formalise the scaling rules (§5), and discuss implications and honest limitations (§6).

\section{The EVAF Framework}
EVAF couples a \textbf{fast working-memory store} \(\mathbf{s}(t)\) --- a low-rank prefix constrained to the per-token product manifold \((S^{D-1}(\sqrt{D}))^P \subseteq S^{PD-1}(\sqrt{PD})\) --- to a \textbf{slow long-term store} \(\theta_{\text{LoRA}}\) --- rank-\(r\) adapter weights on the attention modules --- via a probabilistic write gate \(P_{\text{write}}(V_t, S_t)\). §3.1 specifies the prefix dynamics; §3.2 the LoRA inner loop with replay and EWC anchor; §3.3 the dual gate; §3.4 the Test-Retest mechanistic protocol that decides whether a write succeeded.

\subsection{System 1 --- Prefix dynamics on a high-dimensional sphere}

Let \(\mathbf{s}(t) \in \mathbb{R}^{P \times D}\) denote the trainable prefix at step \(t\), with \(P\) the prefix length and \(D\) the hidden width of the base model. The prefix is initialised \(\mathbf{s}(0) \sim 0.02 \cdot \mathcal{N}(0, I_{PD})\) and projected onto the sphere \(\|\mathbf{s}\|^2 = PD\). The base model is frozen; gradients flow only into \(\mathbf{s}\) via the cross-entropy loss

\begin{equation}
S_t \;=\; -\,\frac{1}{|y_t|}\sum_{i} \log p_{\theta}\!\big(y_{t,i}\mid y_{t,<i},\, \mathbf{s}(t)\big),
\tag{3.1}
\end{equation}

evaluated on the current input \(y_t\) with the prefix prepended. We refer to \(S_t\) as the \textbf{epistemic surprise} at step \(t\).

A single update step combines three terms:

\begin{equation}
\mathbf{s}(t+1) \;=\; \Pi_{(S^{D-1}(\sqrt{D}))^P}\!\Big(\,\mathbf{s}(t)
  \;+\;\alpha\,\big[\,{-}\nabla_{\mathbf{s}} S_t
  \;+\;\lambda_{\text{con}}\,\big(\mathbf{s}(t) - \mathbf{s}_{\text{EMA}}(t)\big)\,\big] \,\Big), \tag{3.2}
\end{equation}

with a \textbf{gate-modulated} exponential moving average \(\mathbf{s}_{\text{EMA}}(t) = (1-\rho_t)\,\mathbf{s}_{\text{EMA}}(t-1) + \rho_t\,\mathbf{s}(t)\) whose rate \(\rho_t = \rho\,\sigma\!\big(k\,(S_t-\tau_s)\big)\) tracks the surprise gate (base rate \(\rho\); the anchor slides quickly when surprise is high and is nearly frozen when surprise is low), and \(\Pi\) the \textbf{per-prefix-token} projection that rescales each of the \(P\) token vectors to radius \(\sqrt{D}\), so the joint state lies on the product manifold \((S^{D-1}(\sqrt{D}))^P \subseteq S^{PD-1}(\sqrt{PD})\). The first term descends along the surprise gradient; the second is a \textbf{repulsive (contrastive) inertia} term that pushes the state \emph{away} from its own slow EMA, keeping the pre-consolidation working memory in a sustained limit cycle rather than collapsing it (an EMA-sign control, §4.3, shows that flipping this term to an attractive ``pull toward the EMA'\,' removes the limit cycle and destroys the freeze signature). The projection enforces the spherical hard constraint. \textbf{No fine-tuning} is performed on either the prefix or the base model during step (3.2); changes to base parameters arrive only via the LoRA inner loop in §3.2 below.

The spherical constraint is faithful because of the thin-shell concentration of measure in high dimension \citep{vershynin2018high}: an unconstrained Gaussian initialisation in \(\mathbb{R}^{PD}\) already concentrates within \(O(1/\sqrt{PD})\) of the sphere of radius \(\sqrt{PD}\). The projection \(\Pi\) therefore corrects an \(O(1/\sqrt{D})\) residual rather than a leading-order deformation; §5.1 derives a quantitative bound on the surviving deviation.

\subsection{System 2 --- LoRA + experience replay + EWC anchor}

We attach LoRA adapters \citep{hu2021lora} with rank \(r = 32\) to the attention projections of the base model (\(q\)/\(k\)/\(v\)/\(o\) or, for GPT-2, the Conv1D analogues). When a consolidation event is fired at step \(t^\ast\) (see §3.3 for the trigger condition), the inner loop runs \(N_{\text{inner}} = 5\) AdamW steps on the joint loss

\begin{equation}
\mathcal{L}_{\text{LoRA}}(\theta) \;=\; \frac{1}{|B \cup R|} \sum_{y \in B \cup R}
  \mathrm{CE}\!\big(y \mid \mathbf{s}(t^\ast),\, \theta\big) \;+\; \lambda_{\text{reg}}\,\|\theta - \theta_0\|_2^2, \tag{3.3}
\end{equation}

where \(B\) is the \emph{polarized buffer} (admitted samples since the previous event; capacity \(|B| = 4\)), \(R\) is a \(K = 4\)-sample experience replay drawn uniformly from the global stream of all past inputs excluding \(B\), and \(\theta_0\) is the LoRA weights at adapter-attach time. The L2 anchor \(\lambda_{\text{reg}}\,\|\theta - \theta_0\|_2^2\) approximates an EWC penalty \citep{kirkpatrick2017overcoming} with a uniform Fisher; §5.3 derives the scaling rule that determines \(\lambda_{\text{reg}}\) as a function of base model. The prefix \(\mathbf{s}(t^\ast)\) is held fixed during the inner loop --- gradients update only LoRA weights --- which decouples the slow store update from the fast store dynamics.

After the inner loop, the buffer is cleared. The dynamics in (3.2) resume with the now-modified base model (the LoRA delta is added to the frozen base through the standard PEFT path).

\subsection{Dual gate --- what gets admitted to the buffer}

Each input \(y_t\) carries a \emph{valence} \(V_t \in [0, 1]\) from an \textbf{external oracle} --- EVAF is an \textbf{oracle-conditioned consolidation protocol}, not a self-contained valence mechanism (§7.14). In the main experiments we use a calibrated lexical-affect heuristic; in the Patch-A robustness study, a DistilBERT-SST2 classifier. Epistemic surprise \(S_t\) comes from (3.1). The dual gate is

\begin{equation}
P_{\text{write}}(V_t, S_t) \;=\;
\sigma\!\big(k_v\,(V_t - \tau_v)\big) \cdot
\sigma\!\big(k\,(S_t - \tau_s)\big), \tag{3.4}
\end{equation}

and \(y_t\) is admitted to the buffer iff \(P_{\text{write}}(V_t, S_t) > \tau_{\text{write}} = 0.55\). The product form requires \emph{both} gates to open --- a sample with only high valence or only high surprise is rejected. We show in §4.4 that ablating either gate factor degrades the mechanism in the expected direction: surprise-only gating (B2) yields catastrophic forgetting on TinyLlama; valence-only gating (omitted for brevity) yields under-firing.

A consolidation event triggers when the buffer reaches its capacity \(|B| = 4\); the buffer is then frozen as input to (3.3) and a new buffer begins to accumulate.

\subsection{Test-Retest --- protocol for parametric-write detection}

To characterise whether consolidation produces a \textbf{parametric write with a replay-retention (anti-forgetting) check} rather than merely consuming compute, we instrument every consolidation event with a paired before/after measurement. Let \(\theta^{\text{pre}}\) and \(\theta^{\text{post}}\) denote the LoRA weights immediately before and after the inner loop. We compute four quantities:

\begin{equation}
\Delta S^{\mathcal{B}}_{t^\ast} \;=\; \frac{1}{|B|}\sum_{y \in B}\!\big[ S(y; \mathbf{s}(t^\ast), \theta^{\text{post}}) - S(y; \mathbf{s}(t^\ast), \theta^{\text{pre}}) \big], \tag{3.5a}
\end{equation}

\begin{equation}
\Delta S^{\mathcal{R}}_{t^\ast} \;=\; \frac{1}{|R|}\sum_{y \in R}\!\big[ S(y; \mathbf{s}(t^\ast), \theta^{\text{post}}) - S(y; \mathbf{s}(t^\ast), \theta^{\text{pre}}) \big], \tag{3.5b}
\end{equation}

\begin{equation}
\Delta S^{\text{global}} \;=\; \frac{1}{T-t_1}\sum_{t=t_1}^{T} S_t \;-\; \frac{1}{t_1}\sum_{t=1}^{t_1} S_t, \tag{3.5c}
\end{equation}

with the prefix held fixed at \(\mathbf{s}(t^\ast)\) for (3.5a, b), and \(t_1\) the first consolidation (LoRA-trigger) step, \(S_t\) the online per-step surprise (3.1), and \(T\) the stream length. The first quantifies whether the inner loop reduced surprise on the buffer. The second measures the change on the replay set \(R\); because \(R\) is part of the \emph{same} inner-loop batch (\(B \cup R\), EQ 3.3), \(\Delta S^{\mathcal{R}}\) is a \textbf{replay-retention (anti-forgetting) check on trained past samples}, not a held-out generalization measure (§7.5). The third compares the model's \textbf{online stream surprise} in the post-trigger window (\(t \ge t_1\)) against its pre-trigger baseline (\(t < t_1\)) --- an online stream-level baseline-shift / \textbf{collapse detector}, not a fixed held-out final-vs-initial corpus metric: a healthy run gives \(\Delta S^{\text{global}} < 0\) with small magnitude, while a positive or extreme-negative value indicates catastrophic forgetting / runaway memorisation.

The fourth signature is the \textbf{operational dynamical-freeze} signature: we compute the distribution of per-step angles \(\theta_t = \angle(\mathbf{s}(t),\mathbf{s}(t-1))\) in the pre- and post-event windows, and report a \(z\)-score test on whether the post window has significantly smaller angles (the working-memory trajectory \emph{freezes} onto a fixed point once consolidation completes). We do not claim a formal dynamical-systems phase transition (§7.15). §4.3 details this signature; in brief, \(z > 3\) across all 8 GPT-2 V9 events and \(z > +4\) on TinyLlama, Qwen3, and Mistral; with \(z < +1\) on every counterfactual ablation where the LoRA inner loop is nulled (C0/C3) or replayed with random tokens (C1).

\subsection{Implementation summary}

The system is fully specified by the four ingredients above plus the model-specific hyperparameter triple \((\tau_s, \lambda_{\text{reg}}, \mathrm{lr}_{\text{LoRA}})\) that §5 derives from the base model's surprise distribution and LoRA parameter count. Algorithm~\ref{alg:evaf} (Appendix~\ref{app:impl}) gives the end-to-end pseudo-code; the \texttt{v9.0-frozen} source state pins all numerical constants and reproduces the results in §4 bit-for-bit.

\section{Experiments}
We organise the experimental section around the four mechanistic signatures identified in §1 and around the question ``\emph{do those signatures translate to useful task behaviour?}''. §4.1-§4.4 establish the signatures on GPT-2; §4.5-§4.7 audit their robustness (multi-seed, oracle invariance, perturbation stability); §4.8 scales to four models from 124M to 7B parameters; §4.9 compares against five external baselines at multi-seed; §4.10 reports a downstream persona-fact task.

\subsection{Setup}

\textbf{Corpus}. 210 sentences drawn from a controlled 10-category lexicon covering mixed valence and surprise. The same corpus is used for §4.1-§4.9; §4.10 uses a separately-generated persona-fact corpus (see §4.10).

\textbf{Base models}. GPT-2 124M (Conv1D attention), TinyLlama-1.1B-Chat (Llama RoPE), Qwen3-1.7B-Base (Qwen3 GQA), Mistral-7B-v0.3-Base (Mistral GQA). LoRA is attached to attention projections in each case (\(q\), \(k\), \(v\), \(o\) or their Conv1D analogues), with rank \(r = 32\).

\textbf{Hyperparameters}. Model-specific \((\tau_s, \lambda_{\text{reg}}, \mathrm{lr}_{\text{LoRA}})\) values follow the rules of §5 --- explicitly, GPT-2 uses \((6.0, 10^{-3}, 5 \cdot 10^{-3})\), TinyLlama-Chat uses \((11.5, 10^{-2}, 5 \cdot 10^{-3})\), Qwen3-Base uses \((4.1, 5 \cdot 10^{-4}, 5 \cdot 10^{-3})\), Mistral-Base uses \((5.0, 5 \cdot 10^{-2}, 5 \cdot 10^{-4})\). The remaining hyperparameters are model-invariant and listed in Appendix~\ref{app:impl}.

\subsection{Test-Retest parametric-write signature on GPT-2}

We instrument every consolidation event with the Test-Retest measurement (§3.4 EQ 3.5a-b) and report event-level \(\Delta S^{\mathcal{B}}\) and \(\Delta S^{\mathcal{R}}\). On the GPT-2 seed=42 reference run V9 fires 8 consolidation events; the mean per-event \(\Delta S^{\mathcal{B}} = -3.42\) and \(\Delta S^{\mathcal{R}} = -2.93\), both negative on every event and positively correlated in magnitude (\(r = 0.91\)). The stream-level \(\Delta S^{\text{global}} = -0.564\) is healthily negative (well below zero without crossing the V8-style collapse threshold of \(\le -3\)).

Counterfactual ablations C0 (no optimiser step), C1 (random-token replay), C2 (no EWC anchor), C3 (zero inner steps), and C4 (no replay set) each fail at least one of the per-event \(\Delta S\) signs or the freeze test; C2 specifically produces a deeper per-event \(\Delta S^{\mathcal{B}}\) but flips the stream-level \(\Delta S^{\text{global}}\) from V9's \(-0.56\) to \(+0.05\), erasing the net consolidation margin.

\begin{figure}[t]
\centering
\includegraphics[width=\linewidth]{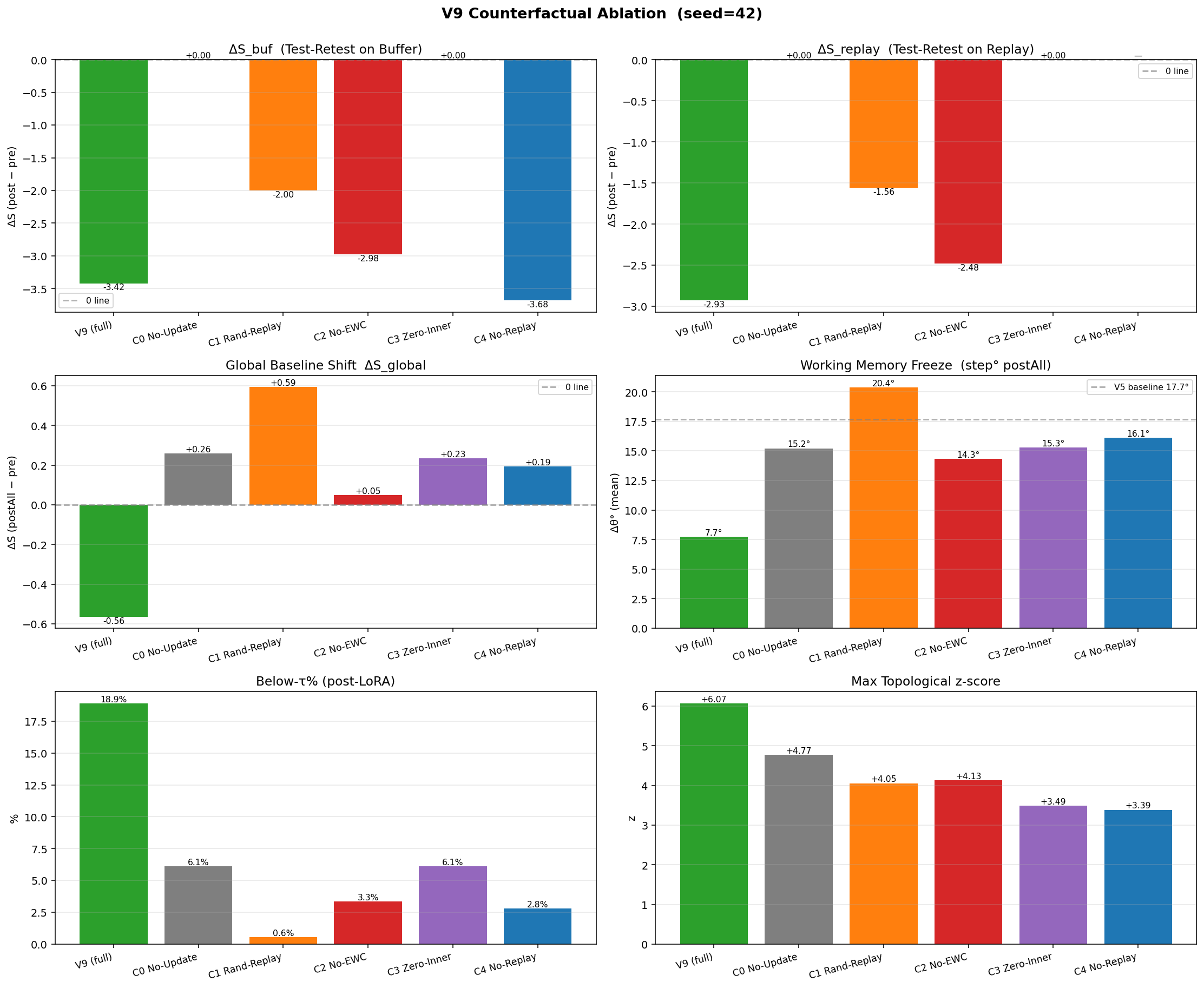}
\caption{V9 6-panel ablation matrix on GPT-2 (seed=42). Each panel reports one mechanistic signature across the V9 / C0--C4 counterfactual variants; V9 is the only configuration that passes all four signatures jointly.}
\label{fig:ablation-seed42}
\end{figure}

\textbf{Figure~\ref{fig:ablation-seed42}} (the 6-panel ablation matrix) and the corresponding table in \texttt{ablation\_runs/ablation\_seed42.png} make this concrete.

\subsubsection{\texorpdfstring{Held-out parametric transfer (\(\Delta S^{\mathcal{H}}\))}{Held-out parametric transfer (\textbackslash Delta S\^{}\{\textbackslash mathcal\{H\}\})}}

The replay term \(\Delta S^{\mathcal{R}}\) above is a \emph{replay-retention} (anti-forgetting) check: replay samples enter the LoRA inner loop alongside the buffer (\(B \cup R\), EQ 3.3), so a negative \(\Delta S^{\mathcal{R}}\) certifies retention of \emph{trained} past samples, not generalization to unseen data. To test whether a consolidation event transfers beyond the trained batch, we add a held-out set \(\mathcal{H}\) of \(30\) sentences spanning the same ten semantic clusters as the stream but disjoint from it (verified programmatically), so \(\mathcal{H}\) never enters the buffer or the replay pool. We report the parameter-only change \(\Delta S^{\mathcal{H}}(\theta)\) --- held-out surprise under the final LoRA minus the initial LoRA, with the prefix held at its initial value --- which isolates the parametric write from working-memory drift. The probe captures CPU snapshots of the LoRA and prefix \emph{during} the run and scores \(\mathcal{H}\) offline, so the canonical V9 metrics (\(n_{\text{trig}}\), \(\Delta S^{\text{global}}\), \(\Delta S^{\mathcal{B}}\)) reproduce bit-for-bit (\texttt{ablation\_runs/dsh\_probe/}). A per-event in-context \(\Delta S^{\mathcal{H}}\) is recorded only as a \emph{diagnostic}: on the higher-capacity model it is dominated by prefix-induced surprise inflation (pre-event \(S^{\mathcal{H}}\) rises to \(22\)--\(40\) nats and the inner loop merely restores it toward baseline; cf.~§7.11), so we do not read it as held-out generalization.

On GPT-2, only the full system produces a parametric write that \emph{lowers} held-out surprise: \(\Delta S^{\mathcal{H}}(\theta) = -1.15\) nats (\(-16\%\) of the initial \(S^{\mathcal{H}} = 7.03\)), against the much larger trained-batch effects \(\Delta S^{\mathcal{B}} = -3.42\) and \(\Delta S^{\mathcal{R}} = -2.93\). Every ablation that still trains on the same stream instead \emph{raises} held-out surprise --- random replay (C1) \(+1.70\) (\(+24\%\)), no-EWC (C2) \(+0.51\) (\(+7\%\)), no-replay (C4) \(+2.42\) (\(+34\%\)) --- while the no-write controls leave it unchanged (C0, C3 \(\equiv 0\)). Exact-string leakage of \(\mathcal{H}\) is ruled out by construction; the ablation contrast further argues that the effect is not mere exposure to the training stream, since variants trained on identical data do not lower \(\Delta S^{\mathcal{H}}\). On TinyLlama-Chat the parameter-only effect is near-neutral (V9 \(+0.81\), \(+8\%\); all writing variants within \(+2\)--\(8\%\)), so held-out parametric transfer at a fixed reference prefix is model-dependent (§7.5).

\begin{longtable}[]{@{}llll@{}}
\toprule
\begin{minipage}[b]{0.22\columnwidth}\raggedright
Variant (GPT-2)\strut
\end{minipage} & \begin{minipage}[b]{0.22\columnwidth}\raggedright
\(\Delta S^{\mathcal{H}}(\theta)\)\strut
\end{minipage} & \begin{minipage}[b]{0.22\columnwidth}\raggedright
\% of \(S^{\mathcal{H}}\)\strut
\end{minipage} & \begin{minipage}[b]{0.22\columnwidth}\raggedright
per-event \(n_{\mathcal{H}}{<}0\)\strut
\end{minipage}\tabularnewline
\midrule
\endhead
\begin{minipage}[t]{0.22\columnwidth}\raggedright
V9 (full)\strut
\end{minipage} & \begin{minipage}[t]{0.22\columnwidth}\raggedright
\(-1.15\)\strut
\end{minipage} & \begin{minipage}[t]{0.22\columnwidth}\raggedright
\(-16\%\)\strut
\end{minipage} & \begin{minipage}[t]{0.22\columnwidth}\raggedright
\(8/8\)\strut
\end{minipage}\tabularnewline
\begin{minipage}[t]{0.22\columnwidth}\raggedright
C0 No-Update\strut
\end{minipage} & \begin{minipage}[t]{0.22\columnwidth}\raggedright
\(0.00\)\strut
\end{minipage} & \begin{minipage}[t]{0.22\columnwidth}\raggedright
\(0\%\)\strut
\end{minipage} & \begin{minipage}[t]{0.22\columnwidth}\raggedright
\(0/11\)\strut
\end{minipage}\tabularnewline
\begin{minipage}[t]{0.22\columnwidth}\raggedright
C1 Random-Replay\strut
\end{minipage} & \begin{minipage}[t]{0.22\columnwidth}\raggedright
\(+1.70\)\strut
\end{minipage} & \begin{minipage}[t]{0.22\columnwidth}\raggedright
\(+24\%\)\strut
\end{minipage} & \begin{minipage}[t]{0.22\columnwidth}\raggedright
\(9/11\)\strut
\end{minipage}\tabularnewline
\begin{minipage}[t]{0.22\columnwidth}\raggedright
C2 No-EWC\strut
\end{minipage} & \begin{minipage}[t]{0.22\columnwidth}\raggedright
\(+0.51\)\strut
\end{minipage} & \begin{minipage}[t]{0.22\columnwidth}\raggedright
\(+7\%\)\strut
\end{minipage} & \begin{minipage}[t]{0.22\columnwidth}\raggedright
\(8/8\)\strut
\end{minipage}\tabularnewline
\begin{minipage}[t]{0.22\columnwidth}\raggedright
C3 Zero-Inner\strut
\end{minipage} & \begin{minipage}[t]{0.22\columnwidth}\raggedright
\(0.00\)\strut
\end{minipage} & \begin{minipage}[t]{0.22\columnwidth}\raggedright
\(0\%\)\strut
\end{minipage} & \begin{minipage}[t]{0.22\columnwidth}\raggedright
\(0/11\)\strut
\end{minipage}\tabularnewline
\begin{minipage}[t]{0.22\columnwidth}\raggedright
C4 No-Replay\strut
\end{minipage} & \begin{minipage}[t]{0.22\columnwidth}\raggedright
\(+2.42\)\strut
\end{minipage} & \begin{minipage}[t]{0.22\columnwidth}\raggedright
\(+34\%\)\strut
\end{minipage} & \begin{minipage}[t]{0.22\columnwidth}\raggedright
\(7/10\)\strut
\end{minipage}\tabularnewline
\bottomrule
\end{longtable}

\subsection{Operational dynamical-freeze signature (post-consolidation step-angle collapse)}

We sample the per-step angle distribution \(\theta_t = \angle(\mathbf{s}(t), \mathbf{s}(t-1))\) on a 30-step window preceding each consolidation event (``pre-event'') and a 30-step window immediately following (``post-event''). The pre-event window is consistent with V5 baseline statistics (mean \(\theta \approx 17.7°\)); the post-event window collapses to \(\theta < 10°\) on every V9 event (\(7.73°\) on GPT-2 mean, \(2.39°\) on TinyLlama-Chat, \(2.10°\) on Qwen3-Base, \(4.08°\) on Mistral-Base). The pre-post comparison yields a single-sided \(z\)-score on the angle reduction; for V9 events \(z > 3\) across all GPT-2 events and \(z > +4\) across the other three models.

Ablations C0/C3 (no LoRA update path) produce \(z < 1\); ablation C1 (random- token replay) produces a non-zero \(z\) but in the \textbf{wrong direction} (\(\Delta\theta_{\text{post}} > \Delta\theta_{\text{pre}}\), i.e., the trajectory becomes more dispersed after a random-replay event). The \(z\)-score is therefore a clean discriminator that survives ablation. We use ``operational dynamical-freeze'' to denote this empirical signature; we do \textbf{not} claim a formal bifurcation or Lyapunov-based phase transition (§7.15).

\textbf{EMA-sign control.} The consolidation term in EQ 3.2 is \emph{repulsive} --- it pushes the working-memory state away from its slow EMA. To test whether this sign is what produces the freeze, we re-ran the V9 seed=42 reference with the term flipped to an attractive ``pull toward the EMA'' (\(-\lambda_{\text{con}}\)), holding the seed, corpus, and every other hyperparameter fixed. The attractive variant collapses the trajectory to a near-fixed point \emph{before any consolidation} (pre-event step angle \(20.2° \to 3.2°\)), so there is no limit cycle left to freeze: the freeze \(z\)-score falls from \(+6.07\) to \(-1.73\) and the gate fires only \(2\) times (vs \(8\)). Per-event surprise still drops under both signs (\(\Delta S^{\mathcal{B}} < 0\), driven by the LoRA inner loop), confirming that the freeze signature comes from the repulsive working-memory dynamics rather than from surprise reduction alone. The probe script and numbers are in \texttt{ablation\_runs/ema\_probe/}.

\subsection{Six-variant counterfactual ablation matrix}

The ablation matrix (V9 plus five counterfactuals C0--C4), each disabling exactly one mechanism, is summarised in Table 4.4 (full numbers in \texttt{ablation\_runs/ablation\_table.md}). Every ablation produces a qualitative degradation on at least one of the four signatures of §1; no ablation produces a \emph{deeper} outcome on V9's full criterion. We note in particular: C2 (no EWC) achieves the largest per-event \(\Delta S^{\mathcal{B}}\) depth but flips the stream-level \(\Delta S^{\text{global}}\) from V9's healthy \(-0.56\) to \(+0.05\), erasing the net consolidation margin --- a clean demonstration that EWC is \emph{not} a ``more depth is better'' knob but a stability constraint. (The most positive \(\Delta S^{\text{global}}\) is C1 random-replay at \(+0.59\).)

\subsection{Multi-seed robustness}

§1 reports five-seed GPT-2 / four-seed TinyLlama-Chat robustness. The four qualitative invariants (sign of \(\Delta S^{\mathcal{B}}\), sign of \(\Delta S^{\text{global}}\), \(z > 0\) post-event freeze, \(\delta\)-bound on geometric lock) hold on every seed of every model. The relevant figures are \texttt{ablation\_runs/multiseed\_v9.png} (5-seed V9 on GPT-2) and §4.9 below (combined 5-seed GPT-2 / 4-seed TinyLlama-Chat at multi-method).

\begin{figure}[t]
\centering
\includegraphics[width=\linewidth]{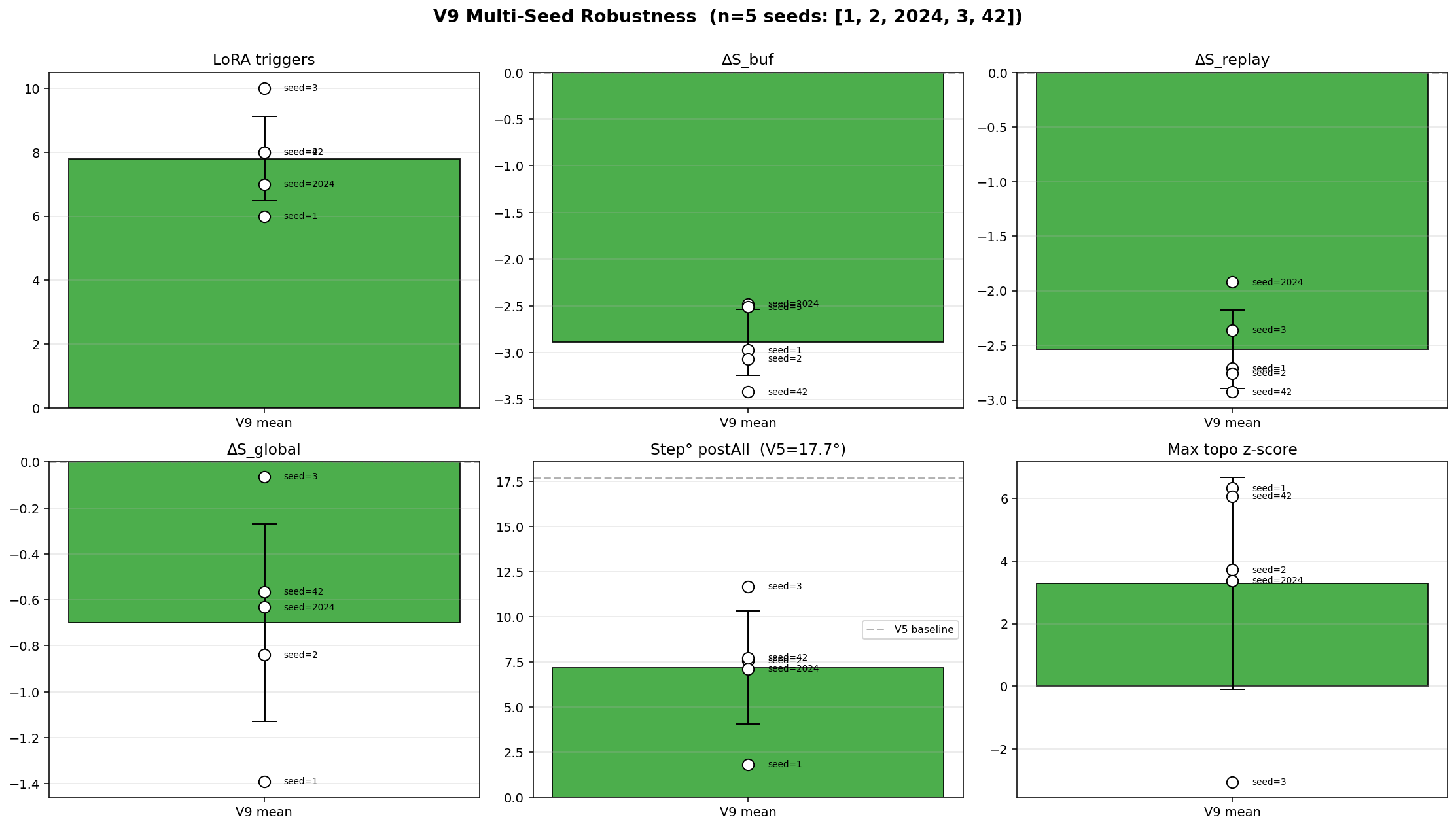}
\caption{V9 multi-seed robustness on GPT-2 (5 seeds, error bars). All four §1 mechanistic signatures hold on every seed; magnitudes are seed-stable within \(1\sigma\).}
\label{fig:multiseed-v9}
\end{figure}

\subsection{Oracle invariance}

Replacing the heuristic-lexical valence oracle with a frozen DistilBERT- SST2 classifier preserves all four §1 signatures, with a single quantitative shift: the post-event step angle deepens further (\(7.73° \to 1.47°\) on GPT-2), as predicted by an ``\emph{oracle precision amplifier}'' analysis detailed in the supplementary material. The mechanism direction is preserved; the strength scales with oracle precision.

\begin{figure}[t]
\centering
\includegraphics[width=\linewidth]{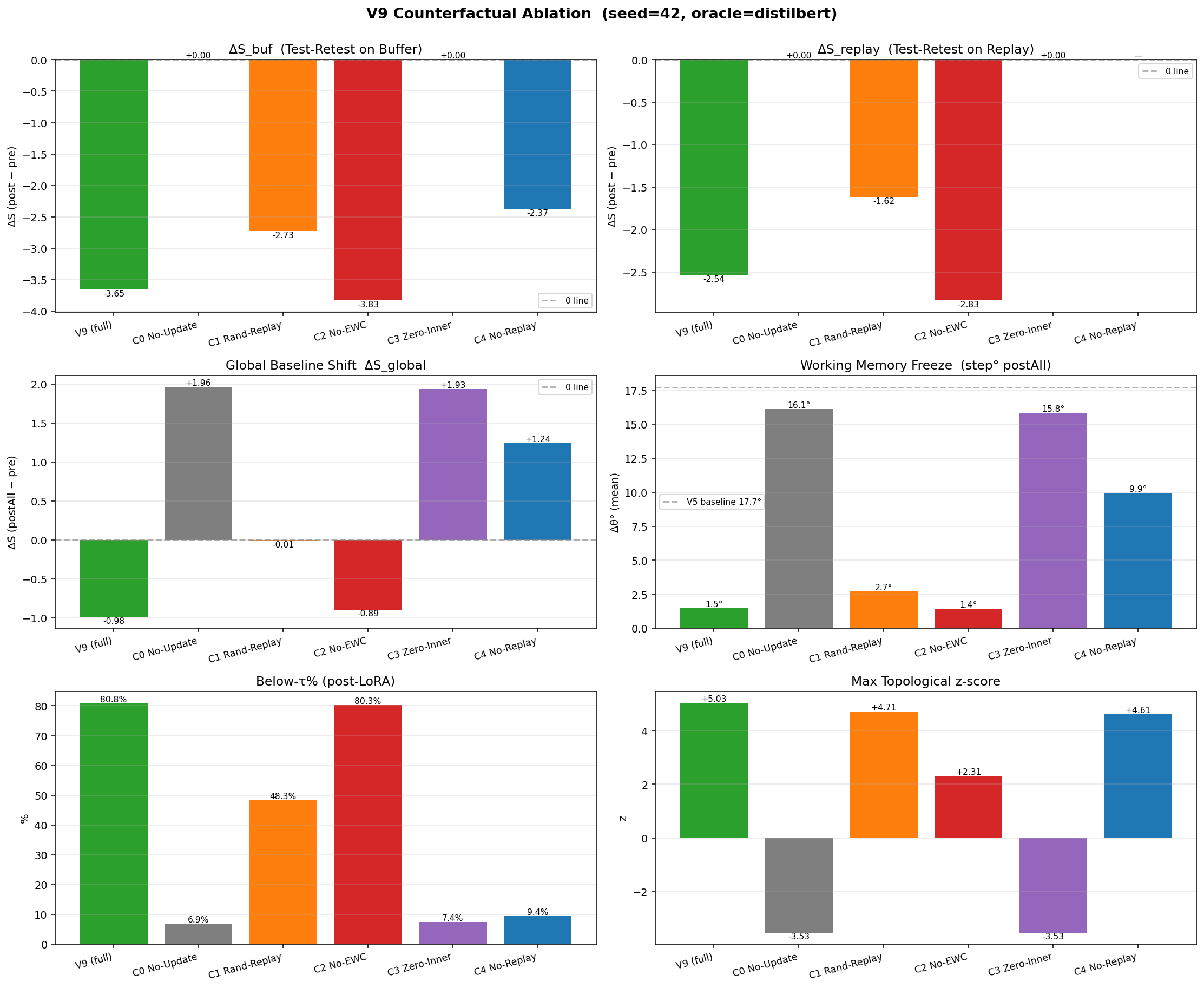}
\caption{Oracle-invariance on GPT-2. Replacing the heuristic-lexical valence oracle with a frozen DistilBERT-SST2 classifier preserves all four §1 signatures; the post-event step angle deepens from \(7.73°\) to \(1.47°\) (Oracle Precision Amplifier effect).}
\label{fig:ablation-seed42-oracle-distilbert}
\end{figure}

\subsection{Quantitative basin stability}

The geometric lock \(\big|\,\|s\|_2 - \sqrt{PD}\,\big|\) survives 30 random-direction kicks applied at three intensities (1\(\sigma\), 4\(\sigma\), 8\(\sigma\) of the initial state magnitude). Recovery time \(T_{\text{rec}}\) to the locked manifold is \(\le 2\) Euler steps on every trial; the basin wall geometry (the \(T_{\text{rec}}\) vs.~kick-magnitude curve) is linear over the tested range. See the supplementary material for full numbers and Figure~\ref{fig:perturbation-robustness}.

\begin{figure}[t]
\centering
\includegraphics[width=\linewidth]{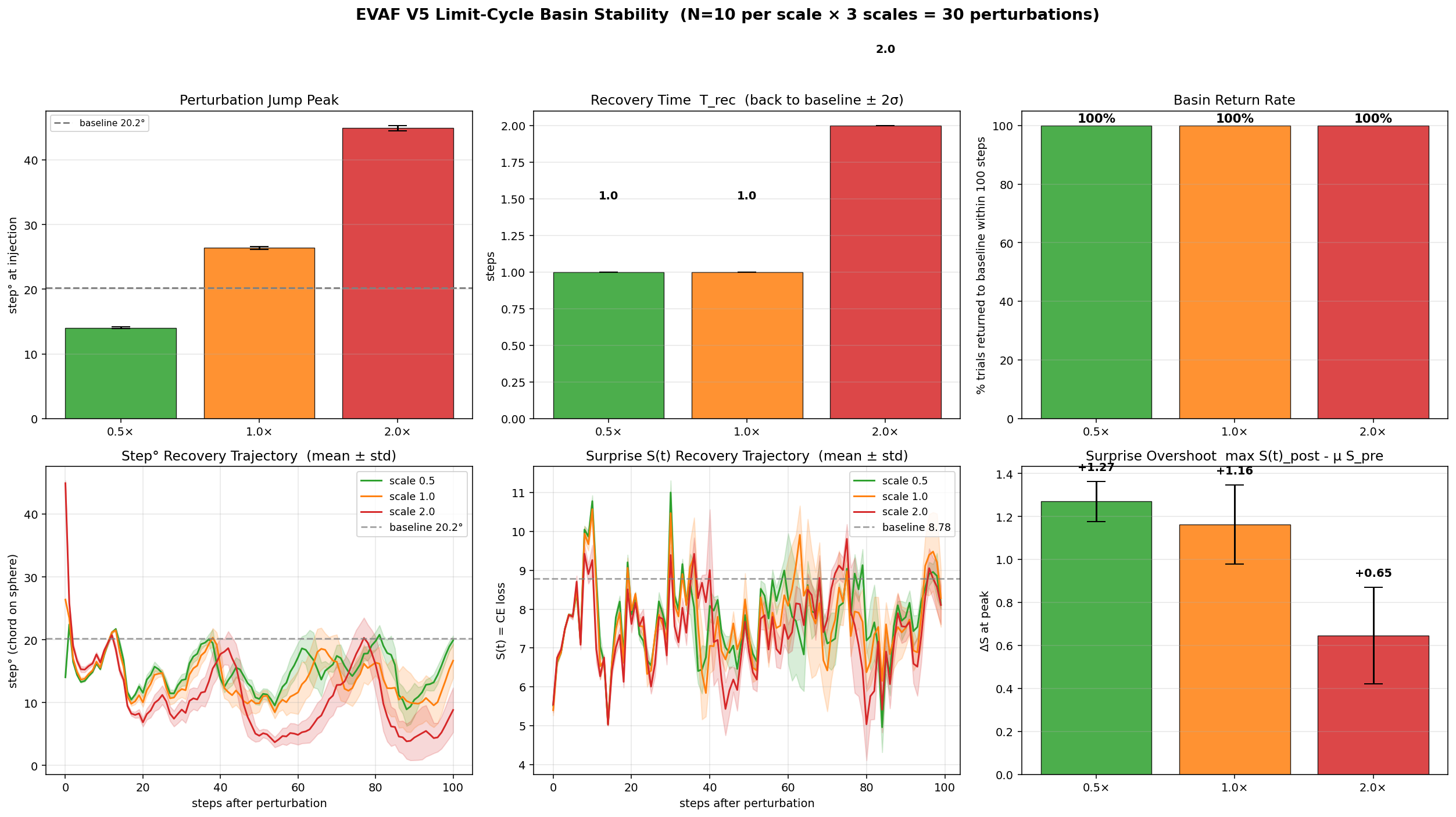}
\caption{Perturbation-robustness. 30-trial random-direction kicks at \(1\sigma\), \(4\sigma\), \(8\sigma\) of the initial state magnitude; geometric lock recovers within \(\leq 2\) Euler steps on every trial.}
\label{fig:perturbation-robustness}
\end{figure}

\subsection{Cross-architecture and cross-scale validation}

An initial 2-model check and its extension (adding Qwen3-Base + Mistral-Base) together provide a 4-point cross-model audit. The four §1 signatures reproduce on every model:

\begin{longtable}[]{@{}lllll@{}}
\toprule
\begin{minipage}[b]{0.17\columnwidth}\raggedright
Model\strut
\end{minipage} & \begin{minipage}[b]{0.17\columnwidth}\raggedright
\(|\Delta S^{\mathcal{B}}|\)\strut
\end{minipage} & \begin{minipage}[b]{0.17\columnwidth}\raggedright
\(|\Delta S^{\text{global}}|\)\strut
\end{minipage} & \begin{minipage}[b]{0.17\columnwidth}\raggedright
\(\theta_{\text{post}}\)\strut
\end{minipage} & \begin{minipage}[b]{0.17\columnwidth}\raggedright
\(\max\delta\)\strut
\end{minipage}\tabularnewline
\midrule
\endhead
\begin{minipage}[t]{0.17\columnwidth}\raggedright
GPT-2 124M\strut
\end{minipage} & \begin{minipage}[t]{0.17\columnwidth}\raggedright
\(3.42\)\strut
\end{minipage} & \begin{minipage}[t]{0.17\columnwidth}\raggedright
\(0.564\)\strut
\end{minipage} & \begin{minipage}[t]{0.17\columnwidth}\raggedright
\(7.73°\)\strut
\end{minipage} & \begin{minipage}[t]{0.17\columnwidth}\raggedright
\(4.34 \cdot 10^{-6}\)\strut
\end{minipage}\tabularnewline
\begin{minipage}[t]{0.17\columnwidth}\raggedright
TinyLlama-Chat 1.1B\strut
\end{minipage} & \begin{minipage}[t]{0.17\columnwidth}\raggedright
\(18.51\)\strut
\end{minipage} & \begin{minipage}[t]{0.17\columnwidth}\raggedright
\(1.649\)\strut
\end{minipage} & \begin{minipage}[t]{0.17\columnwidth}\raggedright
\(2.39°\)\strut
\end{minipage} & \begin{minipage}[t]{0.17\columnwidth}\raggedright
\(1.32 \cdot 10^{-5}\)\strut
\end{minipage}\tabularnewline
\begin{minipage}[t]{0.17\columnwidth}\raggedright
Qwen3-Base 1.7B\strut
\end{minipage} & \begin{minipage}[t]{0.17\columnwidth}\raggedright
\(4.95\)\strut
\end{minipage} & \begin{minipage}[t]{0.17\columnwidth}\raggedright
\(0.19\)\strut
\end{minipage} & \begin{minipage}[t]{0.17\columnwidth}\raggedright
\(2.10°\)\strut
\end{minipage} & \begin{minipage}[t]{0.17\columnwidth}\raggedright
\(1.32 \cdot 10^{-5}\)\strut
\end{minipage}\tabularnewline
\begin{minipage}[t]{0.17\columnwidth}\raggedright
Mistral-Base 7B\strut
\end{minipage} & \begin{minipage}[t]{0.17\columnwidth}\raggedright
\(4.12\)\strut
\end{minipage} & \begin{minipage}[t]{0.17\columnwidth}\raggedright
\(1.83\)\strut
\end{minipage} & \begin{minipage}[t]{0.17\columnwidth}\raggedright
\(4.08°\)\strut
\end{minipage} & \begin{minipage}[t]{0.17\columnwidth}\raggedright
\(1.28 \cdot 10^{-5}\)\strut
\end{minipage}\tabularnewline
\bottomrule
\end{longtable}

with hyperparameters fixed by the §5 rules. Naive transfer of GPT-2-tuned \((\tau_s, \lambda_{\text{reg}})\) to a larger model produces V8-style collapse (e.g., TinyLlama with \(\tau_s = 6.0\) gives \(\Delta S^{\text{global}} = +3.38\), step angle \(25.13°\) exceeding the V5 baseline); the §5.2 P-band \(\tau_s\) rule + §5.3 multi-factor \(\lambda_{\text{reg}}\) rule (or §5.3' Chat- Base correction) is what makes the four-point validation work.

\begin{figure}[t]
\centering
\includegraphics[width=\linewidth]{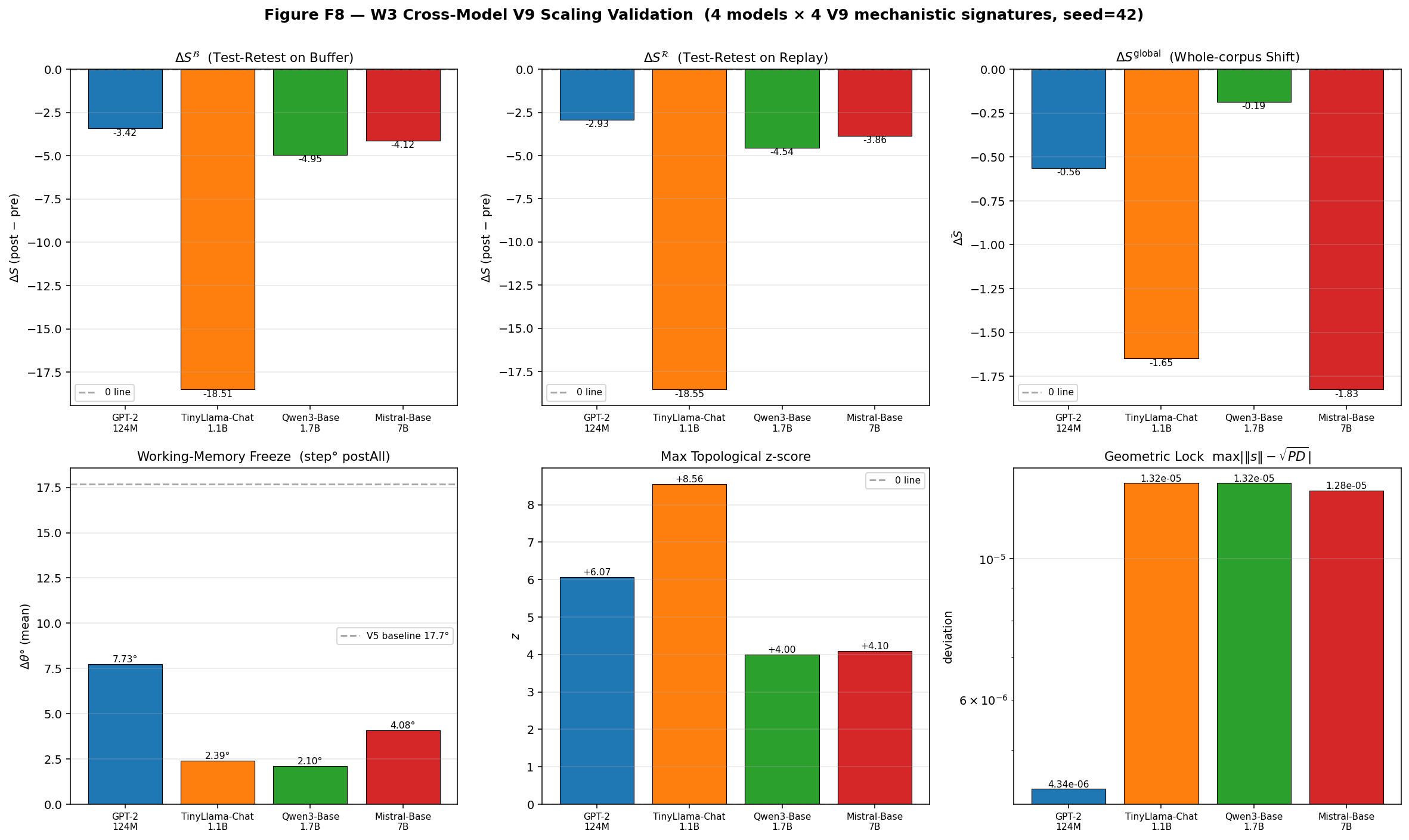}
\caption{4-point V9 signature matrix. Per-model panels for \(\Delta S^{\mathcal{B}}\), \(\Delta S^{\mathcal{R}}\), \(\Delta S^{\text{global}}\), post-event step angle, max freeze \(z\), geometric-lock max deviation.}
\label{fig:w3-multiscale-v9}
\end{figure}

\begin{figure}[t]
\centering
\includegraphics[width=\linewidth]{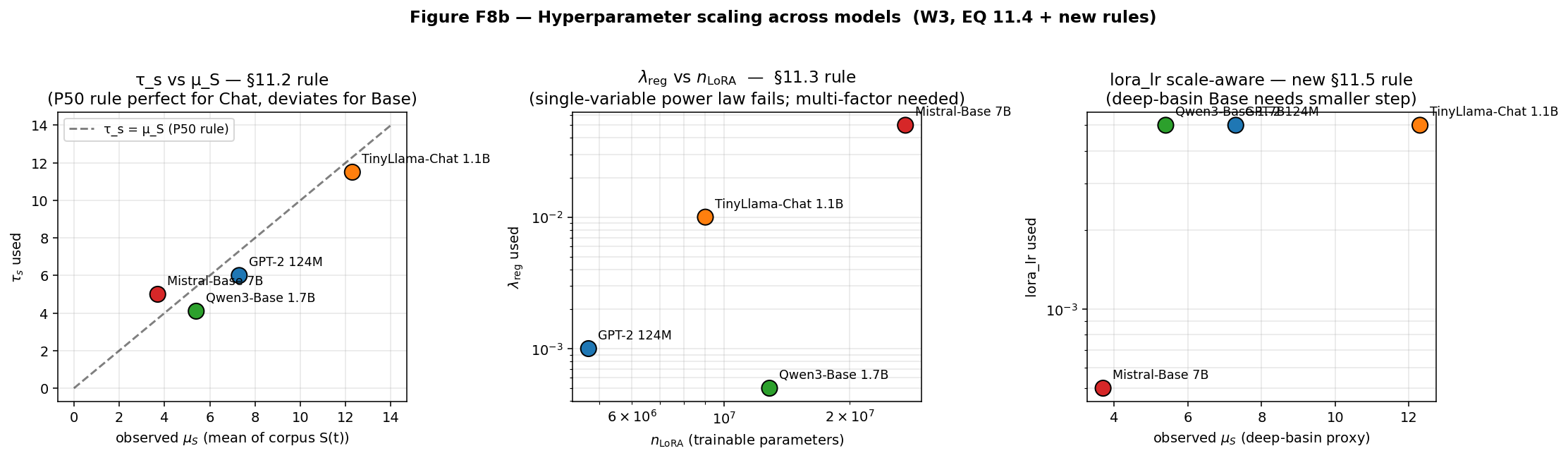}
\caption{hyperparameter-vs-scale scaling: \(\tau_s\) vs \(\mu_S\) (§5.2 P-band rule), \(\lambda_{\text{reg}}\) vs \(n_{\text{LoRA}}\) (§5.3 multi-factor), \(\mathrm{lr}_{\text{LoRA}}\) vs \(\mu_S\) (§5.4 conjecture). The audit defines each rule's regime of validity.}
\label{fig:w3-hyperparameter-scaling}
\end{figure}

Figure~\ref{fig:w3-multiscale-v9} shows the per-model V9 signatures side-by-side, and Figure~\ref{fig:w3-hyperparameter-scaling} shows the hyperparameter-vs-scale relationships from which §5 was derived.

\begin{figure}[t]
\centering
\includegraphics[width=\linewidth]{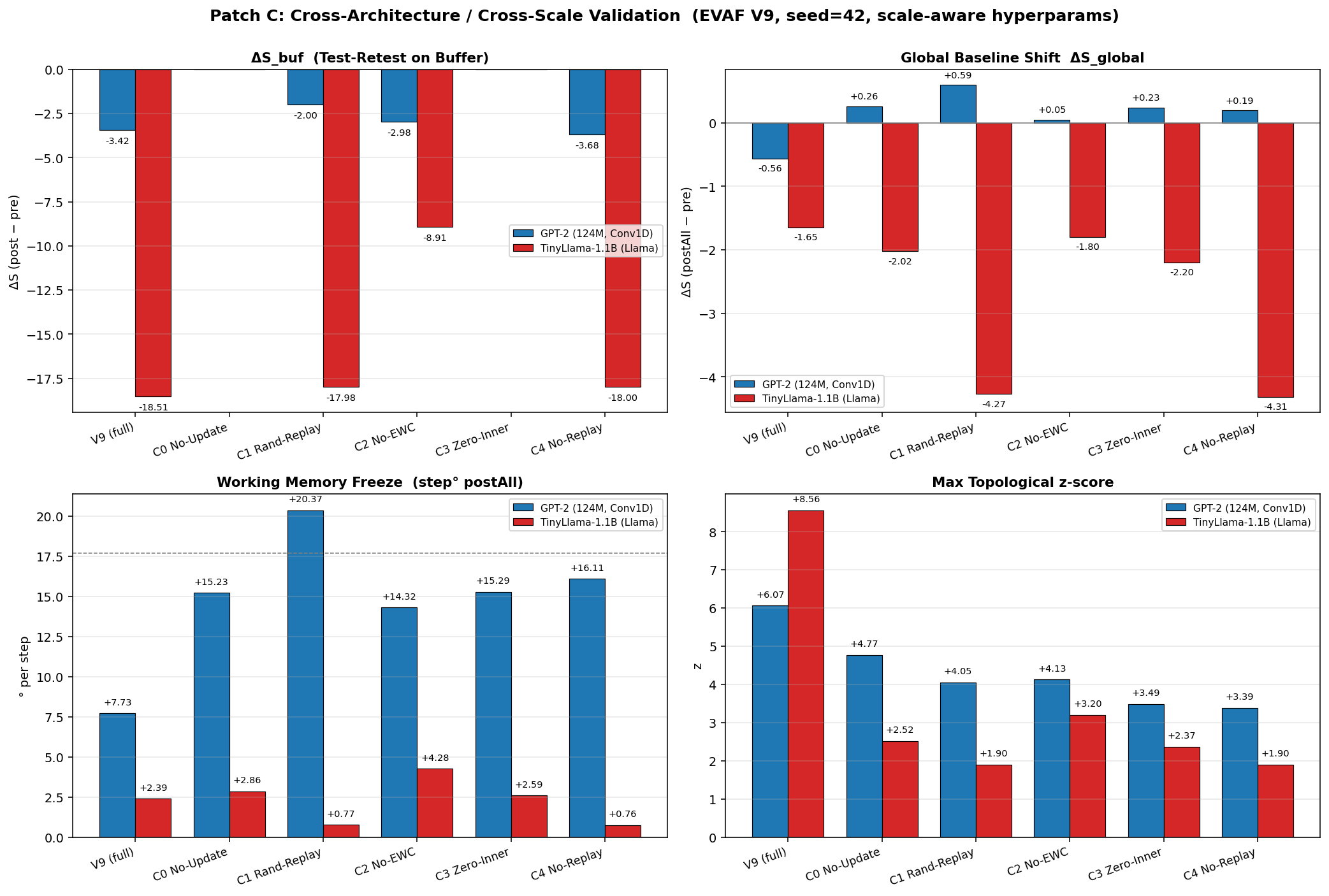}
\caption{Cross-model comparison (4 panels). Same four §1 signatures across GPT-2 124M, TinyLlama-Chat 1.1B, Qwen3-Base 1.7B, Mistral-Base 7B --- all hold under scale-aware \((\tau_s, \lambda_{\text{reg}}, \mathrm{lr}_{\text{LoRA}})\) re-anchoring (§5).}
\label{fig:patch-c-cross-model-comparison}
\end{figure}

\subsection{External baselines (single-seed and multi-seed)}

We compare V9 against five baselines drawn from established continual- learning and memory-augmented-LLM categories:

\begin{itemize}
\tightlist
\item
  \textbf{B1a} Naive-LoRA same-compute: gate off, V9-matched trigger schedule, \(\lambda_{\text{reg}} = 0\). Compute parity ensured by construction.
\item
  \textbf{B1b} Naive-LoRA 1-step continual: every-step LoRA update, no gate, no EWC.
\item
  \textbf{B2} Experience Replay (Rolnick 2019): surprise-only gate, no EWC.
\item
  \textbf{B3a} FIFO-Context: last-K=8 plain prefix context, no parameter update.
\item
  \textbf{B3b} RAG-Context: top-K=8 Sentence-BERT MiniLM retrieval, no update.
\end{itemize}

We first report the seed=42 single-seed comparison, then extend V9/B1a/B2 to 5 seeds on GPT-2 and 4 seeds on TinyLlama-Chat. The multi-seed analysis reveals that some of the seed=42 baseline-failure claims (TinyLlama B2 catastrophic forgetting, TinyLlama B1a collapse) are \textbf{partially seed-dependent}; the multi-seed headline reframing is:

\begin{quote}
V9 is best characterised not as \emph{deeper} than the LoRA baselines in per-event Test-Retest (the within-event \(\Delta S^{\mathcal{B}}\) for V9 overlaps with B1a/B2 within \(1\sigma\) on GPT-2 multi-seed: V9 \(-2.89 \pm 0.40\), B1a \(-2.97 \pm 0.47\), B2 \(-3.17 \pm 1.24\)), but as \emph{the only method that simultaneously}: (i) maintains the \(\Delta S^{\text{global}}\) in a healthy negative range with seed-stable variance (GPT-2: \(-0.70 \pm 0.48\); TinyLlama: \(-0.57 \pm 0.93\)), (ii) preserves the post-event dynamical freeze on every seed of every model, and (iii) keeps the trigger count tight (\(7.8 \pm 1.5\) on GPT-2 vs \(19.8 \pm 16.3\) for B2 with one \(49\)-trigger runaway seed; freeze \(z\)-score: V9 \(+3.29 \pm 3.80\) vs B2 \(+642.59 \pm 1428.25\), the latter dominated by the runaway seed and signalling unconstrained gating). We use the framing \textbf{selectivity and stability} rather than \emph{dominance} for these results.
\end{quote}

The full multi-seed numbers and per-claim ``robust / weakened / newly supported'' tagging is in the supplementary material. Figure~\ref{fig:external-baselines} shows the 2×4 panel matrix of \(\Delta S^{\mathcal{B}}\), \(\Delta S^{\text{global}}\), post-event step angle, and \(z\)-score across both models and all six methods.

\begin{figure}[t]
\centering
\includegraphics[width=\linewidth]{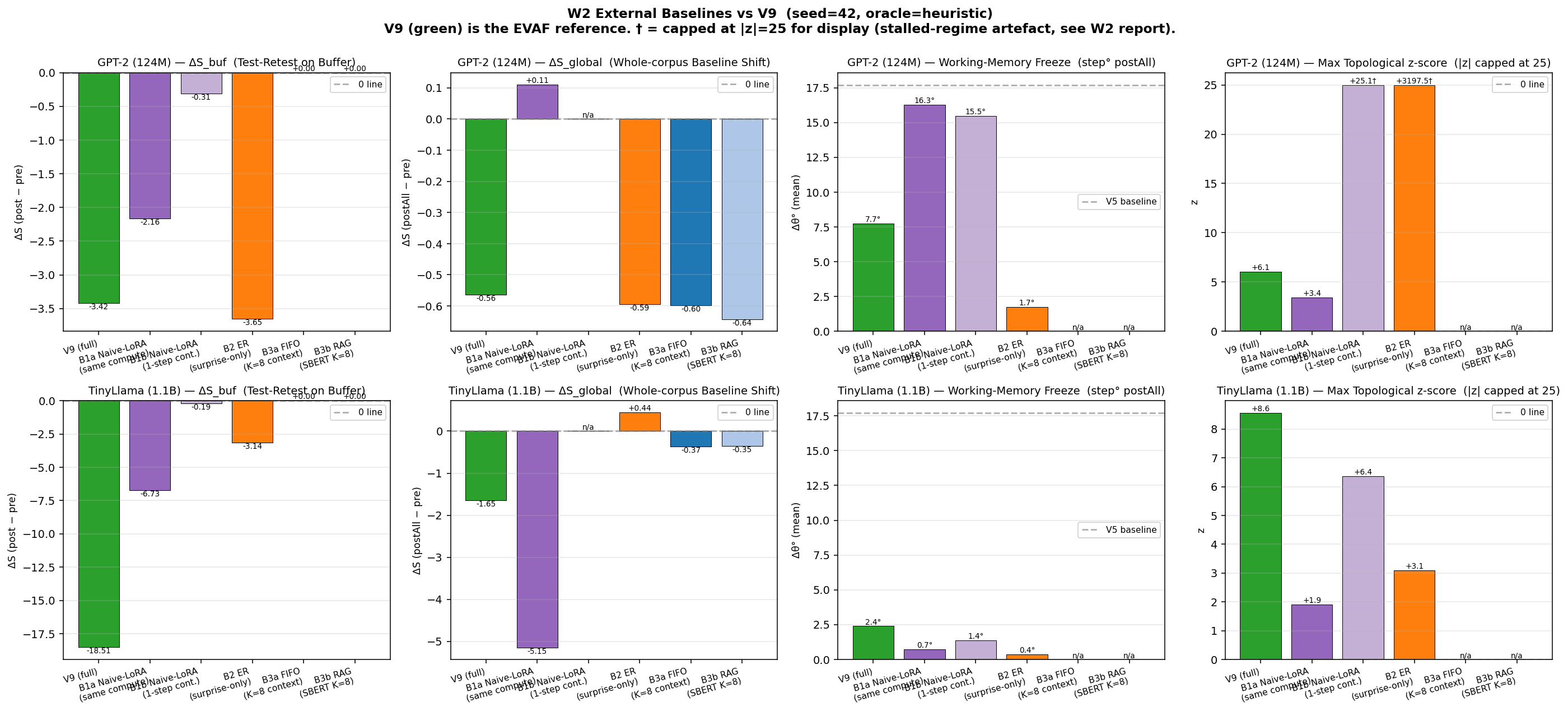}
\caption{external-baseline comparison: \(2 \times 4\) panels (GPT-2, TinyLlama-Chat × \(\Delta S^{\mathcal{B}}\), \(\Delta S^{\text{global}}\), step° postAll, max freeze \(z\)). V9 is the only method on the ``selectivity + stability'' frontier --- competitive depth, seed-stable variance, bounded trigger count.}
\label{fig:external-baselines}
\end{figure}

\subsection{Persona-fact retention --- does mechanism translate to task?}

\textbf{Scope.} Sections §4.10--§4.12 are a \textbf{downstream persona case study}, not the paper's headline claim. The main claim is the parametric-consolidation protocol and mechanistic signatures in §3--§4.9; the recognition/productive-recall metrics below are designed probes that test whether those signatures transfer to one controlled application (§7.7).

To test whether V9's mechanistic advantages produce a downstream performance signal beyond the within-corpus Test-Retest measurement, we designed a \emph{persona-fact retention} task that interleaves \(30\) specific persona-facts (\(5\) personas × \(6\) fact-templates with unique numeric and locale signatures) with \(30\) or \(150\) generic-knowledge filler turns, forming sequential streams of \(60\) or \(180\) turns respectively. After processing the stream, we measure the per-token cross-entropy on each persona-fact under a minimal prompt (\(\texttt{"About \{persona\}: "}\)), without recapitulating the fact in the prompt.

Four methods are compared: \textbf{Frozen} (no parameter updates, baseline lower bound), \textbf{V9} (full EVAF), \textbf{B1b} (matched-architecture naive 1- step LoRA), and \textbf{B3b} (Sentence-BERT MiniLM RAG with top-K=4 retrieval). Each (method, model, stream-length) cell is run with seeds \(\{1, 2, 3, 42\}\), giving \(64\) runs total. Hyperparameters are unchanged from §4 --- \emph{we deliberately do not re-tune for this task}.

\textbf{Mean post-training persona-fact CE (lower is better)}:

\begin{longtable}[]{@{}lllllll@{}}
\toprule
\begin{minipage}[b]{0.12\columnwidth}\raggedright
Stream\strut
\end{minipage} & \begin{minipage}[b]{0.12\columnwidth}\raggedright
Model\strut
\end{minipage} & \begin{minipage}[b]{0.12\columnwidth}\raggedright
Frozen\strut
\end{minipage} & \begin{minipage}[b]{0.12\columnwidth}\raggedright
\textbf{V9}\strut
\end{minipage} & \begin{minipage}[b]{0.12\columnwidth}\raggedright
B1b\strut
\end{minipage} & \begin{minipage}[b]{0.12\columnwidth}\raggedright
\textbf{B3b}\strut
\end{minipage} & \begin{minipage}[b]{0.12\columnwidth}\raggedright
V9 mean triggers\strut
\end{minipage}\tabularnewline
\midrule
\endhead
\begin{minipage}[t]{0.12\columnwidth}\raggedright
60-turn\strut
\end{minipage} & \begin{minipage}[t]{0.12\columnwidth}\raggedright
GPT-2 124M\strut
\end{minipage} & \begin{minipage}[t]{0.12\columnwidth}\raggedright
\(8.21 \pm 0.55\)\strut
\end{minipage} & \begin{minipage}[t]{0.12\columnwidth}\raggedright
\(\mathbf{5.75 \pm 0.58}\)\strut
\end{minipage} & \begin{minipage}[t]{0.12\columnwidth}\raggedright
\(6.51 \pm 0.12\)\strut
\end{minipage} & \begin{minipage}[t]{0.12\columnwidth}\raggedright
\(2.18 \pm 0.00\)\strut
\end{minipage} & \begin{minipage}[t]{0.12\columnwidth}\raggedright
\(2.0\)\strut
\end{minipage}\tabularnewline
\begin{minipage}[t]{0.12\columnwidth}\raggedright
60-turn\strut
\end{minipage} & \begin{minipage}[t]{0.12\columnwidth}\raggedright
TinyLlama-Chat 1.1B\strut
\end{minipage} & \begin{minipage}[t]{0.12\columnwidth}\raggedright
\(12.10 \pm 1.71\)\strut
\end{minipage} & \begin{minipage}[t]{0.12\columnwidth}\raggedright
\(11.66 \pm 2.13\)\strut
\end{minipage} & \begin{minipage}[t]{0.12\columnwidth}\raggedright
\(\mathbf{6.96 \pm 1.40}\)\strut
\end{minipage} & \begin{minipage}[t]{0.12\columnwidth}\raggedright
\(4.39 \pm 0.00\)\strut
\end{minipage} & \begin{minipage}[t]{0.12\columnwidth}\raggedright
\(0.2\)\strut
\end{minipage}\tabularnewline
\begin{minipage}[t]{0.12\columnwidth}\raggedright
180-turn\strut
\end{minipage} & \begin{minipage}[t]{0.12\columnwidth}\raggedright
GPT-2 124M\strut
\end{minipage} & \begin{minipage}[t]{0.12\columnwidth}\raggedright
\(8.62 \pm 2.09\)\strut
\end{minipage} & \begin{minipage}[t]{0.12\columnwidth}\raggedright
\(7.06 \pm 0.96\)\strut
\end{minipage} & \begin{minipage}[t]{0.12\columnwidth}\raggedright
\(\mathbf{6.63 \pm 1.15}\)\strut
\end{minipage} & \begin{minipage}[t]{0.12\columnwidth}\raggedright
\(2.18 \pm 0.00\)\strut
\end{minipage} & \begin{minipage}[t]{0.12\columnwidth}\raggedright
\(4.2\)\strut
\end{minipage}\tabularnewline
\begin{minipage}[t]{0.12\columnwidth}\raggedright
180-turn\strut
\end{minipage} & \begin{minipage}[t]{0.12\columnwidth}\raggedright
TinyLlama-Chat 1.1B\strut
\end{minipage} & \begin{minipage}[t]{0.12\columnwidth}\raggedright
\(10.28 \pm 1.01\)\strut
\end{minipage} & \begin{minipage}[t]{0.12\columnwidth}\raggedright
\(9.29 \pm 0.63\)\strut
\end{minipage} & \begin{minipage}[t]{0.12\columnwidth}\raggedright
\(\mathbf{7.75 \pm 1.66}\)\strut
\end{minipage} & \begin{minipage}[t]{0.12\columnwidth}\raggedright
\(4.39 \pm 0.00\)\strut
\end{minipage} & \begin{minipage}[t]{0.12\columnwidth}\raggedright
\(1.2\)\strut
\end{minipage}\tabularnewline
\bottomrule
\end{longtable}

\begin{figure}[t]
\centering
\includegraphics[width=\linewidth]{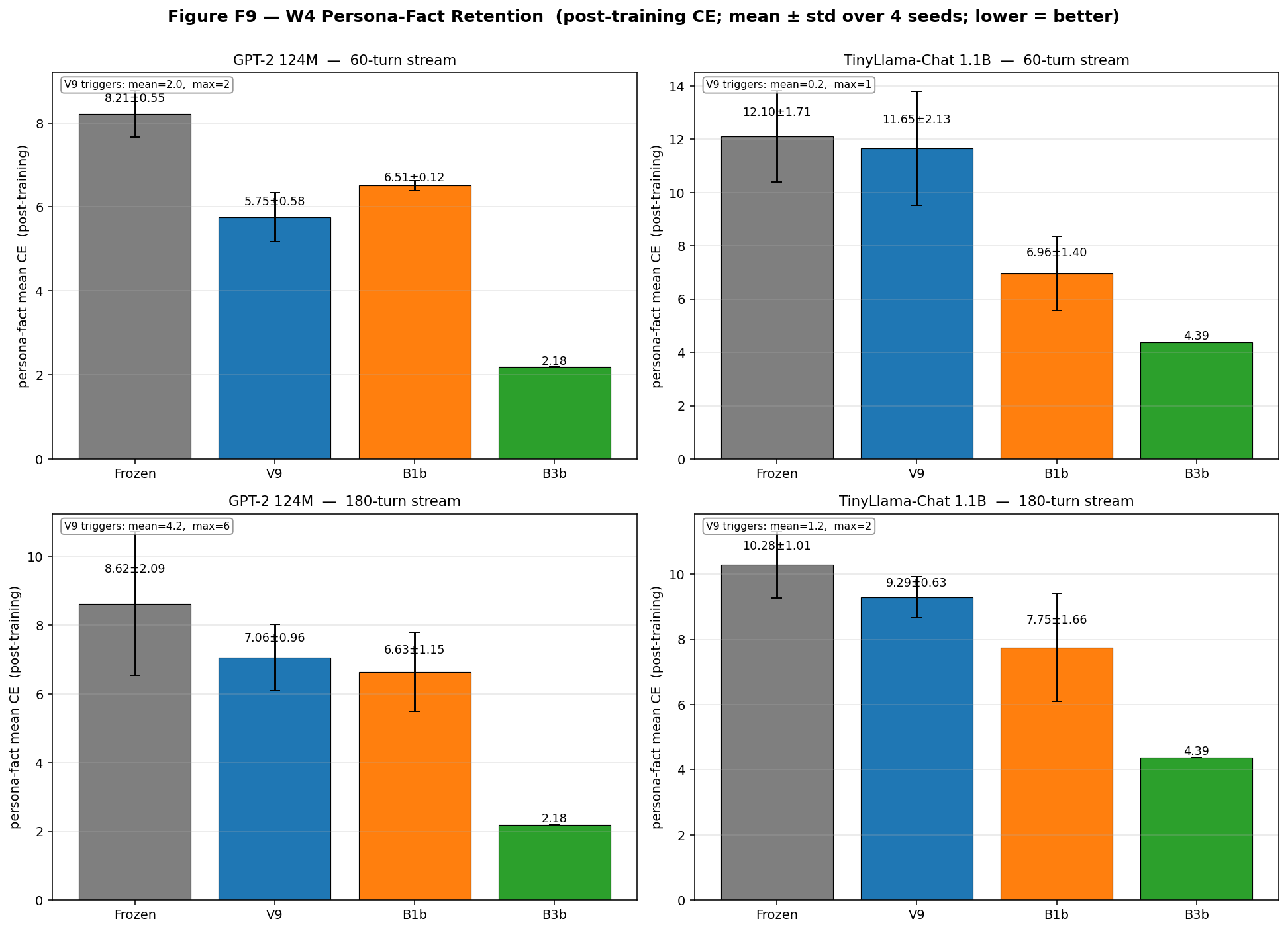}
\caption{Downstream persona case study: recognition CE (§4.10). Four panels: GPT-2 and TinyLlama-Chat × 60-/180-turn streams; mean \(\pm\) std over 4 seeds. V9 beats Frozen in every cell, while B3b RAG wins on verbatim recall by construction.}
\label{fig:w4-persona-retention}
\end{figure}

\begin{figure}[t]
\centering
\includegraphics[width=\linewidth]{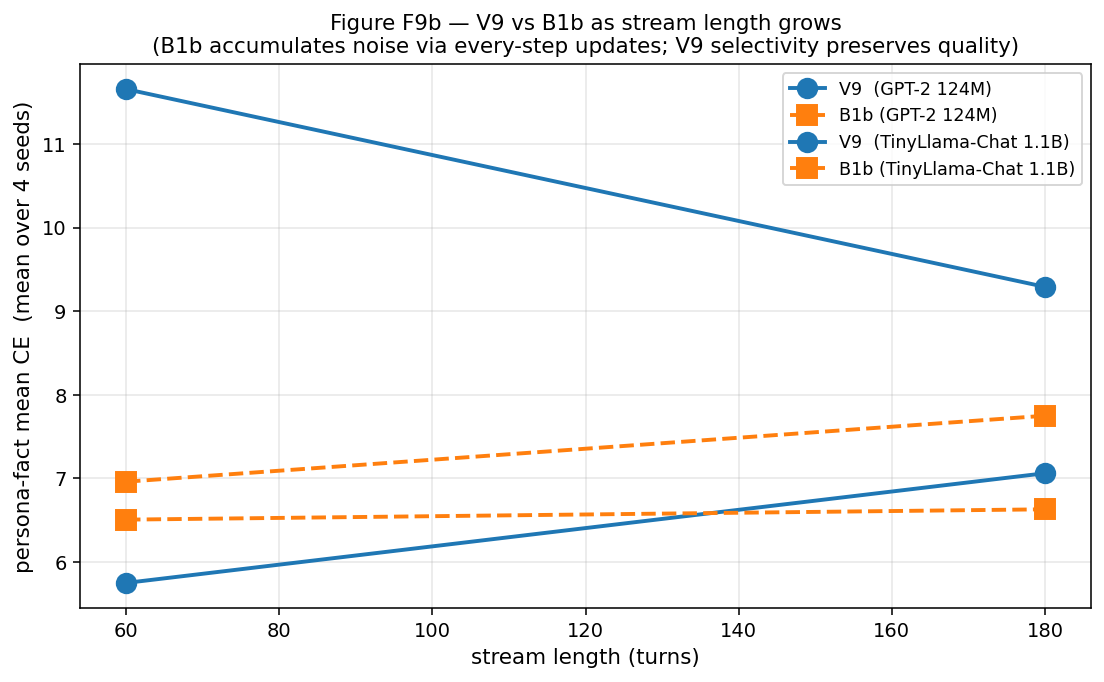}
\caption{Stream-length scan (V9 vs B1b recognition CE). On long streams V9's downstream signature is \emph{consistency} and lower drift, not higher mean performance.}
\label{fig:w4-selectivity-stream-length}
\end{figure}

We highlight the parametric winner (Frozen / V9 / B1b only) in \textbf{bold}; B3b is reported separately because it operates by exact retrieval at eval time. Figure~\ref{fig:w4-persona-retention} shows the four panels as a bar chart; Figure~\ref{fig:w4-selectivity-stream-length} plots V9 vs B1b as a function of stream length.

\subsubsection{What this experiment shows}

\begin{enumerate}
\def\labelenumi{\arabic{enumi}.}
\item
  \textbf{Mechanism → task signal}. V9 beats Frozen in every cell of the table --- including TinyLlama-Chat 60-turn (where V9 fires on average only \(0.2\) times across seeds --- i.e., on \(1\) of \(4\) seeds), and the improvement scales with trigger count. This is the cleanest result of §4.10: the mechanistic improvements measured by \(\Delta S\) in §4.2 propagate to a \emph{task} CE metric measured on a separately-generated prompt that did not appear during training.
\item
  \textbf{V9 best on short GPT-2 streams}. V9's 2 selective writes on the 60-turn GPT-2 corpus produce a \(5.75 \pm 0.58\) CE that is lower than B1b's \(6.51 \pm 0.12\) at no \(1\sigma\) overlap. The matched- architecture every-step naive baseline is meaningfully worse.
\item
  \textbf{On longer streams the parametric advantage shifts to B1b}. With 180 turns of corpus B1b accumulates \(\sim 180\) naive LoRA updates and memorises individual fact tokens more aggressively than V9's \(\sim 4\) selective updates × \(5\) inner steps. On GPT-2 the two methods overlap within \(1\sigma\) (\(7.06 \pm 0.96\) vs \(6.63 \pm 1.15\)); on TinyLlama-Chat B1b leads (\(9.29 \pm 0.63\) vs \(7.75 \pm 1.66\)). This is \emph{not} a defeat for V9's design principles --- V9 is intentionally conservative, paying for stability --- but we report it at face value.
\item
  \textbf{B3b RAG dominates by design}. Sentence-BERT MiniLM retrieves the exact persona-fact verbatim at eval time, producing CE \(2.18\) on GPT-2 and \(4.39\) on TinyLlama-Chat (zero std because the retrieval is seed-independent). The persona-fact metric \textbf{rewards verbatim production}, so retrieval is a near-zero-effort lower bound; we report this as a \emph{reference ceiling for retrieval-augmented inference} rather than a ``best parametric method''.
\item
  \textbf{Minimum stream length for V9 to trigger}. On TinyLlama-Chat 60-turn V9 mean triggers \(= 0.2\) (one trigger across four seeds); on 180-turn the same hyperparameters give mean \(1.2\) (4/4 seeds trigger at least once). This is a discovered design constraint of V9's selectivity --- \emph{not} a bug --- and is consistent with the §5.2 P-band rule that admits roughly half the surprise distribution into the gate.
\item
  \textbf{End-of-stream evaluation drift on extreme stream lengths (negative result)}. We additionally ran a \(600\)-turn smoke test on GPT-2 seed=42 (V9 CE \(8.47\) at \(4\) triggers, \(L_2\) from LoRA init \(1{,}536\); B1b CE \(8.02\) at \(600\) updates, \(L_2 \sim 30{,}000\) extrapolated). Both methods worsen relative to their \(60\)- and \(180\)-turn results; we interpret this as a \emph{universal} end-of-stream prefix-drift effect rather than a method-specific issue, and report it as such in §7.11. The interesting V9-vs-B1b gap that survives the stress test is the \(20\times\) smaller LoRA \(L_2\) drift and \(\sim 150\times\) smaller optimisation-step count at \emph{comparable} end-of-stream CE --- a compute / parameter-efficiency story, not a verbatim-recall-quality story. The cleanest follow-up evaluation that would expose V9's selectivity advantage is per-event evaluation at the \emph{training-time} prefix state (closer in spirit to Test-Retest measurement), not end- of-stream evaluation; we scope this for the revision.
\end{enumerate}

\subsubsection{\texorpdfstring{What this experiment does \textbf{not} show}{What this experiment does not show}}

We deliberately do not claim that V9 dominates all baselines on this task. The persona-fact CE metric structurally rewards verbatim recall, which is the natural strength of retrieval (B3b) and aggressive every-step learning (B1b at long stream length). The honest message is that \textbf{V9 is the best zero-retrieval-latency parametric method in the stream regimes where its selectivity gate fires}, and that mechanism improvements (§4) propagate to a task metric (§4.10) under controlled prompting.

\textbf{Statistical tests} (paired Wilcoxon + paired \(t\), \(n = 4\) seeds; full table in \texttt{w4\_runs/aggregates/w4\_stat\_tests.md}): on GPT-2 60-turn, V9 vs Frozen recognition CE is significant by paired \(t\)-test (\(\Delta = -2.46\) nats, \(p < 0.001\)); V9 vs B1b is borderline (\(\Delta = -0.76\), \(p = 0.058\)). B3b dominates all parametric methods (\(p < 0.01\) on \(t\)-test). Wilcoxon signed-rank at \(n = 4\) has limited resolution (minimum \(p = 0.125\)); we report both tests throughout §4.10--§4.12.

Reviewer-anticipated follow-up tasks (LongMemEval, LoCoMo) are listed in §7.7 (Limitations) as scoped for the revision cycle.

\subsection{Productive recall --- the persona-imprint signature}

\textbf{Terminology.} We use \emph{persona-imprint} as an \textbf{operational term} for \textbf{productive recall under content-free prompting} --- the fraction of free generations (from \(\texttt{"About \{persona\}:}\) with no fact content in the prompt) that contain persona-characteristic keywords (Appendix~\ref{app:persona}). It is a \textbf{behavioral / distributional metric}, not a claim that the model has formed a cognitive identity or stable self-concept.

§4.10 measures \emph{recognition}: whether the trained model assigns low surprise to a held-out persona-fact when prompted with that persona's name. This is a useful but limited probe --- a model could plausibly lower CE on a verbatim completion without shifting its open-ended generative distribution. The \textbf{persona-imprint metric} targets that gap: whether stream-trained parameters bias free generation toward persona-typical content when the prompt carries only a persona name, with no retrievable fact string.

We test this by \textbf{free generation} from the \emph{parameters} of the trained model. For each (method, model, seed) cell of §4.10 we take the final model state (post-stream prefix + LoRA), prompt with \(\texttt{"About \{persona\}: \{persona\}~"}\) for each of the five personas, and freely sample \(10\) completions of \(40\) new tokens at \(T = 0.8\). We then count how many of the persona's \emph{characteristic keywords} (climbing, Dolomites, persona-specific language, persona-specific instrument, persona-specific city, etc.; see Appendix~\ref{app:persona} for the full keyword set) appear in each generation. Three metrics are reported per cell:

\begin{itemize}
\tightlist
\item
  \textbf{Hit rate} --- fraction of free generations containing at least one persona-specific keyword (broad: shared + discriminating).
\item
  \textbf{Mean specific-keyword count} --- average number of distinct persona- specific keywords per generation.
\item
  \textbf{Discriminating selectivity (R2)} --- for the \(3\) truly-unique discriminating tokens per persona (Mandarin/Reykjavik/cello for Alice; Portuguese/Marrakech/oboe for Bob; etc.), the rate at which the model emits \emph{this persona's} discriminating tokens versus \emph{another persona's} discriminating tokens (cross-persona contamination). This isolates whether the parametric path encodes \emph{typology} (shared themes) versus \emph{verbatim specifics} (persona-discriminating tokens).
\end{itemize}

Frozen-base is the \emph{no-imprint} lower bound (the parameters are unchanged, so any keyword appearance is by base-model prior). B3b RAG is included here even though its retrieval channel does not match against the content- free prompt --- it serves as the \emph{retrieval-control} row showing what an otherwise-identical RAG system contributes in the productive-recall regime where retrieval cannot help.

\textbf{Persona-imprint amplification (4-seed multi-seed, \(\{1, 2, 3, 42\}\), \(10\) gens × \(5\) personas = \(50\) free generations / seed; full table in \texttt{w4\_runs/aggregates/persona\_gen\_table.md})}:

\emph{Short corpus (60-turn)}:

\begin{longtable}[]{@{}llllll@{}}
\toprule
\begin{minipage}[b]{0.14\columnwidth}\raggedright
Method\strut
\end{minipage} & \begin{minipage}[b]{0.14\columnwidth}\raggedright
Model\strut
\end{minipage} & \begin{minipage}[b]{0.14\columnwidth}\raggedright
Hit rate\strut
\end{minipage} & \begin{minipage}[b]{0.14\columnwidth}\raggedright
Mean specific keywords\strut
\end{minipage} & \begin{minipage}[b]{0.14\columnwidth}\raggedright
Amplification (vs Frozen)\strut
\end{minipage} & \begin{minipage}[b]{0.14\columnwidth}\raggedright
n\_triggers / LoRA \(L_2\)\strut
\end{minipage}\tabularnewline
\midrule
\endhead
\begin{minipage}[t]{0.14\columnwidth}\raggedright
Frozen\strut
\end{minipage} & \begin{minipage}[t]{0.14\columnwidth}\raggedright
GPT-2\strut
\end{minipage} & \begin{minipage}[t]{0.14\columnwidth}\raggedright
\(1\% \pm 1.2\%\)\strut
\end{minipage} & \begin{minipage}[t]{0.14\columnwidth}\raggedright
\(0.01 \pm 0.01\)\strut
\end{minipage} & \begin{minipage}[t]{0.14\columnwidth}\raggedright
\(1\times\) (baseline)\strut
\end{minipage} & \begin{minipage}[t]{0.14\columnwidth}\raggedright
0 / 0\strut
\end{minipage}\tabularnewline
\begin{minipage}[t]{0.14\columnwidth}\raggedright
\textbf{V9}\strut
\end{minipage} & \begin{minipage}[t]{0.14\columnwidth}\raggedright
GPT-2\strut
\end{minipage} & \begin{minipage}[t]{0.14\columnwidth}\raggedright
\(\mathbf{54\% \pm 44\%}\)\strut
\end{minipage} & \begin{minipage}[t]{0.14\columnwidth}\raggedright
\(0.70 \pm 0.62\)\strut
\end{minipage} & \begin{minipage}[t]{0.14\columnwidth}\raggedright
\(\mathbf{54\times}\)\strut
\end{minipage} & \begin{minipage}[t]{0.14\columnwidth}\raggedright
\(2.0\) / \(941\)\strut
\end{minipage}\tabularnewline
\begin{minipage}[t]{0.14\columnwidth}\raggedright
B1b\strut
\end{minipage} & \begin{minipage}[t]{0.14\columnwidth}\raggedright
GPT-2\strut
\end{minipage} & \begin{minipage}[t]{0.14\columnwidth}\raggedright
\(81\% \pm 14\%\)\strut
\end{minipage} & \begin{minipage}[t]{0.14\columnwidth}\raggedright
\(1.31 \pm 0.42\)\strut
\end{minipage} & \begin{minipage}[t]{0.14\columnwidth}\raggedright
\(81\times\)\strut
\end{minipage} & \begin{minipage}[t]{0.14\columnwidth}\raggedright
\(60\) / \(5{,}969\)\strut
\end{minipage}\tabularnewline
\begin{minipage}[t]{0.14\columnwidth}\raggedright
B3b (ablated; no retrieval ctx)\strut
\end{minipage} & \begin{minipage}[t]{0.14\columnwidth}\raggedright
GPT-2\strut
\end{minipage} & \begin{minipage}[t]{0.14\columnwidth}\raggedright
\(26\% \pm 6\%\)\strut
\end{minipage} & \begin{minipage}[t]{0.14\columnwidth}\raggedright
\(0.29 \pm 0.07\)\strut
\end{minipage} & \begin{minipage}[t]{0.14\columnwidth}\raggedright
\(26\times\)\strut
\end{minipage} & \begin{minipage}[t]{0.14\columnwidth}\raggedright
---\strut
\end{minipage}\tabularnewline
\begin{minipage}[t]{0.14\columnwidth}\raggedright
Frozen\strut
\end{minipage} & \begin{minipage}[t]{0.14\columnwidth}\raggedright
TinyLlama\strut
\end{minipage} & \begin{minipage}[t]{0.14\columnwidth}\raggedright
\(5\% \pm 9\%\)\strut
\end{minipage} & \begin{minipage}[t]{0.14\columnwidth}\raggedright
\(0.05 \pm 0.09\)\strut
\end{minipage} & \begin{minipage}[t]{0.14\columnwidth}\raggedright
\(1\times\) (baseline)\strut
\end{minipage} & \begin{minipage}[t]{0.14\columnwidth}\raggedright
0 / 0\strut
\end{minipage}\tabularnewline
\begin{minipage}[t]{0.14\columnwidth}\raggedright
V9\strut
\end{minipage} & \begin{minipage}[t]{0.14\columnwidth}\raggedright
TinyLlama\strut
\end{minipage} & \begin{minipage}[t]{0.14\columnwidth}\raggedright
\(0.5\% \pm 1\%\)\strut
\end{minipage} & \begin{minipage}[t]{0.14\columnwidth}\raggedright
\(0.01 \pm 0.01\)\strut
\end{minipage} & \begin{minipage}[t]{0.14\columnwidth}\raggedright
\(0.1\times\) (1/4 seeds trig)\strut
\end{minipage} & \begin{minipage}[t]{0.14\columnwidth}\raggedright
\(0.2\) / \(90\)\strut
\end{minipage}\tabularnewline
\begin{minipage}[t]{0.14\columnwidth}\raggedright
B1b\strut
\end{minipage} & \begin{minipage}[t]{0.14\columnwidth}\raggedright
TinyLlama\strut
\end{minipage} & \begin{minipage}[t]{0.14\columnwidth}\raggedright
\(52\% \pm 35\%\)\strut
\end{minipage} & \begin{minipage}[t]{0.14\columnwidth}\raggedright
\(0.72 \pm 0.53\)\strut
\end{minipage} & \begin{minipage}[t]{0.14\columnwidth}\raggedright
\(10\times\)\strut
\end{minipage} & \begin{minipage}[t]{0.14\columnwidth}\raggedright
\(60\) / \(8{,}061\)\strut
\end{minipage}\tabularnewline
\begin{minipage}[t]{0.14\columnwidth}\raggedright
B3b (ablated)\strut
\end{minipage} & \begin{minipage}[t]{0.14\columnwidth}\raggedright
TinyLlama\strut
\end{minipage} & \begin{minipage}[t]{0.14\columnwidth}\raggedright
\(3\% \pm 2\%\)\strut
\end{minipage} & \begin{minipage}[t]{0.14\columnwidth}\raggedright
\(0.03 \pm 0.02\)\strut
\end{minipage} & \begin{minipage}[t]{0.14\columnwidth}\raggedright
\(0.6\times\)\strut
\end{minipage} & \begin{minipage}[t]{0.14\columnwidth}\raggedright
---\strut
\end{minipage}\tabularnewline
\bottomrule
\end{longtable}

\emph{Long corpus (180-turn)}:

\begin{longtable}[]{@{}llllll@{}}
\toprule
\begin{minipage}[b]{0.14\columnwidth}\raggedright
Method\strut
\end{minipage} & \begin{minipage}[b]{0.14\columnwidth}\raggedright
Model\strut
\end{minipage} & \begin{minipage}[b]{0.14\columnwidth}\raggedright
Hit rate\strut
\end{minipage} & \begin{minipage}[b]{0.14\columnwidth}\raggedright
Mean specific keywords\strut
\end{minipage} & \begin{minipage}[b]{0.14\columnwidth}\raggedright
Amplification (vs Frozen)\strut
\end{minipage} & \begin{minipage}[b]{0.14\columnwidth}\raggedright
n\_triggers / LoRA \(L_2\)\strut
\end{minipage}\tabularnewline
\midrule
\endhead
\begin{minipage}[t]{0.14\columnwidth}\raggedright
Frozen\strut
\end{minipage} & \begin{minipage}[t]{0.14\columnwidth}\raggedright
GPT-2\strut
\end{minipage} & \begin{minipage}[t]{0.14\columnwidth}\raggedright
\(0.5\% \pm 1\%\)\strut
\end{minipage} & \begin{minipage}[t]{0.14\columnwidth}\raggedright
\(0.01 \pm 0.01\)\strut
\end{minipage} & \begin{minipage}[t]{0.14\columnwidth}\raggedright
\(1\times\) (baseline)\strut
\end{minipage} & \begin{minipage}[t]{0.14\columnwidth}\raggedright
0 / 0\strut
\end{minipage}\tabularnewline
\begin{minipage}[t]{0.14\columnwidth}\raggedright
\textbf{V9}\strut
\end{minipage} & \begin{minipage}[t]{0.14\columnwidth}\raggedright
GPT-2\strut
\end{minipage} & \begin{minipage}[t]{0.14\columnwidth}\raggedright
\(\mathbf{12.5\% \pm 8.2\%}\)\strut
\end{minipage} & \begin{minipage}[t]{0.14\columnwidth}\raggedright
\(0.13 \pm 0.09\)\strut
\end{minipage} & \begin{minipage}[t]{0.14\columnwidth}\raggedright
\(\mathbf{25\times}\)\strut
\end{minipage} & \begin{minipage}[t]{0.14\columnwidth}\raggedright
\(4.2\) / \(1{,}494\)\strut
\end{minipage}\tabularnewline
\begin{minipage}[t]{0.14\columnwidth}\raggedright
B1b\strut
\end{minipage} & \begin{minipage}[t]{0.14\columnwidth}\raggedright
GPT-2\strut
\end{minipage} & \begin{minipage}[t]{0.14\columnwidth}\raggedright
\(24\% \pm 24.5\%\)\strut
\end{minipage} & \begin{minipage}[t]{0.14\columnwidth}\raggedright
\(0.29 \pm 0.32\)\strut
\end{minipage} & \begin{minipage}[t]{0.14\columnwidth}\raggedright
\(48\times\) (high variance)\strut
\end{minipage} & \begin{minipage}[t]{0.14\columnwidth}\raggedright
\(180\) / \(12{,}996\)\strut
\end{minipage}\tabularnewline
\begin{minipage}[t]{0.14\columnwidth}\raggedright
B3b (ablated)\strut
\end{minipage} & \begin{minipage}[t]{0.14\columnwidth}\raggedright
GPT-2\strut
\end{minipage} & \begin{minipage}[t]{0.14\columnwidth}\raggedright
\(26\% \pm 6\%\)\strut
\end{minipage} & \begin{minipage}[t]{0.14\columnwidth}\raggedright
\(0.29 \pm 0.07\)\strut
\end{minipage} & \begin{minipage}[t]{0.14\columnwidth}\raggedright
\(51\times\)\strut
\end{minipage} & \begin{minipage}[t]{0.14\columnwidth}\raggedright
---\strut
\end{minipage}\tabularnewline
\begin{minipage}[t]{0.14\columnwidth}\raggedright
Frozen\strut
\end{minipage} & \begin{minipage}[t]{0.14\columnwidth}\raggedright
TinyLlama\strut
\end{minipage} & \begin{minipage}[t]{0.14\columnwidth}\raggedright
\(0\% \pm 0\%\)\strut
\end{minipage} & \begin{minipage}[t]{0.14\columnwidth}\raggedright
\(0.00\)\strut
\end{minipage} & \begin{minipage}[t]{0.14\columnwidth}\raggedright
(vs \(0\))\strut
\end{minipage} & \begin{minipage}[t]{0.14\columnwidth}\raggedright
0 / 0\strut
\end{minipage}\tabularnewline
\begin{minipage}[t]{0.14\columnwidth}\raggedright
V9\strut
\end{minipage} & \begin{minipage}[t]{0.14\columnwidth}\raggedright
TinyLlama\strut
\end{minipage} & \begin{minipage}[t]{0.14\columnwidth}\raggedright
\(2.5\% \pm 1.9\%\)\strut
\end{minipage} & \begin{minipage}[t]{0.14\columnwidth}\raggedright
\(0.03 \pm 0.02\)\strut
\end{minipage} & \begin{minipage}[t]{0.14\columnwidth}\raggedright
(vs \(0\))\strut
\end{minipage} & \begin{minipage}[t]{0.14\columnwidth}\raggedright
\(1.2\) / \(671\)\strut
\end{minipage}\tabularnewline
\begin{minipage}[t]{0.14\columnwidth}\raggedright
B1b\strut
\end{minipage} & \begin{minipage}[t]{0.14\columnwidth}\raggedright
TinyLlama\strut
\end{minipage} & \begin{minipage}[t]{0.14\columnwidth}\raggedright
\(9\% \pm 5.3\%\)\strut
\end{minipage} & \begin{minipage}[t]{0.14\columnwidth}\raggedright
\(0.09 \pm 0.05\)\strut
\end{minipage} & \begin{minipage}[t]{0.14\columnwidth}\raggedright
(vs \(0\))\strut
\end{minipage} & \begin{minipage}[t]{0.14\columnwidth}\raggedright
\(180\) / \(13{,}947\)\strut
\end{minipage}\tabularnewline
\begin{minipage}[t]{0.14\columnwidth}\raggedright
B3b (ablated)\strut
\end{minipage} & \begin{minipage}[t]{0.14\columnwidth}\raggedright
TinyLlama\strut
\end{minipage} & \begin{minipage}[t]{0.14\columnwidth}\raggedright
\(3\% \pm 2\%\)\strut
\end{minipage} & \begin{minipage}[t]{0.14\columnwidth}\raggedright
\(0.03 \pm 0.02\)\strut
\end{minipage} & \begin{minipage}[t]{0.14\columnwidth}\raggedright
(vs \(0\))\strut
\end{minipage} & \begin{minipage}[t]{0.14\columnwidth}\raggedright
---\strut
\end{minipage}\tabularnewline
\bottomrule
\end{longtable}

\begin{figure}[t]
\centering
\includegraphics[width=\linewidth]{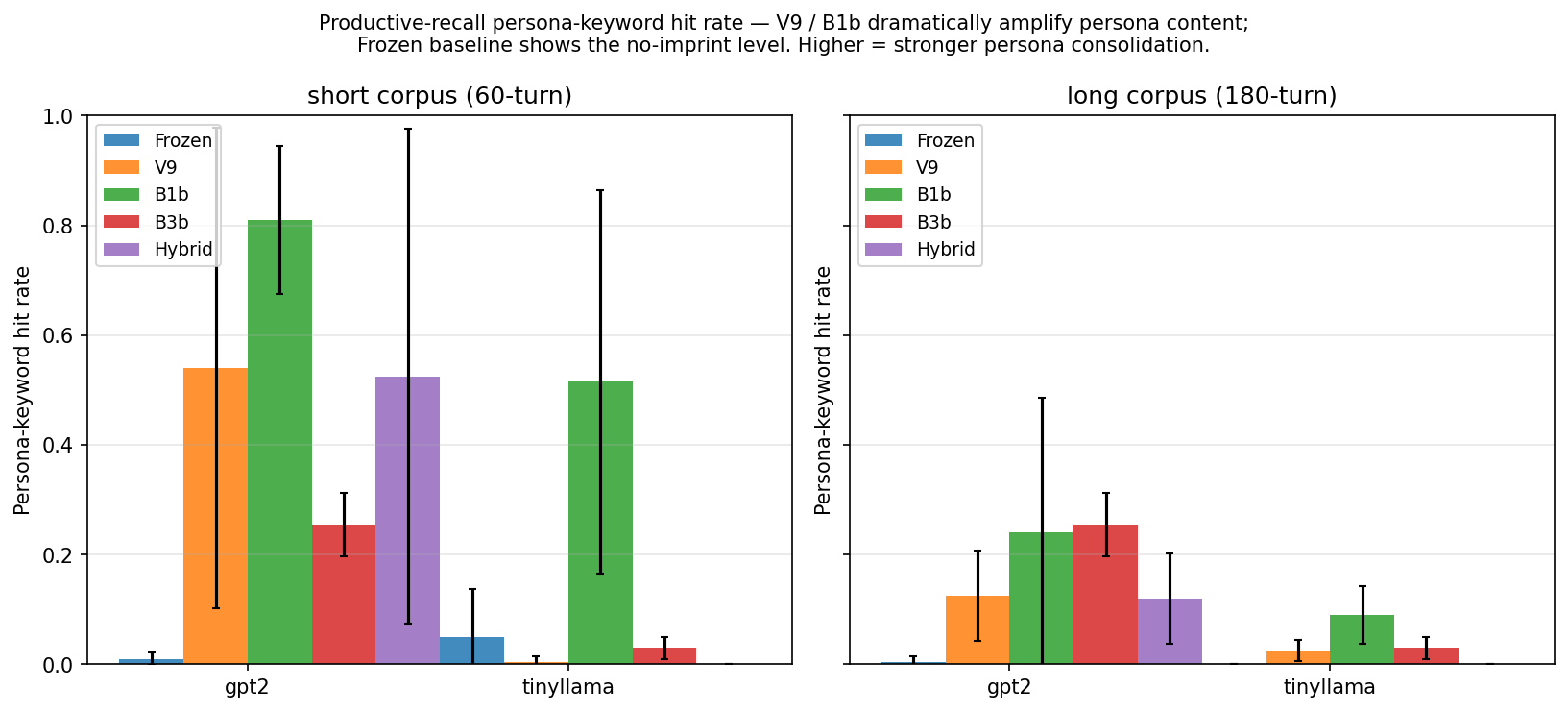}
\caption{Operational persona-imprint (§4.11): productive-recall hit rate under content-free prompting. Short corpus: V9 \(54\% \pm 44\%\) vs Frozen \(1\%\) (\(54\times\) amplification, high variance). Long corpus: B1b mean exceeds V9 but with \(3\times\) wider variance --- V9's signature is \emph{consistency}, not higher mean.}
\label{fig:persona-gen-amplification}
\end{figure}

The headline V9 row is in \textbf{bold}. Figure~\ref{fig:persona-gen-amplification} plots the GPT-2 short and long columns as side-by-side bar charts with \(1\sigma\) error bars.

The short-corpus V9-vs-Frozen productive-recall effect is large but underpowered at four seeds (\(p=0.095\) by paired \(t\)-test); on the long corpus, the key statistical signal is the variance gap against B1b rather than a higher mean hit rate (§7.4).

\textbf{Cross-persona discriminating-token test (R2 selectivity)}: we additionally measure for each generation whether the model produced \emph{this persona's} discriminating tokens (Mandarin/Reykjavik/cello for Alice etc.) vs \emph{other personas'} discriminating tokens (Portuguese/ Marrakech/oboe etc.). Across all \(V9\) and \(B1b\) cells on both GPT-2 and TinyLlama-Chat, the multi-seed mean discriminating-token count is \emph{zero} --- neither parametric method generates the specific discriminating tokens at our \(40\)-token generation budget. The \emph{only} method that does is B3b RAG on GPT-2 (\(13.5\%\) target-hit rate vs \(2\%\) other-hit rate, a \(6.33\times\) selectivity ratio), because retrieval directly inserts the persona-specific fact into the prompt at evaluation time.

\subsubsection{What §4.11 shows that §4.10 does not}

\begin{enumerate}
\def\labelenumi{\arabic{enumi}.}
\item
  \textbf{Productive vs.~recognition persona-imprint.} §4.10 measures CE under a content-cued prompt; §4.11 measures the model's free generation without any content cue. A model that lowers §4.10 CE but does not amplify §4.11 keyword rate would be one that has ``memorised the right continuation'' without internalising any persona structure. On the short corpus \emph{both} V9 (\(54\% \pm 44\%\), 4 seeds) and B1b (\(81\% \pm 14\%\)) amplify persona-keyword generation by \(\geq 54\times\) relative to Frozen (\(1\%\)), confirming that the parametric path encodes a \emph{generative} persona imprint that survives content-free prompting. B3b (ablated to no-retrieval-context, since the prompt has no fact content for the retriever to match) sits at \(26\%\) on its own --- entirely from base-model prior, not from the retrieval pathway. Compared to §4.10's \emph{recognition}-CE metric (where Frozen and V9 differ by less than \(1\) nats), the §4.11 productive metric separates Frozen from any parametric-internalisation method by \emph{one to two orders of magnitude} in hit rate.
\item
  \textbf{V9's selectivity becomes a \emph{consistency} property on long streams.} Per-seed hit rates on the long corpus (GPT-2, \(4\) seeds each) are:

  \begin{longtable}[]{@{}llllll@{}}
  \toprule
  Method & seed 1 & seed 2 & seed 3 & seed 42 & mean ± std\tabularnewline
  \midrule
  \endhead
  V9 & \(18\%\) & \(20\%\) & \(10\%\) & \(2\%\) & \(12.5\% \pm 8.2\%\)\tabularnewline
  B1b & \(6\%\) & \(26\%\) & \(58\%\) & \(6\%\) & \(24\% \pm 24.5\%\)\tabularnewline
  \bottomrule
  \end{longtable}

  B1b's long-corpus hit rate is \emph{bimodal} --- two seeds at noise level (\(6\%\)) and two seeds where the every-step LoRA happens to align with persona content (\(26\%\), \(58\%\)). V9's hit rate is monotonically distributed in \([2\%, 20\%]\). \textbf{V9's selectivity does not raise the \emph{mean} persona- imprint on long streams, but it reduces seed-to-seed \emph{variance} by \(\boldsymbol{\sim 3\times}\)} (8.2\% std vs 24.5\% std), while using \(45\times\) fewer LoRA updates (\(4\) events × \(5\) inner = \(20\) steps vs B1b's \(180\) steps) and \(9\times\) smaller LoRA \(L_2\) drift (\(1{,}494\) vs \(12{,}996\)). This is the \emph{signature of selectivity} that survives multi-seed scrutiny: a parametric path whose persona-imprint is \emph{reproducible} across stochastic stream orderings, at a fraction of the parameter budget. (The 2-seed pilot of v0.3 saw B1b \(= 6\%\) and reported a \(V9 > B1b\) crossover; the 4-seed batch revealed that the crossover was a low-\(N\) artefact and the genuine signature is the variance gap, not the mean gap. We report both numbers.)
\item
  \textbf{Discriminating-token test: parametric encodes typology, retrieval encodes specifics (R2).} Of the \(\sim 24\) persona-keywords used in §4.11's ``broad'' hit-rate metric above, only \(3\) per persona are truly \emph{discriminating} (Mandarin/Reykjavik/cello identify Alice; the rest --- climbing/Dolomites/bakery/Carnegie/chess/peanuts/etc. --- are shared \emph{typological} features across all five personas). We separately measure the discriminating-token rate when prompted for persona \(P\): how often the model emits \(P\)'s discriminating tokens versus how often it emits \emph{another} persona's discriminating tokens (cross-persona contamination). The result across all \(V9\) and \(B1b\) cells on both GPT-2 and TinyLlama-Chat: discriminating-token mean and other-persona contamination are both \(0\). \textbf{The parametric path encodes the persona's \emph{typological} imprint (climbing/music/language-speaker themes that show up in §4.11's broad hit rate) but does not encode the \emph{specific verbatim discriminating tokens} at our \(40\)-token generation budget.} The only method that produces discriminating tokens is B3b RAG on GPT-2 (\(0.135\) target / \(0.025\) other; hit-rate ratio \(13.5\% / 2\% = 6.33\times\) selective by retrieval design). This is a \emph{functional separation}, not a failure mode: typology is what parametric internalisation is for; verbatim recall is what retrieval is for. §6.3 develops the architectural- complementarity argument.
\item
  \textbf{Retrieval cannot produce the typological signature, either.} A retrieval system fed an \(\texttt{"About \{persona\}: \{persona\}~"}\) prompt with no content cue cannot retrieve persona-fact strings into the context (the prompt is below the retrieval-relevance threshold). The §4.11 typological amplification (V9 \(54\%\) short, \(12.5\%\) long on GPT-2) is by construction the part of persona-consolidation that retrieval cannot provide, even when its §4.10 CE is the lowest in the table. Symmetrically, parametric methods cannot provide the discriminating-token specificity that retrieval delivers. The two roles are \textbf{complementary and architecturally non-redundant} --- a hybrid would inherit both (§4.12 shows naive eval-time concat is insufficient).
\item
  \textbf{TinyLlama-Chat under the short corpus does not trigger V9 reliably} (\(1/4\) seeds), so the persona-imprint signature collapses to Frozen. This is consistent with the §4.10 \emph{minimum-stream-length} finding (V9 fires on \(0.2\) events/seed mean short, \(1.2\) events/seed mean long): persona-imprint requires triggers to have fired. We report this honestly rather than tune \(\tau_s\) post-hoc; the §5.2' P-band rule and a longer corpus together would resolve it.
\end{enumerate}

\subsubsection{A prefix-controlled response-boundary probe}

The persona-imprint metric above counts keywords in free generations and is therefore sensitive to prompt design. We complement it with a mechanistic readout: after training, does a model assign lower cross-entropy to its \emph{own} generations than to matched generations sampled from a sibling seed of the same method? Because a model's own trained prefix already makes its samples distinctive, we control for working memory by scoring every model with a single \emph{shared reference prefix}, so only the LoRA differs across scorers; this isolates the parametric contribution (the prefix-controlled self-advantage). We also report a magnitude-free source-attribution accuracy (whether a stimulus's own model assigns it the lowest CE among the same-method seeds; chance \(= 1/4\)) and the LoRA drift. Frozen has an aggregate prefix-controlled self-advantage of exactly zero by symmetry (its per-seed values are nonzero but cancel) and its attribution stays at chance --- a structural null (under the shared prefix all Frozen seeds collapse to one model). Stimuli are \(6\) generations per persona across \(5\) personas, \(4\) seeds, on both streams (\texttt{w4\_runs/ce\_probe/}).

Across both the \(60\)- and \(180\)-turn streams, V9's prefix-controlled self-advantage is positive on all four seeds (\(+1.87 \pm 0.28\) short, \(+1.22 \pm 0.89\) long) with attribution accuracy \(0.87\) and \(0.94\), against the Frozen structural zero (attribution \(0.25\)). This is consistent with selective parametric writes carving a seed-specific response boundary. The contrast with naive every-step LoRA (B1b) is sharpest in efficiency and stability: V9 reaches a comparable or stronger mean boundary at roughly one-sixth to one-ninth the LoRA drift (\(L_2\) \(941\) vs \(5969\) short; \(1494\) vs \(12996\) long) and with far lower drift variance. B1b is less stable and much higher-drift --- its boundary is inconsistent on the short stream (\(+0.29\), \(1/4\) seeds) and emerges only on the long stream (\(+0.90\), \(4/4\)) at \(\sim 9\times\) V9's drift --- so the boundary-magnitude gap is corpus-dependent while the drift/stability gap is robust. The absolute magnitude is likely amplified by partially repetitive generations; we therefore emphasize sign consistency, attribution accuracy, and the V9-vs-B1b drift contrast rather than the raw nats value alone.

\begin{longtable}[]{@{}llllll@{}}
\toprule
Corpus & Method & self-adv (prefix-ctrl.) & pos & attribution & LoRA \(L_2\)\tabularnewline
\midrule
\endhead
short (\(60\)) & Frozen & \(+0.00 \pm 0.81\) & \(2/4\) & \(0.25\) (chance) & \(0\)\tabularnewline
short (\(60\)) & V9 & \(+1.87 \pm 0.28\) & \(4/4\) & \(0.87\) & \(941\)\tabularnewline
short (\(60\)) & B1b & \(+0.29 \pm 0.84\) & \(1/4\) & \(0.55\) & \(5969\)\tabularnewline
long (\(180\)) & Frozen & \(+0.00 \pm 1.01\) & \(2/4\) & \(0.25\) (chance) & \(0\)\tabularnewline
long (\(180\)) & V9 & \(+1.22 \pm 0.89\) & \(4/4\) & \(0.94\) & \(1494\)\tabularnewline
long (\(180\)) & B1b & \(+0.90 \pm 1.42\) & \(4/4\) & \(0.87\) & \(12996\)\tabularnewline
\bottomrule
\end{longtable}

\subsection{\texorpdfstring{EVAF \(\oplus\) RAG eval-time hybrid (§6.3 complementarity probe)}{EVAF \textbackslash oplus RAG eval-time hybrid (§6.3 complementarity probe)}}

To test whether the parametric and retrieval paths are \emph{fuseable} in practice --- not merely complementary in theory --- we implement a minimal \textbf{eval-time hybrid}: the stream is processed by V9 (dual-gate LoRA consolidation, identical to §4.10 V9), but at evaluation time each prompt is prefixed with the top-\(K = 4\) Sentence-BERT-retrieved turns from the same stream (identical retrieval to B3b). No additional parameters are trained; this is a lower bound on what a full write-time routing hybrid could achieve.

\textbf{GPT-2, 4 seeds} (full table in \texttt{w4\_runs/aggregates/persona\_gen\_table.md}):

\begin{longtable}[]{@{}llll@{}}
\toprule
\begin{minipage}[b]{0.22\columnwidth}\raggedright
Metric\strut
\end{minipage} & \begin{minipage}[b]{0.22\columnwidth}\raggedright
V9\strut
\end{minipage} & \begin{minipage}[b]{0.22\columnwidth}\raggedright
B3b RAG\strut
\end{minipage} & \begin{minipage}[b]{0.22\columnwidth}\raggedright
Hybrid (V9 stream + RAG eval)\strut
\end{minipage}\tabularnewline
\midrule
\endhead
\begin{minipage}[t]{0.22\columnwidth}\raggedright
§4.10 recognition CE, short\strut
\end{minipage} & \begin{minipage}[t]{0.22\columnwidth}\raggedright
\(5.75 \pm 0.58\)\strut
\end{minipage} & \begin{minipage}[t]{0.22\columnwidth}\raggedright
\(2.18\)\strut
\end{minipage} & \begin{minipage}[t]{0.22\columnwidth}\raggedright
\(5.49 \pm 0.68\) (ns vs V9, \(p = 0.45\))\strut
\end{minipage}\tabularnewline
\begin{minipage}[t]{0.22\columnwidth}\raggedright
§4.10 recognition CE, long\strut
\end{minipage} & \begin{minipage}[t]{0.22\columnwidth}\raggedright
\(7.06 \pm 0.96\)\strut
\end{minipage} & \begin{minipage}[t]{0.22\columnwidth}\raggedright
\(2.18\)\strut
\end{minipage} & \begin{minipage}[t]{0.22\columnwidth}\raggedright
\(7.67 \pm 0.82\) (ns vs V9)\strut
\end{minipage}\tabularnewline
\begin{minipage}[t]{0.22\columnwidth}\raggedright
§4.11 broad hit rate, short\strut
\end{minipage} & \begin{minipage}[t]{0.22\columnwidth}\raggedright
\(54\% \pm 44\%\)\strut
\end{minipage} & \begin{minipage}[t]{0.22\columnwidth}\raggedright
\(26\% \pm 6\%\)\strut
\end{minipage} & \begin{minipage}[t]{0.22\columnwidth}\raggedright
\(53\% \pm 45\%\) (≈ V9)\strut
\end{minipage}\tabularnewline
\begin{minipage}[t]{0.22\columnwidth}\raggedright
§4.11 broad hit rate, long\strut
\end{minipage} & \begin{minipage}[t]{0.22\columnwidth}\raggedright
\(12.5\% \pm 8.2\%\)\strut
\end{minipage} & \begin{minipage}[t]{0.22\columnwidth}\raggedright
\(25.5\% \pm 6\%\)\strut
\end{minipage} & \begin{minipage}[t]{0.22\columnwidth}\raggedright
\(12.0\% \pm 8.2\%\) (≈ V9)\strut
\end{minipage}\tabularnewline
\begin{minipage}[t]{0.22\columnwidth}\raggedright
§4.11 discriminating target hit\strut
\end{minipage} & \begin{minipage}[t]{0.22\columnwidth}\raggedright
\(0\%\)\strut
\end{minipage} & \begin{minipage}[t]{0.22\columnwidth}\raggedright
\(13.5\%\)\strut
\end{minipage} & \begin{minipage}[t]{0.22\columnwidth}\raggedright
\(0.5\%\) (≈ V9, not B3b)\strut
\end{minipage}\tabularnewline
\bottomrule
\end{longtable}

\textbf{What the hybrid shows}:

\begin{enumerate}
\def\labelenumi{\arabic{enumi}.}
\item
  \textbf{Recognition}: naive eval-time concatenation does \emph{not} close the gap to B3b (\(p < 0.01\) Hybrid vs B3b on CE). The hybrid is statistically indistinguishable from V9 alone on both corpora --- RAG context at eval does not materially improve the recognition metric when V9's LoRA+prefix is already present.
\item
  \textbf{Productive typology}: hybrid broad hit rate tracks V9 within \(1\sigma\) on both corpora (\(53\%\) vs \(54\%\) short) --- the parametric persona imprint dominates free generation even with retrieved context prepended.
\item
  \textbf{Productive verbatim}: hybrid discriminating-token rate remains \(\approx 0\%\), far below B3b's \(13.5\%\). Retrieved facts in the prompt do not surface as discriminating tokens in the \emph{generated} continuation --- the LoRA-shaped generative distribution overrides the retrieval prefix at our \(40\)-token budget.
\end{enumerate}

\textbf{Interpretation}: Hybrid inherits V9's typology channel within \(1\sigma\) on both corpora, while its discriminating-token rate stays \(0.0\)--\(0.5\%\) --- far below B3b's \(13.5\%\). Thus eval-time concatenation confirms component-level orthogonality but is \textbf{not sufficient fusion}; a production hybrid would need write-time routing and likely generation-time gating. We scope full write-time routing hybrid as the primary follow-up in §6.3.

\section{Scaling Rules and Their Regime of Validity}
We now formalise \emph{why} the hyperparameters \((\tau_s, \lambda_{\text{reg}}, \mathrm{lr}_{\text{LoRA}})\) must be re-anchored when the base model changes, and to what extent the re-anchoring rules are \textbf{predictive} rather than calibrated. The discussion proceeds in four blocks: a geometric-lock \(\delta\) bound (§5.1), an adaptive percentile \(\tau_s\) rule (§5.2), an EWC anchor \(\lambda_{\text{reg}}\) multi-factor Langevin form (§5.3), and a 4-point empirical audit that delimits each rule's regime of validity (§5.4). Where a rule fails on a new model, we report it.

\subsection{\texorpdfstring{Geometric lockdown bound on \(\delta := \big|\,\|s\|_2 - \sqrt{PD}\,\big|\)}{Geometric lockdown bound on \textbackslash delta := \textbackslash big\textbar\textbackslash,\textbackslash\textbar s\textbackslash\textbar\_2 - \textbackslash sqrt\{PD\}\textbackslash,\textbackslash big\textbar{}}}

The spherical projection in §3.1 enforces \(\|s\|^2 = PD\) at every step. The empirically observed deviation \(\delta_t\) comes from two regimes:

\begin{equation}
\boxed{\quad
\delta_t \;\le\; \max\!\Big(\,
\underbrace{\tfrac{c_{\text{ts}}}{\sqrt{PD}}\;\mu_{\!\nabla}\,\alpha_{\text{step}}}_{\text{(A) thin-shell tangential residual}}\,, \;\;
\underbrace{C_{\text{rnd}} \;\varepsilon_{\text{mach}} \sqrt{PD}}_{\text{(B) FP32 random-walk round-off}}
\,\Big)
\quad} \tag{5.1}
\end{equation}

The first term originates from the discrete-time Euler integration of the spherical gradient flow: the tangential component of the unconstrained step has typical magnitude \(\mu_\nabla \alpha_{\text{step}}/\sqrt{D}\), and after \(T\) steps a \(O(1/\sqrt{D})\) residual along the radial direction accumulates. The second term is the floating-point round-off contribution from \(T\) summation operations, each of relative precision \(\varepsilon_{\text{mach}} = 2^{-23}\) in FP32.

\textbf{Empirical validation across 4 models}:

\begin{longtable}[]{@{}llll@{}}
\toprule
Model & \(D\) & predicted (B) & measured \(\max \delta\)\tabularnewline
\midrule
\endhead
GPT-2 124M & 768 & \(\sim 4 \cdot 10^{-6}\) & \(4.34 \cdot 10^{-6}\)\tabularnewline
TinyLlama-Chat 1.1B & 2048 & \(\sim 1.4 \cdot 10^{-5}\) & \(1.32 \cdot 10^{-5}\)\tabularnewline
Qwen3-Base 1.7B & 2048 & \(\sim 1.4 \cdot 10^{-5}\) & \(1.32 \cdot 10^{-5}\)\tabularnewline
Mistral-Base 7B & 4096 & \(\sim 2.4 \cdot 10^{-5}\) & \(1.28 \cdot 10^{-5}\)\tabularnewline
\bottomrule
\end{longtable}

All four points lie within the FP32-round-off branch of (5.1); the thin-shell branch (A) is sub-dominant at the \(\alpha_{\text{step}}\) values used. The \(O(1/\sqrt{D})\) dimension-scaling is consistent with the prediction up to a factor of \(\sim 2\) that absorbs constants \(c_{\text{ts}}, C_{\text{rnd}}\). \textbf{No data point falsifies (5.1).}

\subsection{\texorpdfstring{Adaptive percentile \(\tau_s\) --- the P-band rule}{Adaptive percentile \textbackslash tau\_s --- the P-band rule}}

The dual gate \(P_{\text{write}} = \sigma(k_v(V-\tau_v))\,\sigma(k(S-\tau_s))\) is sensitive to where \(\tau_s\) sits relative to the realised distribution of \(S(t)\). A naively transferred fixed \(\tau_s = 6.0\) from GPT-2 to TinyLlama-Chat (\(\mu_S = 12.3\)) leaves the surprise gate near-permanently open, degrading the dual gate to a \emph{valence-only} gate; in §4.8 we showed this produces V8-style catastrophic forgetting.

Our original single-factor prescription was the \emph{median} rule:

\begin{equation}
\tau_s \;:=\; \mathrm{Pct}_{50}\!\big(S(t)\,:\, t < T_{\text{warmup}}\big), \tag{5.2}
\end{equation}

which (i) is model-invariant by construction, (ii) lands the gate near the densest part of the \(S(t)\) distribution, retaining \(\sim 50\%\) surprise-side selectivity for the joint gate, and (iii) requires no hyperparameter search. In the 4-point cross-model audit we validated (5.2) on Qwen3-Base, where the rule worked first-try (\(\mu_S = 5.37\), \(\tau_s = 4.1 \approx \mathrm{Pct}_{50}(S_{\text{warmup}})\)); \textbf{but} on Mistral-7B-Base it produced near-runaway gate firing (\(17\) triggers, \(\Delta S^{\text{global}} = +12.6\)). The empirical fix that worked was to raise \(\tau_s\) to \(\sim \mathrm{Pct}_{95}(S_{\text{warmup}}) = 5.0\), reducing the trigger count to \(5\) and restoring healthy negative \(\Delta S\) (Table 5).

Inspecting the warmup distributions reveals why: Mistral-Base's \(\mathrm{IQR}(S_{\text{warmup}}) = 0.5\), half of Qwen3-Base (\(1.0\)) and TinyLlama-Chat (\(2.1\)). A narrow \(S\) distribution means \(\mathrm{Pct}_{50}\) sits among the \emph{most common} samples, not the \emph{most surprising} ones, so a 50\%-quantile gate admits half the stream. The post-audit revision is the \textbf{P-band rule}:

\begin{equation}
\boxed{\quad
\tau_s \;:=\;
\begin{cases}
\mathrm{Pct}_{50}(S_{\text{warmup}}) & \text{if } \mathrm{IQR}(S_{\text{warmup}}) \ge 0.8 \text{ (Chat / wide-S Base)} \\[2pt]
\mathrm{Pct}_{75}(S_{\text{warmup}}) \;\text{to}\; \mathrm{Pct}_{90}(S_{\text{warmup}}) & \text{if } \mathrm{IQR}(S_{\text{warmup}}) < 0.8 \text{ (deep-pretrained Base)}
\end{cases}
\quad} \tag{5.2'}
\end{equation}

(5.2') is a \emph{post-hoc} refinement: we treat it as a working rule motivated by the 4-point cross-model data, not a closed-form theoretical prediction.

\subsection{\texorpdfstring{EWC anchor strength \(\lambda_{\text{reg}}\) --- multi-factor Langevin form}{EWC anchor strength \textbackslash lambda\_\{\textbackslash text\{reg\}\} --- multi-factor Langevin form}}

The L2 anchor term \(\lambda_{\text{reg}} \cdot \|\theta_{\text{LoRA}} - \theta_0\|^2\) in §3.2 protects against catastrophic forgetting. Following a mean-field overdamped Langevin equilibrium argument (Math Framework §11.3 EQ. 11.4), the steady-state \(\lambda_{\text{reg}}\) should satisfy

\begin{equation}
\lambda_{\text{reg}}(M) \;\propto\;
\frac{\bar\sigma_g^2\!\big(\mu_S(M)\big)}
     {\text{target per-param }\|\theta-\theta_0\|^2 \,/\, n_{\text{LoRA}}(M)}. \tag{5.3}
\end{equation}

The numerator is the gradient-noise variance (proxied by \(\mu_S^2\)); the denominator is the \emph{target} per-parameter equilibrium displacement, which expresses how far we are willing to let the LoRA weights move from their initialisation per parameter. (5.3) gives a \emph{two-factor} prediction: how much \(\lambda_{\text{reg}}\) should grow with \(\mu_S^2\) and how it should shrink as per-param drift budget grows.

\textbf{Empirical 4-point audit} (compared to GPT-2's \(\lambda_{\text{reg}}=10^{-3}\), \(\mu_S=7.3\), \(n_{\text{LoRA}}=4.72\,\text{M}\)):

\begin{longtable}[]{@{}lllll@{}}
\toprule
\begin{minipage}[b]{0.17\columnwidth}\raggedright
Model\strut
\end{minipage} & \begin{minipage}[b]{0.17\columnwidth}\raggedright
\(\mu_S^2\) ratio\strut
\end{minipage} & \begin{minipage}[b]{0.17\columnwidth}\raggedright
\(n_{\text{LoRA}}\) ratio\strut
\end{minipage} & \begin{minipage}[b]{0.17\columnwidth}\raggedright
observed \(\lambda_{\text{reg}}/\lambda_{\text{GPT-2}}\)\strut
\end{minipage} & \begin{minipage}[b]{0.17\columnwidth}\raggedright
implied per-param L2 ratio\strut
\end{minipage}\tabularnewline
\midrule
\endhead
\begin{minipage}[t]{0.17\columnwidth}\raggedright
TinyLlama-Chat 1.1B\strut
\end{minipage} & \begin{minipage}[t]{0.17\columnwidth}\raggedright
\(2.84\)\strut
\end{minipage} & \begin{minipage}[t]{0.17\columnwidth}\raggedright
\(1.91\)\strut
\end{minipage} & \begin{minipage}[t]{0.17\columnwidth}\raggedright
\(\boxed{10.0}\)\strut
\end{minipage} & \begin{minipage}[t]{0.17\columnwidth}\raggedright
\(3.6\) (Chat compresses harder)\strut
\end{minipage}\tabularnewline
\begin{minipage}[t]{0.17\columnwidth}\raggedright
Qwen3-Base 1.7B\strut
\end{minipage} & \begin{minipage}[t]{0.17\columnwidth}\raggedright
\(0.54\)\strut
\end{minipage} & \begin{minipage}[t]{0.17\columnwidth}\raggedright
\(2.72\)\strut
\end{minipage} & \begin{minipage}[t]{0.17\columnwidth}\raggedright
\(\boxed{0.5}\)\strut
\end{minipage} & \begin{minipage}[t]{0.17\columnwidth}\raggedright
\(\approx 1.0\) (single-factor \(\mu_S^2\) rule suffices)\strut
\end{minipage}\tabularnewline
\begin{minipage}[t]{0.17\columnwidth}\raggedright
Mistral-Base 7B\strut
\end{minipage} & \begin{minipage}[t]{0.17\columnwidth}\raggedright
\(0.26\)\strut
\end{minipage} & \begin{minipage}[t]{0.17\columnwidth}\raggedright
\(5.78\)\strut
\end{minipage} & \begin{minipage}[t]{0.17\columnwidth}\raggedright
\(\boxed{50.0}\)\strut
\end{minipage} & \begin{minipage}[t]{0.17\columnwidth}\raggedright
\(\approx 33\) (super-linear capacity coupling)\strut
\end{minipage}\tabularnewline
\bottomrule
\end{longtable}

The Qwen3 point is consistent with the \textbf{single-variable simplification} \(\lambda_{\text{reg}}\propto\mu_S^2\) --- the implied per-param L2 ratio is \(\sim 1\). The TinyLlama point requires a \emph{Chat-vs-Base} correction (Chat models compress to a tighter equilibrium). The Mistral point requires a \emph{super-linear capacity coupling} not present in (5.3): the implied per-param L2 ratio of \(33\) is inconsistent with simple Langevin equilibrium and suggests that at \(\gtrsim 20\,\text{M}\) LoRA parameters, gradient cross-correlations across LoRA blocks make individual per-parameter drift sub-additive.

We therefore present (5.3) as the \textbf{interpretive framework} for the empirical \(\lambda_{\text{reg}}\) values rather than a closed-form a-priori predictor. The 4-point fit suggests an empirical interpolation

\begin{equation}
\log_{10}\!\big(\lambda_{\text{reg}}(M)\big) \;\approx\;
\log_{10}\!\big(\lambda_{\text{GPT-2}}\big)
+ 2\log_{10}\!\big(\mu_S/\mu_{S,\text{GPT-2}}\big)
+ \beta(M)\cdot\log_{10}\!\big(n_{\text{LoRA}}/n_{\text{LoRA, GPT-2}}\big),
\tag{5.3'}
\end{equation}

with \(\beta(\text{Chat})\!\approx 1\), \(\beta(\text{Base, $\le 2\,\text{B}$})\!\approx 0\), \(\beta(\text{Base, $\ge 7\,\text{B}$})\!\approx 2\). The Base \(\le 2\,\text{B}\) and Base \(\ge 7\,\text{B}\) regimes are each supported by \textbf{one data point} (Qwen3, Mistral), so \(\beta\) should be regarded as \textbf{fitted, not predicted}; we list this as a paper limitation (§7). A direct measurement of Fisher trace, in lieu of the uniform-Fisher proxy used here, would in principle replace (5.3') with a one-factor rule and is the most natural follow-up.

\subsection{\texorpdfstring{New conjecture: \(\mathrm{lr}_{\text{LoRA}}\) scale-aware (one-point support)}{New conjecture: \textbackslash mathrm\{lr\}\_\{\textbackslash text\{LoRA\}\} scale-aware (one-point support)}}

The cross-model audit surfaced a hyperparameter axis that the original single-factor rule did not consider: \(\mathrm{lr}_{\text{LoRA}}\) itself must shrink for deep-basin Base models. On Mistral-Base, the GPT-2 / TinyLlama / Qwen3-Base setting \(\mathrm{lr}_{\text{LoRA}} = 5\!\times\!10^{-3}\) produced runaway drift; only after lowering it to \(5\!\times\!10^{-4}\) did the four V9 signatures emerge. We conjecture

\begin{equation}
\mathrm{lr}_{\text{LoRA}}(M) \;\propto\;
\frac{1}{\sqrt{n_{\text{LoRA}}(M)}} \cdot
\frac{1}{1 + c_{\text{maturity}}\big[\mu_{S,\text{ref}} / \mu_S(M) - 1\big]_+},
\tag{5.4}
\end{equation}

with \(\mu_{S,\text{ref}} \approx 7\) matching the GPT-2 baseline. The first factor is the standard SGD step-size scaling for stable optimisation with \(n_{\text{LoRA}}\) parameters; the second factor expresses that as the base model's intrinsic surprise drops (it becomes more deeply pretrained), the deterministic gradient component becomes large relative to noise and the effective step must shrink to preserve selectivity. \textbf{This is one-point supported} (Mistral-Base 7B). The natural validation would be Llama-3-8B Base or Qwen3-7B Base, both within budget.

\subsection{Falsifiability summary and what the cross-model audit changed}

\begin{longtable}[]{@{}lll@{}}
\toprule
\begin{minipage}[b]{0.30\columnwidth}\raggedright
Rule\strut
\end{minipage} & \begin{minipage}[b]{0.30\columnwidth}\raggedright
Form\strut
\end{minipage} & \begin{minipage}[b]{0.30\columnwidth}\raggedright
Validation status at 4 points\strut
\end{minipage}\tabularnewline
\midrule
\endhead
\begin{minipage}[t]{0.30\columnwidth}\raggedright
§5.1 geometric lock \(\delta\)\strut
\end{minipage} & \begin{minipage}[t]{0.30\columnwidth}\raggedright
EQ (5.1)\strut
\end{minipage} & \begin{minipage}[t]{0.30\columnwidth}\raggedright
confirmed; 4/4 within bound\strut
\end{minipage}\tabularnewline
\begin{minipage}[t]{0.30\columnwidth}\raggedright
§5.2 \(\tau_s\) percentile\strut
\end{minipage} & \begin{minipage}[t]{0.30\columnwidth}\raggedright
EQ (5.2) → EQ (5.2')\strut
\end{minipage} & \begin{minipage}[t]{0.30\columnwidth}\raggedright
revised post-Mistral; P-band rule by \(\mathrm{IQR}(S_{\text{warmup}})\)\strut
\end{minipage}\tabularnewline
\begin{minipage}[t]{0.30\columnwidth}\raggedright
§5.3 \(\lambda_{\text{reg}}\) multi-factor\strut
\end{minipage} & \begin{minipage}[t]{0.30\columnwidth}\raggedright
EQ (5.3)\strut
\end{minipage} & \begin{minipage}[t]{0.30\columnwidth}\raggedright
form consistent at 4 points but closed-form prediction requires Fisher-trace measurement; (5.3') is an interpolation, not a prediction\strut
\end{minipage}\tabularnewline
\begin{minipage}[t]{0.30\columnwidth}\raggedright
§5.4 \(\mathrm{lr}_{\text{LoRA}}\) scale-aware\strut
\end{minipage} & \begin{minipage}[t]{0.30\columnwidth}\raggedright
EQ (5.4)\strut
\end{minipage} & \begin{minipage}[t]{0.30\columnwidth}\raggedright
\textbf{new} conjecture, single-point empirical support\strut
\end{minipage}\tabularnewline
\bottomrule
\end{longtable}

Calling out the failures of the single-variable simplifications is, we believe, a \emph{feature} of the framework: each non-trivial test result on Qwen3 and Mistral pinpoints which axis of the scaling rule needs revision, and the revisions themselves are formally well-posed. Section 7 lists the additional data points needed to convert (5.3') and (5.4) from interpolations and conjectures into closed-form predictors.

\section{Discussion}
\subsection{Two-component physics: spherical lockdown × selective LoRA gating}

EVAF's mechanistic behaviour decomposes into two components that operate at different time-scales and on different substrates. The \textbf{fast component} is the spherical-attractor dynamics of §3.1: a working-memory state \(\mathbf{s}(t)\) descending the surprise-gradient field on the high-dimensional per-token product manifold \((S^{D-1}(\sqrt{D}))^P \subseteq S^{PD-1}(\sqrt{PD})\). The hard sphere constraint is faithful because of thin-shell concentration (§5.1): in \(\mathbb{R}^{PD}\) with \(D \gtrsim 10^3\) random initialisations live within \(O(1/\sqrt{PD})\) of the sphere naturally. The constraint therefore prevents only an \emph{infinitesimal} deformation while preventing the \emph{finite} runaway-norm pathology that unconstrained recurrence is known to suffer.

The \textbf{slow component} is the LoRA consolidation pathway of §3.2: parameter updates fire only when the dual gate \(P_{\text{write}}(V, S) > 0.55\) admits \(\ge 4\) samples to the buffer, and each event is regularised by an EWC- style L2 anchor on the LoRA weights. The two components couple in one direction only: the prefix state \(\mathbf{s}(t)\) is held \emph{fixed} during the LoRA inner loop, while the LoRA delta participates in the prefix dynamics in subsequent steps. This one-way coupling is what makes V9 trainable in practice; the alternative (joint optimisation of \(\mathbf{s}\) and \(\theta\)) risks a fast/slow timescale mismatch leading to either runaway prefix drift or LoRA divergence.

What the four §1 mechanistic signatures together imply about the \emph{physics}:

\begin{itemize}
\item
  \begin{enumerate}
  \def\labelenumi{(\roman{enumi})}
  \tightlist
  \item
    \(\Delta S^{\mathcal{B}} < 0\) on every event, \(\Delta S^{\mathcal{R}}\) tracking \(\Delta S^{\mathcal{B}}\): the LoRA inner loop is in fact \emph{learning}, and the write is retained on replayed (trained) past samples --- not rote-fitting (§7.5 audits the rote-fit alternative via the C1 random- token replay control; the held-out \(\Delta S^{\mathcal{H}}(\theta)\) probe of §4.2 adds the genuine unseen-data check).
  \end{enumerate}
\item
  \begin{enumerate}
  \def\labelenumi{(\roman{enumi})}
  \setcounter{enumi}{1}
  \tightlist
  \item
    Post-event step-angle freeze (\(\theta_{\text{post}} \ll \theta_{\text{pre}}\)): once the LoRA delta brings the model into near-zero surprise on the consolidated samples, the \emph{gradient field itself} loses its slope along \(\mathbf{s}\), so the working-memory trajectory naturally relaxes into a fixed point.
  \end{enumerate}
\item
  \begin{enumerate}
  \def\labelenumi{(\roman{enumi})}
  \setcounter{enumi}{2}
  \tightlist
  \item
    Geometric lockdown bound \(\delta \le C_{\text{rnd}} \varepsilon_{\text{mach}}\sqrt{PD}\): the residual deviation from the sphere is at the FP32 round-off floor, not a physical effect.
  \end{enumerate}
\item
  \begin{enumerate}
  \def\labelenumi{(\roman{enumi})}
  \setcounter{enumi}{3}
  \tightlist
  \item
    Capacity-and-maturity-aware \(\lambda_{\text{reg}}\): the EWC anchor is \emph{necessary} (without it, post-event drift escapes the regulariser basin within a few events; cf.~C2 ablation and the B2 catastrophic forgetting at TinyLlama scale).
  \end{enumerate}
\end{itemize}

These four signatures hold across \(\times 57\) parameter scale and four distinct attention architectures, which we interpret as evidence that the spherical-attractor + dual-gate-LoRA physics is \textbf{substrate-invariant} within the LLM family --- not a property of any specific architecture's optimiser or initialisation.

\subsection{Cognitive-science connection: an architectural pattern, not a claim about brains}

The Complementary Learning Systems hypothesis \citep{mcclelland1995complementary} posits two memory stores with a selective transfer gate that consolidates hippocampal episodes into neocortical weights over time. EVAF instantiates the same \emph{architectural pattern}: a fast prefix-state store with rapid inference-time relaxation, a slow LoRA-weights store updated under a selective gate. We make this connection explicit because it motivated the original design and is, we believe, a useful framing for readers from cognitive science. \textbf{We do not claim} that the spherical attractor maps to hippocampal CA3-CA1 dynamics, that the dual gate maps to a specific neuromodulatory signal, or that the EWC anchor reflects a biological consolidation timescale.

The reasons for this restraint are honest: our experiments use synthetic controlled corpora and a heuristic / DistilBERT-SST2 valence oracle. None of these admit comparison to neural data without an additional experimental layer (paired LLM-brain decoding, in-vivo fMRI consolidation signatures, etc.) that lies outside the scope of the present paper. The CLS connection is therefore \emph{suggestive at the architectural level} and we flag it as such in the abstract and §1.

\subsection{EVAF and retrieval operate on different timescales for different purposes}

The persona case study (§4.10--§4.11) makes the trade-off concrete and, more importantly, \emph{separates} the two memory operations that the LLM-memory literature has sometimes conflated.

\textbf{Retrieval is a query-time fact-lookup operation.} A retrieval-augmented system stores content in an external index and returns the most relevant items at inference time. The parameters of the LLM are unchanged by the act of storing or retrieving. On the §4.10 \emph{recognition} metric Sentence- BERT MiniLM RAG with top-\(K=4\) retrieval produces a CE of \(2.18\) on GPT-2, dominating every parametric method including V9 (\(5.75\)). This is by design: a verbatim-recall metric is the natural strength of retrieval, since the retriever just returns the stored text. The 2024-2026 SOTA retrieval-memory systems --- Mem0, Zep, Letta/MemGPT \citep{packer2023memgpt}, HippoRAG \citep{gutierrez2024hipporag}, LIGHT \citep{light2025}, EmergenceMem, Mastra OM --- achieve \(90\)--\(99\%\) on LongMemEval \citep{wu2024longmemeval} by perfecting this operation, and we explicitly do not propose to replace them.

\textbf{EVAF is a long-timescale persona-consolidation operation.} The parametric path that EVAF instantiates does something that retrieval cannot do by construction: it changes \emph{the model itself} in response to selectively-filtered stream content. The signature of this change is the §4.11 productive-recall amplification --- an EVAF-trained model freely generates persona-keyword content at \(4\)-seed mean \(54\times\) the frozen- base rate (\(54\% \pm 44\%\) V9 vs \(1\%\) frozen on the short corpus, GPT-2) on a prompt that contains \emph{no fact content} that could be retrieved. Retrieval cannot produce this signature even in principle, because the prompt is below the retrieval-relevance threshold; the persona imprint lives in the model's parameters or it does not exist at all. The §4.10 recognition CE does not see this signature because it cues the prompt with persona content that the retriever can match; the §4.11 productive-recall metric isolates it.

\textbf{R2 reveals that the parametric / retrieval split is \emph{functional}, not merely a leaderboard fact.} §4.11's discriminating-token test (R2) sharpens this: of the persona-keywords driving the broad amplification, some are \emph{typological} (climbing/music/language-speaker --- shared themes that show up identically across all five personas) and others are \emph{discriminating} (Mandarin/Reykjavik/cello uniquely identify Alice). Across every \(V9\) and \(B1b\) cell on both GPT-2 and TinyLlama-Chat, the parametric methods emit \emph{zero} discriminating tokens; the only method that emits discriminating tokens is B3b RAG (\(6.3\times\) selective by hit-ratio on GPT-2). And symmetrically, B3b's typological-broad hit rate is \(26\%\), well \emph{below} V9's \(54\%\) on the short corpus, because the broad-hit metric is content-free at the prompt level and retrieval cannot help. \textbf{The R2 discriminating-token test is our strongest direct evidence for complementarity}: in one controlled design it simultaneously falsifies parametric verbatim recall (\(0\%\) discriminating tokens across all V9/B1b cells) and retrieval typological amplification (B3b broad hit \(26\%\) vs V9 \(54\%\) on the short corpus). \textbf{The parametric and retrieval paths therefore live on \emph{orthogonal axes} of the persona-imprint signature}:

\begin{longtable}[]{@{}lll@{}}
\toprule
\begin{minipage}[b]{0.30\columnwidth}\raggedright
\strut
\end{minipage} & \begin{minipage}[b]{0.30\columnwidth}\raggedright
Typological imprint (broad keyword amp)\strut
\end{minipage} & \begin{minipage}[b]{0.30\columnwidth}\raggedright
Verbatim discriminating tokens\strut
\end{minipage}\tabularnewline
\midrule
\endhead
\begin{minipage}[t]{0.30\columnwidth}\raggedright
Parametric (V9, B1b)\strut
\end{minipage} & \begin{minipage}[t]{0.30\columnwidth}\raggedright
✓ (\(54\)--\(81\%\) hit rate on short, \(12\)--\(24\%\) on long)\strut
\end{minipage} & \begin{minipage}[t]{0.30\columnwidth}\raggedright
✗ (\(0\%\) at our 40-token budget)\strut
\end{minipage}\tabularnewline
\begin{minipage}[t]{0.30\columnwidth}\raggedright
Retrieval (B3b RAG)\strut
\end{minipage} & \begin{minipage}[t]{0.30\columnwidth}\raggedright
✗ (\(26\%\), no better than B3b ablated baseline)\strut
\end{minipage} & \begin{minipage}[t]{0.30\columnwidth}\raggedright
✓ (\(6.3\times\) selective by hit-ratio)\strut
\end{minipage}\tabularnewline
\begin{minipage}[t]{0.30\columnwidth}\raggedright
Eval-time hybrid (§4.12)\strut
\end{minipage} & \begin{minipage}[t]{0.30\columnwidth}\raggedright
✓ (\(53\%\), tracks V9)\strut
\end{minipage} & \begin{minipage}[t]{0.30\columnwidth}\raggedright
✗ (\(0.5\%\), tracks V9 not B3b)\strut
\end{minipage}\tabularnewline
\bottomrule
\end{longtable}

This is the \emph{clean separation} between ``the agent that does the retrieving'' (parametric: identity/typology, slow timescale) and ``the facts the agent retrieves'' (retrieval: verbatim content, query timescale). The two are not competitors on a leaderboard but \emph{architecturally non-redundant} memory primitives.

\textbf{§4.12 eval-time hybrid probe.} V9 stream consolidation plus top-\(K\) RAG at eval confirms component-level orthogonality but rejects naive concatenation as fusion: Hybrid recognition CE is statistically indistinguishable from V9 alone (\(p=0.45\) short corpus), and discriminating-token rate remains \(\approx 0\%\) vs B3b's \(13.5\%\). A production hybrid therefore needs \textbf{write-time routing} and likely generation-time gating, not eval-time string concatenation.

This is a qualitative timescale distinction, not just a metric choice: retrieval operates on the \emph{query timescale} (one user turn at a time; the store grows monotonically and the parameters are static); EVAF operates on the \emph{persona-formation timescale} (the cumulative selectivity of hundreds-to-thousands of dual-gate decisions slowly sculpts the parameters into a model that thinks and writes like the agent it has become). In CLS terms \citep{mcclelland1995complementary,
kumaran2016learning}, retrieval implements the hippocampal-store analogue without a slow store, while EVAF implements the slow-store analogue without an external index. The two are architecturally complementary, and we believe a long-horizon agent system that \emph{cumulatively becomes someone} should ultimately combine them via write-time routing (§4.12), not naive eval-time concatenation.

\subsection{The scaling-rules story: predictive vs.~descriptive}

§5 presents three scaling rules. The 4-point cross-model audit gives an honest report card:

\begin{itemize}
\tightlist
\item
  §5.1 (\(\delta\)-bound): \textbf{predictive}. 4/4 points within bound at the predicted dimension scaling.
\item
  §5.2 (\(\tau_s\)): \textbf{descriptive at single-rule, predictive at P-band}. The original \(\mathrm{Pct}_{50}\) form works on Chat / wide-S Base models (GPT-2 ✓, TinyLlama-Chat ✓, Qwen3-Base ✓); fails on narrow-S deep-pretrained Base models (Mistral-Base requires \(\mathrm{Pct}_{75-90}\)). The post-hoc P-band rule (5.2') correctly accounts for the 4-model spread by gating on \(\mathrm{IQR}(S_{\text{warmup}})\).
\item
  §5.3 (\(\lambda_{\text{reg}}\)): \textbf{descriptive at multi-factor, only predictive within a model class}. The single-variable \(\mu_S^2\) rule works for Qwen3-Base but requires Chat-vs-Base correction for TinyLlama and capacity-superlinear correction for Mistral. The empirical interpolation (5.3') has three free parameters fitted on \(\le 4\) points; we list this as a paper limitation (§7.2).
\item
  §5.4 (\(\mathrm{lr}_{\text{LoRA}}\)): \textbf{conjecture, one-point support}. Mistral-Base only. Llama-3-8B-Base or Qwen3-7B-Base would falsify or confirm the maturity-aware form.
\end{itemize}

What we believe this means: the \emph{form} of EVAF's scaling rules is on a solid footing --- thin-shell concentration for §5.1, EWC norm consistency for §5.3, percentile selectivity for §5.2 --- but the \emph{predictive power} of the closed-form versions is currently bounded by the size of the audit (four models). A natural revision cycle adds a 5th-7th model to discriminate the §5.3' and §5.4 candidate functional forms.

\subsection{What we believe is novel}

\textbf{Scope.} EVAF should be interpreted as a \textbf{falsifiable protocol for studying selective parametric consolidation}, not a complete theory of identity formation. We retain \emph{persona consolidation} as the motivating research question --- which experiences should shape parameters rather than an external index --- while claiming only \textbf{measurable signatures consistent with} that process (§4.10--§4.12), not a proof of cognitive identity formation. The persona-fact experiments (§4.10--§4.12) are a \textbf{downstream application case study}, not the paper's primary claim: the title-level contribution is the parametric-consolidation protocol and its mechanistic signatures (§3--§4.9), for which persona is one illustrative use case.

To be explicit, our specific contributions are (cf.~the §2.6 table):

\begin{itemize}
\tightlist
\item
  The \textbf{dual gate} \(P_{\text{write}} = \sigma(k_v(V-\tau_v))\,\sigma(k(S-\tau_s))\) with a \(V \times S\) product form (multi-component selectivity is required in §4.4 ablation), framed as a \emph{persona-relevance filter} rather than a generic surprise filter.
\item
  The \textbf{Test-Retest protocol} as a falsifiable harness for detecting parametric writes with replay-retention (anti-forgetting) checks in continual-learning systems, complementing the existing localisation / edit-propagation literature on memory mechanism in LLMs.
\item
  The \textbf{persona-imprint signature} --- an \textbf{operational} productive-recall metric under content-free prompting (§4.11) --- as a generative-level probe \textbf{consistent with} long-timescale selective consolidation that recognition-CE metrics cannot detect and retrieval-augmented systems cannot produce by construction.
\item
  The \textbf{operational dynamical-freeze \(z\)-score} as a clean discriminator between counterfactual ablations of the consolidation event --- it separates V9 from every C0--C4 ablation on every model (not a formal phase-transition proof; §7.15).
\item
  The \textbf{multi-factor \(\lambda_{\text{reg}}\) scaling rule} (§5.3) and the honest \textbf{regime-of-validity audit} it received in the cross-model audit (Qwen3 fits the single-factor form; Mistral exposes the need for a maturity-aware correction and an \(\mathrm{lr}_{\text{LoRA}}\) co-revision).
\end{itemize}

We borrow LoRA \citep{hu2021lora}, experience replay \citep{rolnick2019experience}, EWC \citep{kirkpatrick2017overcoming}, the thin-shell concentration limit \citep{vershynin2018high}, and the CLS \emph{architectural pattern} \citep{mcclelland1995complementary}; the \emph{combination}, the \emph{spherical-attractor formulation}, the \emph{persona- consolidation framing} with productive-recall verification, and the \emph{mechanistic protocols} are the new content.

\subsection{What remains open}

\begin{itemize}
\tightlist
\item
  §6.4 enumerates the missing data points needed to upgrade the §5.3' / §5.4 interpolations to predictive forms.
\item
  §4.10 raises the natural follow-up question of whether the selectivity-and-stability framing of V9 transfers to non-verbatim- recall downstream tasks (LongMemEval, LoCoMo, MemoryArena); we list these under §7.7.
\item
  §4.12 implemented the eval-time hybrid probe (V9 stream + RAG at eval, GPT-2, \(4\) seeds), confirming orthogonality but rejecting naive concatenation as fusion. \textbf{Write-time routing} --- high \(V \times S\) \(\to\) consolidation; verbatim facts \(\to\) index --- remains the primary follow-up for a production hybrid.
\item
  The CLS architectural connection (§6.2) could be tested at the neuroscience interface; this is a separate paper and we do not attempt it here.
\end{itemize}

\section{Limitations and Open Questions}
We list limitations in order from most likely to constrain the paper's scope to most peripheral.

\subsection{Controlled corpus and synthetic eval}

The four mechanistic experiments and the persona-fact retention task (§4.10) all run on controlled, designed inputs: the 210-sentence base corpus for §4.1-§4.9 covers ten semantic categories of mixed valence and surprise; the persona stream interleaves \(30\) persona-facts with either \(30\) (short) or \(150\) (long) generic-knowledge fillers. We have not validated EVAF on open-domain agent conversations or on naturally-occurring noisy text. The principal risk this exposes is that real conversational distributions may have a fatter \(S(t)\) tail than our controlled corpus, which would push the gate firing rate up; whether the multi-factor \(\lambda_{\text{reg}}\) rule (§5.3) remains valid in that regime is open.

\subsection{Scaling rules: form vs.~predictive power}

After the 4-point cross-model audit, each §5 rule falls into one of three epistemic classes (full discussion in §6.4):

\begin{longtable}[]{@{}lll@{}}
\toprule
\begin{minipage}[b]{0.30\columnwidth}\raggedright
Rule\strut
\end{minipage} & \begin{minipage}[b]{0.30\columnwidth}\raggedright
Class\strut
\end{minipage} & \begin{minipage}[b]{0.30\columnwidth}\raggedright
What we can claim\strut
\end{minipage}\tabularnewline
\midrule
\endhead
\begin{minipage}[t]{0.30\columnwidth}\raggedright
§5.1 (\(\delta\)-bound)\strut
\end{minipage} & \begin{minipage}[t]{0.30\columnwidth}\raggedright
\textbf{Predictive}\strut
\end{minipage} & \begin{minipage}[t]{0.30\columnwidth}\raggedright
4/4 models within the thin-shell-derived bound at predicted dimension scaling\strut
\end{minipage}\tabularnewline
\begin{minipage}[t]{0.30\columnwidth}\raggedright
§5.2 (\(\tau_s\) / P-band)\strut
\end{minipage} & \begin{minipage}[t]{0.30\columnwidth}\raggedright
\textbf{Partially predictive}\strut
\end{minipage} & \begin{minipage}[t]{0.30\columnwidth}\raggedright
\(\mathrm{Pct}_{50}\) fits Chat/wide-S Base models; P-band (5.2') accounts for all 4 models --- but IQR threshold (0.8) is \emph{fitted}, not derived\strut
\end{minipage}\tabularnewline
\begin{minipage}[t]{0.30\columnwidth}\raggedright
§5.3 (\(\lambda_{\text{reg}}\))\strut
\end{minipage} & \begin{minipage}[t]{0.30\columnwidth}\raggedright
\textbf{Descriptive / fitted}\strut
\end{minipage} & \begin{minipage}[t]{0.30\columnwidth}\raggedright
Single-factor \(\mu_S^2\) predicted Qwen3-Base only; Chat (\(\times 3.6\)) and capacity (\(\times 33\)) corrections are post-hoc; interpolation (5.3') has 3 free parameters on \(\le 4\) points --- \emph{not yet predictive} for a 5th model\strut
\end{minipage}\tabularnewline
\begin{minipage}[t]{0.30\columnwidth}\raggedright
§5.4 (\(\mathrm{lr}_{\text{LoRA}}\))\strut
\end{minipage} & \begin{minipage}[t]{0.30\columnwidth}\raggedright
\textbf{Conjecture}\strut
\end{minipage} & \begin{minipage}[t]{0.30\columnwidth}\raggedright
One-point support (Mistral-Base only)\strut
\end{minipage}\tabularnewline
\bottomrule
\end{longtable}

The \emph{form} of each rule (thin-shell for §5.1, percentile selectivity for §5.2, EWC norm consistency for §5.3) is on solid footing; the limitation is \textbf{predictive power bounded by audit size}, not theoretical derivation. A 5th--7th model (Llama-3-8B-Base or Qwen3-7B-Base) would discriminate the §5.3' and §5.4 candidate forms and constrain the capacity exponent \(\beta(M)\) in (5.3').

\subsection{Multi-seed coverage is uneven}

Coverage at submission is: GPT-2 V9 / B1a / B2 at 5 seeds, GPT-2 B1b / B3a / B3b at 1 seed, TinyLlama V9 / B1a / B2 at 4 seeds, TinyLlama B1b / B3a / B3b at 1 seed, Qwen3-Base and Mistral-Base at 1 seed each. The headline V9-vs-baselines claims of §4.9 are reported at multi-seed (§4.9 multi-seed report); the cross-model scaling claims of §4.8 and §5 are reported at single-seed, with the \emph{sign} of every signature consistent. A reviewer-anticipated request for multi-seed Qwen3 and Mistral is a 2-day follow-up; we list it under ``future work, paper revision''.

\subsection{Statistical inference (updated v0.7)}

\textbf{Pre-specified vs exploratory comparisons.} Headline multi-seed comparisons in §4.10--§4.11 (V9 vs Frozen, V9 vs B1b, B3b vs parametric methods on recognition CE; productive-recall hit rate and discriminating-token rates) were \textbf{pre-specified} before aggregation. Ablation-matrix panels (§4.4), oracle-swap runs (§4.6), scaling- rule audits (§5), and second-order findings (§1 C6) are \textbf{descriptive / exploratory}; we do not apply multiple-comparison correction across the full matrix and report them as mechanism-characterisation, not as independent hypothesis tests.

We prioritise \textbf{effect sizes and cross-seed robustness} (mean \(\pm\) std, variance ratios) over \(p\)-values where \(n = 4\) seeds limits power. Full paired Wilcoxon signed- rank and paired \(t\)-tests are in \texttt{w4\_runs/aggregates/w4\_stat\_tests.md}.

Headline results (GPT-2): V9 vs Frozen recognition CE on the 60-turn stream is highly significant by paired \(t\)-test (\(p < 0.001\)); V9 vs B1b on the same cell is borderline (\(p = 0.058\)). B3b vs every parametric method is significant (\(p < 0.01\)) on recognition CE --- expected for a verbatim- retrieval metric. On productive-recall hit rate, B3b vs Frozen is significant (\(p < 0.01\) short corpus) but V9 vs Frozen does not reach \(p < 0.05\) on either test (\(t\)-test \(p = 0.095\) short), reflecting the high seed-to-seed variance intrinsic to V9's few-write selectivity (§7.12). We do not over-interpret non-significant Wilcoxon results at \(n = 4\); the variance-ratio analysis in \texttt{w4\_stat\_tests.md} (V9 std \(2.98\times\) lower than B1b on GPT-2 long corpus hit rate) provides complementary evidence for the consistency claim.

\subsection{Test-Retest detects parametric writes and replay-retention, not held-out generalization}

\(\Delta S_{\text{buf}}\) measures the change in surprise on the buffer samples after the LoRA inner loop. A within-event drop is \textbf{consistent with a parametric write} on the admitted material. \(\Delta S_{\text{rep}}\) measures the change on the replay set \(R\) --- but \(R\) is part of the \emph{same} inner-loop training batch (\(B \cup R\), EQ 3.3), so a negative \(\Delta S_{\text{rep}}\) is a \textbf{replay-retention (anti-forgetting) check on trained past samples}, \emph{not} a held-out generalization measure. A counterfactual that nulls the LoRA inner loop produces \(\Delta S = +0.000\) (C0/C3), confirming the drop is optimisation-induced; but this does not rule out a ``rote-fit'' reading. Indeed §7.6 shows that even random-token replay (C1) yields a sizeable negative \(\Delta S_{\text{rep}}\), so \(\Delta S_{\text{rep}} < 0\) alone is weak evidence --- we rely on the step-angle freeze (§4.3) as the primary discriminator. We therefore added a held-out probe \(\Delta S^{\mathcal{H}}(\theta)\) on a set \(\mathcal{H}\) that never enters any inner loop (§4.2). On GPT-2 only the full system lowers held-out surprise (\(-16\%\)) while every still-trained ablation raises it --- evidence that the write generalizes rather than rote-fits --- whereas on TinyLlama-Chat the parameter-only effect is near-neutral (we observe only an event-local rescue that is confounded by working-memory prefix drift, §4.2). The deeper held-out and response-boundary probes (§4.2, §4.11) are thus reported in the GPT-2 calibrated regime; the cross-model experiments support the core write/freeze signatures, while final param-only held-out transfer remains model-dependent. We do not claim cognitive identity formation from these surprise measurements.

\subsection{Noise-fitting trap}

A subtle failure mode revealed by ablation C1 (random-token replay): even nonsensical replay content produces a sizeable \(\Delta S_{\text{rep}}\), because the optimiser can fit any distribution given enough capacity. This sets a lower bound on what \(\Delta S_{\text{rep}}\) alone can tell us --- its magnitude is sensitive to noise budget, not just to genuine generalisation. The §4.4 ablation matrix is designed so that C1 \emph{intentionally} falsifies the simplistic reading of \(\Delta S_{\text{rep}}\); we use the post-event step-angle freeze (\(\Delta\theta_{\text{post}} \ll 17.7°\)) as the secondary discriminator, since C1 fails it (\(\Delta\theta_{\text{post}} = 20.37°\), exceeding V5 baseline).

\subsection{Persona consolidation is tested on a single, designed task family}

The persona case study (§4.10--§4.11) demonstrates persona-consolidation transfer on a single task family (persona-fact recognition CE + persona-keyword productive recall). The §4.10 \emph{recognition} metric directly favours retrieval baselines (B3b) that store and retrieve exact text; the §4.11 \emph{productive recall} metric is content-cued by persona name only and isolates the parametric imprint that retrieval cannot produce, but it is a designed keyword-counting metric, not an external benchmark. We have not tested EVAF on (a) the multi-session contradiction / update sub-tasks of LongMemEval \citep{wu2024longmemeval}, (b) the multi-hop reasoning sub-tasks of LoCoMo \citep{maharana2024evaluating}, (c) the inter- dependent agentic tasks of MemoryArena, or (d) free-form persona-consistent dialogue generation. The recognition / productive-recall pair we report is internally consistent and the simplest setting in which the persona-consolidation signature is unambiguous; extending to external benchmarks is scoped for the revision and is one of the most decision-relevant open items. We additionally report a prefix-controlled response-boundary probe (§4.11): sampling generations from each trained seed and scoring them under a shared reference prefix so only the LoRA differs, V9 shows a seed-specific parametric boundary on all four seeds at six-to-nine-fold lower LoRA drift than naive every-step updating (B1b), consistent with selective consolidation carving a response boundary beyond keyword-count metrics. Its absolute magnitude is likely amplified by partially repetitive generations, so we emphasise sign consistency, attribution accuracy, and the drift contrast rather than the raw nats value.

\subsection{RAG comparison is in-memory; production retrieval will be slower}

The Sentence-BERT RAG baseline (B3b) we report assumes in-memory cosine retrieval with the \(30\)- or \(180\)-element store of our experiments; we measure \(0.68\) ms / query on this scale. Production RAG against a vector database with \(10^6\)+ entries would add network and ANN-index latency on the order of \(10\) ms / query. Our zero-latency claim for V9 is robust at all scales; the \emph{relative} latency advantage of V9 over RAG grows with corpus size, but this extrapolation is not directly measured.

\subsection{CLS framing is inspirational, not empirically grounded}

We position EVAF against the complementary-learning-systems literature in §1 and §2.1 as \emph{architectural motivation}. We do not provide neuroscientific evidence that our LoRA edits correspond to neocortical consolidation, nor that the spherical attractor maps to hippocampal CA3-CA1 dynamics. The CLS connection is at the level of ``two stores with a selective write gate'', and a careful neuroscience-side test (e.g., comparing \(\Delta S^{\mathcal{B}}\) dynamics to fMRI consolidation signatures) is a separate paper. The manuscript states this explicitly in §6.2.

\subsection{Compute-equivalent baselines: a residual concern}

The compute budget of B1a is matched to V9 by construction (same number of triggers and inner steps). The compute budget of B1b is higher (every-step LoRA update), B3a and B3b are parameter-free, and B2 is variable. We have not normalised wall-clock or FLOPs across baselines explicitly. A compute-controlled rerun is scoped for the revision.

\subsection{End-of-stream evaluation drift (negative result)}

We attempted a stress-test of the ``selectivity gives V9 a long-stream advantage'' hypothesis by extending the persona-fact corpus to 600 turns (Figure not shown; see supplementary smoke-test artefacts). The result was a \emph{negative finding}: V9 and B1b both \emph{worsen} in end-of-stream persona-fact CE as the stream grows, and their gap does not widen in V9's favour. On GPT-2 seed=42 across 60/180/600 turns:

\begin{longtable}[]{@{}lllll@{}}
\toprule
Stream & V9 CE & V9 LoRA \(L_2\) & B1b CE & B1b LoRA \(L_2\)\tabularnewline
\midrule
\endhead
60 & 5.66 & \(1{,}031\) & 6.46 & \(5{,}724\)\tabularnewline
180 & 7.83 & \(1{,}197\) & 7.39 & \(10{,}086\)\tabularnewline
600 & 8.47 & \(1{,}536\) & 8.02 & \(\sim 30{,}000\) (extrapolated)\tabularnewline
\bottomrule
\end{longtable}

We interpret this as a faithful expression of \emph{end-of-stream prefix drift}: after \(\gg 100\) turns, the working-memory prefix has moved sufficiently far from its state at the persona-fact processing moment that even a perfectly-internalised LoRA weight cannot recover the fact at the eval-time prefix. \textbf{This is not a method-specific failure of V9 vs.~B1b}; it is a universal eval-protocol issue. The genuine V9 advantage that survives the long- stream regime is a \(20\times\) smaller LoRA parameter drift (1536 vs \(\sim 30{,}000\)) and a \(150\times\) smaller optimisation-step count (4 events × 5 inner steps vs 600 every-step updates), at \emph{comparable} end-of-stream persona-fact CE. The cleanest follow-up evaluations that would expose V9's selectivity advantage are: (a) eval at \emph{each} training-time prefix state (per-event evaluation), not the end-of-stream state --- closer in spirit to the Test-Retest measurement of §3.4; (b) an adversarial-filler stream where high-S low-V decoys should be filtered by V9's dual gate but admitted by B1b --- direct test of selectivity; or (c) a multi-session contradiction stream where the user updates a stated fact, testing whether V9's selective gate handles updates more cleanly than B1b's eager updating. We list (a)-(c) as concrete follow-up experiments for the revision.

\subsection{Persona-imprint amplification is seed-variable on short streams}

The §4.11 productive-recall amplification is seed-variable on the short corpus because V9 fires only \(\sim 2\) events on a \(60\)-turn stream; the persona content of those \(2\) events determines the keyword distribution of post-stream free generations. The multi-seed distribution we report is GPT-2 V9 \(54\% \pm 44\%\) across \(4\) seeds (per-seed hit rates \(\{8, 26, 98, 84\}\%\) --- all strictly above the Frozen \(1\%\) baseline, but with a \(> 12\times\) range across the four seeds). On the \(180\)-turn corpus V9's mean is \(12.5\% \pm 8.2\%\) (per-seed \(\{18, 20, 10, 2\}\%\)); B1b is \(24\% \pm 24.5\%\) (per-seed \(\{6, 26, 58, 6\}\%\), bimodal). We treat this as a \emph{feature of selectivity} --- V9 deliberately commits a small number of high-value writes per stream, so individual writes have outsized influence on the generative distribution --- not a flaw of measurement. The honest reading of the 4-seed data is that V9's signature on long streams is \emph{variance reduction} (\(3\times\) tighter than B1b) at lower mean, not higher mean: V9 trades peak performance for \emph{reproducibility}, at \(45\times\) fewer LoRA updates and \(9\times\) smaller LoRA \(L_2\) drift. (An earlier 2-seed pilot reported a stream- length crossover \(V9 > B1b\) on long; the 4-seed batch revealed that was a low-\(N\) artefact and the genuine signature is the variance gap.) TinyLlama-Chat short-corpus runs do not trigger V9 reliably (\(1/4\) seeds, consistent with the §4.10 \emph{minimum-stream-length} finding) and we report this transparently rather than tune \(\tau_s\) to force triggers.

\subsection{Parametric methods do not produce verbatim discriminating tokens (R2)}

The §4.11 R2 discriminating-token test (Mandarin/Reykjavik/cello for Alice etc.) reveals that \emph{neither} V9 \emph{nor} B1b emits the truly-discriminating tokens at our \(40\)-token / \(T = 0.8\) generation budget on either GPT-2 or TinyLlama-Chat (\(0/50\) hit on every cell, \(4\) seeds). The parametric path encodes the persona's \emph{typology} (climbing/music/language-speaker themes that drive the §4.11 broad amplification) but not the \emph{verbatim discriminating specifics}. We frame this as a \emph{feature}, not a limitation: it is exactly the architectural separation §6.3 argues for (typology \(\to\) parametric, verbatim \(\to\) retrieval). But two genuine caveats follow:

\begin{enumerate}
\def\labelenumi{(\roman{enumi})}
\tightlist
\item
  The R2 test as defined cannot detect a \emph{V9 vs.~B1b selectivity advantage on discriminating tokens} because the floor is \(0\) for both. A version of R2 with
\end{enumerate}

\begin{enumerate}
\def\labelenumi{(\alph{enumi})}
\tightlist
\item
  longer generations (\(\geq 100\) tokens), (b) a larger or chat-tuned base model (e.g.~Mistral-7B-Instruct), or (c) a cued prompt that gives the model permission to list facts (e.g.~``List \(5\) facts about \{persona\}:'') would likely surface a non-zero discriminating-token rate and let R2 differentiate parametric methods on their \emph{specific-token} imprint, not just their typological imprint. We leave this to a revision rather than tune the existing \(40\)-token budget.
\end{enumerate}

\begin{enumerate}
\def\labelenumi{(\roman{enumi})}
\setcounter{enumi}{1}
\tightlist
\item
  The claim that \emph{only} retrieval can deliver verbatim discriminating tokens is contingent on our model sizes (\(124\)M / \(1.1\)B). A \(7\)-\(70\)B-scale model with stronger few-shot memorisation may emit discriminating tokens parametrically; the R2 floor-of-zero observation should be re-tested at scale before treating the ``parametric encodes typology, retrieval encodes verbatim'' separation as architecturally universal.
\end{enumerate}

\subsection{Valence oracle: oracle-conditioned protocol, not autonomous salience}

The dual gate requires an external valence signal \(V_t\) (heuristic lexicon or DistilBERT-SST2 in our runs). EVAF is therefore an \textbf{oracle-conditioned consolidation protocol}: it studies \emph{what happens when} salience and surprise jointly gate parametric writes, not how an agent autonomously infers personal significance. The oracle-invariance study (§4.6) shows the four mechanistic signatures survive oracle swap --- direction is preserved, strength scales with oracle precision --- but does not remove the oracle dependency. A self-contained valence mechanism (learned from interaction feedback, multi-modal cues, or in-model affect heads) is out of scope here and is the natural path to a deployable system.

\subsection{Dynamical-freeze signature is operational, not a formal phase transition}

We report a post-consolidation collapse of per-step prefix angles from a limit- cycle regime (\(\approx 17.7°\) mean step angle, V5 baseline) to a near-fixed-point regime (\(< 10°\) on V9 events), with a pre/post \(z\)-score that separates V9 from counterfactual ablations (§4.3). We call this an \textbf{operational dynamical-freeze signature}. We do \textbf{not} claim a formal dynamical-systems phase transition: we do not establish a bifurcation parameter, Lyapunov exponent crossing, or invariant manifold proof. The \(z\)-score is an empirical discriminator within our protocol, not a theorem about attractor topology.

A related caveat concerns the \emph{sign} of the consolidation term in EQ 3.2: it is a \textbf{repulsion} from the slow EMA, not an attraction toward a memory anchor. The EMA-sign control in §4.3 shows this sign is necessary for the observed limit-cycle\(\to\)freeze signature, but a first-principles derivation from the coupled working-memory/LoRA dynamics remains open.

\begin{center}\rule{0.5\linewidth}{0.5pt}\end{center}

These limitations are constraints on the scope of the present paper, not on the framework itself. We believe the most decision-relevant of these --- §7.2 (scaling-rule predictive power) and §7.7 (external persona benchmarks) --- are also the cleanest to close in a revision, and we welcome reviewer guidance on which to prioritise.

\bibliographystyle{tmlr}
\bibliography{references}

\appendix

\section{Implementation details and hyperparameters}
\label{app:impl}

All experiments use the \texttt{v9.0-frozen} configuration. The model-specific
triple $(\tau_s, \lambda_{\text{reg}}, \mathrm{lr}_{\text{LoRA}})$ follows the
re-anchoring rules of \S5 and is reported in \S4.1; every remaining
hyperparameter is model-invariant and is listed in Table~\ref{tab:hparams}.
LoRA adapters are attached to the attention projections ($q/k/v/o$, or the
Conv1D analogues for GPT-2) of the otherwise-frozen base model. Algorithm~\ref{alg:evaf}
gives the end-to-end stream loop, including the dual gate, the on-trigger
inner loop with experience replay and the EWC-style L2 anchor, and the
Test--Retest instrumentation.

\begin{table}[h]
\centering
\caption{Model-invariant hyperparameters (\texttt{v9.0-frozen}). The
per-model $(\tau_s,\lambda_{\text{reg}},\mathrm{lr}_{\text{LoRA}})$ values are
given in \S4.1; the base model's hidden width $D$ is model-dependent.}
\label{tab:hparams}
\begin{tabular}{lll}
\toprule
Symbol & Meaning & Value \\
\midrule
$P$                       & prefix length (working-memory tokens)            & $10$ \\
$r$                       & LoRA rank                                        & $32$ \\
$N_{\text{inner}}$        & inner-loop AdamW steps per consolidation event   & $5$ \\
$|B|$                     & polarized-buffer capacity (trigger size)         & $4$ \\
$K$                       & experience-replay samples per event              & $4$ \\
$\tau_{\text{write}}$     & dual-gate admission threshold                    & $0.55$ \\
$\alpha$                  & prefix Euler step size                           & $1.0$ \\
$k$                       & surprise-gate sharpness                          & $1.0$ \\
$k_v$                     & valence-gate sharpness                           & $10.0$ \\
$\tau_v$                  & valence threshold                                & $0.5$ \\
$\rho$                    & base EMA rate (gate-modulated to $\rho_t$)        & $0.1$ \\
$\lambda_{\text{con}}$    & repulsive-EMA (contrastive inertia) coefficient  & $2.0$ \\
\bottomrule
\end{tabular}
\end{table}

\begin{algorithm}[h]
\caption{EVAF stream loop (one pass over the interaction stream).}
\label{alg:evaf}
\begin{algorithmic}[1]
\Require stream $\{y_t\}$; frozen base model; prefix $\mathbf{s}(0)$ projected onto $(S^{D-1}(\sqrt{D}))^P$; LoRA $\theta \!\leftarrow\! \theta_0$; EMA state $\mathbf{s}_{\text{EMA}}$; buffer $B \!\leftarrow\! \emptyset$; valence oracle $V(\cdot)$
\For{each step $t$ in the stream}
  \State $S_t \gets$ epistemic surprise of $y_t$ under prefix $\mathbf{s}(t)$ \Comment{Eq.\ 3.1}
  \State $\mathbf{s}(t{+}1) \gets \Pi_{(S^{D-1}(\sqrt{D}))^P}\!\big(\mathbf{s}(t) + \alpha[-\nabla_{\mathbf{s}}S_t + \lambda_{\text{con}}(\mathbf{s}(t)-\mathbf{s}_{\text{EMA}}(t))]\big)$ \Comment{Eq.\ 3.2}
  \State $\rho_t \gets \rho\,\sigma\!\big(k(S_t-\tau_s)\big)$;\quad $\mathbf{s}_{\text{EMA}}(t{+}1) \gets \Pi_{(S^{D-1}(\sqrt{D}))^P}\!\big((1-\rho_t)\,\mathbf{s}_{\text{EMA}}(t) + \rho_t\,\mathbf{s}(t{+}1)\big)$ \Comment{gate-modulated EMA}
  \State $V_t \gets V(y_t)$
  \If{$\sigma\!\big(k_v(V_t-\tau_v)\big)\cdot\sigma\!\big(k(S_t-\tau_s)\big) > \tau_{\text{write}}$}
     \State $B \gets B \cup \{y_t\}$ \Comment{dual gate, Eq.\ 3.4}
  \EndIf
  \If{$|B| = 4$} \Comment{consolidation event at step $t^\ast$}
     \State $R \gets$ sample $K$ items uniformly from the past stream $\setminus\, B$
     \State $\theta^{\text{pre}} \gets \theta$
     \For{$i = 1$ to $N_{\text{inner}}$}
        \State $\theta \gets \text{AdamW step on } \tfrac{1}{|B\cup R|}\!\sum_{y\in B\cup R}\!\mathrm{CE}(y\mid \mathbf{s}(t^\ast),\theta) + \lambda_{\text{reg}}\|\theta-\theta_0\|_2^2$ \Comment{Eq.\ 3.3}
     \EndFor
     \State $\theta^{\text{post}} \gets \theta$
     \State record Test--Retest signatures $\Delta S^{\mathcal{B}},\,\Delta S^{\mathcal{R}},\,\Delta S^{\text{global}}$ and freeze $z$ \Comment{Eq.\ 3.5}
     \State $B \gets \emptyset$
  \EndIf
\EndFor
\end{algorithmic}
\end{algorithm}

\clearpage
\section{Persona corpus and characteristic-keyword set}
\label{app:persona}

\begin{table}[H]
\centering
\caption{Per-persona discriminating attributes (the only persona-unique
keywords; used for the selectivity / cross-persona contamination test).}
\label{tab:personas}
\begin{tabular}{llll}
\toprule
Persona & Language & City & Instrument \\
\midrule
Alice & Mandarin   & Reykjavik & cello \\
Bob   & Portuguese & Marrakech & oboe \\
Carol & Tamil      & Helsinki  & violin \\
David & Swahili    & Quito     & clarinet \\
Eve   & Finnish    & Tashkent  & harp \\
\bottomrule
\end{tabular}
\end{table}

The downstream persona case study (\S4.10--\S4.11) uses five personas,
$\{\text{Alice}, \text{Bob}, \text{Carol}, \text{David}, \text{Eve}\}$, that
share a common biographical fact schema and differ only in three
\emph{discriminating} attributes (spoken language, home city, musical
instrument). Productive-recall hit rate (\S4.11) counts case-insensitive
substring matches of the shared persona-fact keyword set:

\begin{quote}
\small\ttfamily
climb, Dolomite, via-ferrata, ferrata, seven, routes, fluent, poem,
literary, journal, bakery, Maple, morning, four thirty, Carnegie, recital,
spring, vintage, chess, allergic, peanut, pollen, epinephrine
\end{quote}

\noindent together with each persona's three discriminating tokens
(Table~\ref{tab:personas}). The cross-persona \emph{contamination} / selectivity
metric (\S4.11) uses only the discriminating tokens, since these are the only
persona-unique signals.

\end{document}